\documentclass[10pt,twocolumn,letterpaper]{article}

\PassOptionsToPackage{nameinlink}{cleveref}

\usepackage[pagenumbers]{cvpr} % To force page numbers, e.g. for an arXiv version

\definecolor{cvprblue}{rgb}{0.21,0.49,0.74}
\usepackage[pagebackref,breaklinks,colorlinks,allcolors=cvprblue]{hyperref}

\usepackage{graphicx}
\usepackage{amsmath}
\usepackage{amssymb}
\usepackage{booktabs}
\usepackage{bm}

\usepackage[table]{xcolor}
\usepackage{tikz, pgf, pgfplots}
\usetikzlibrary{spy,calc,arrows.meta,positioning,backgrounds,intersections}
\pgfplotsset{compat=1.18}
\usepackage{tikz-3dplot} % The package for 3D rotations
\usepackage{multirow}
\usepackage[normalem]{ulem}
\usepackage{colortbl}

\newlength{\figurewidth}

\newcommand{\boldparagraph}[1]{\smallskip\noindent{\bf #1}~}

\newcounter{inline}
\newcommand{\inc}{\stepcounter{inline}(\theinline)}

\newcommand{\ours}{Cross-View Splatter\xspace} % name of method

\usepackage[pagebackref,breaklinks,colorlinks]{hyperref}

\usepackage[capitalize]{cleveref}
\crefname{section}{Sec.}{Secs.}
\Crefname{section}{Section}{Sections}
\Crefname{table}{Table}{Tables}
\crefname{table}{Tab.}{Tabs.}

\def\paperID{30603} % *** Enter the Paper ID here
\def\confName{CVPR}
\def\confYear{2026}

\begin{document}

\title{\ours: Feed-Forward View Synthesis with Georeferenced Images}

\author{
Matias Turkulainen$^{1}$\footnotemark \and
Akshay Krishnan$^{2}$ \and
Filippo Aleotti$^{3}$ \and
Mohamed Sayed$^{3}$ \and
Guillermo Garcia-Hernando$^{3}$ \and
Juho Kannala$^{1,4}$ \and
Arno Solin$^{1,5}$ \and
Gabriel Brostow$^{6}$ \and
Daniyar Turmukhambetov$^{3}$ \\
{\normalsize
$^{1}$Aalto University \quad
$^{2}$Georgia Tech \quad
$^{3}$Niantic Spatial \quad
$^{4}$University of Oulu \quad
$^{5}$ELLIS Institute Finland \quad
$^{6}$UCL
}
}

%%%%%%%% Teaser Figure
\twocolumn[{%
\renewcommand\twocolumn[1][]{#1}%
\maketitle
\begin{center}
    \centering
    \captionsetup{type=figure}
    \vspace*{-2em}
    \includegraphics[width=0.85\textwidth,trim={0 2cm 0 0}, clip]{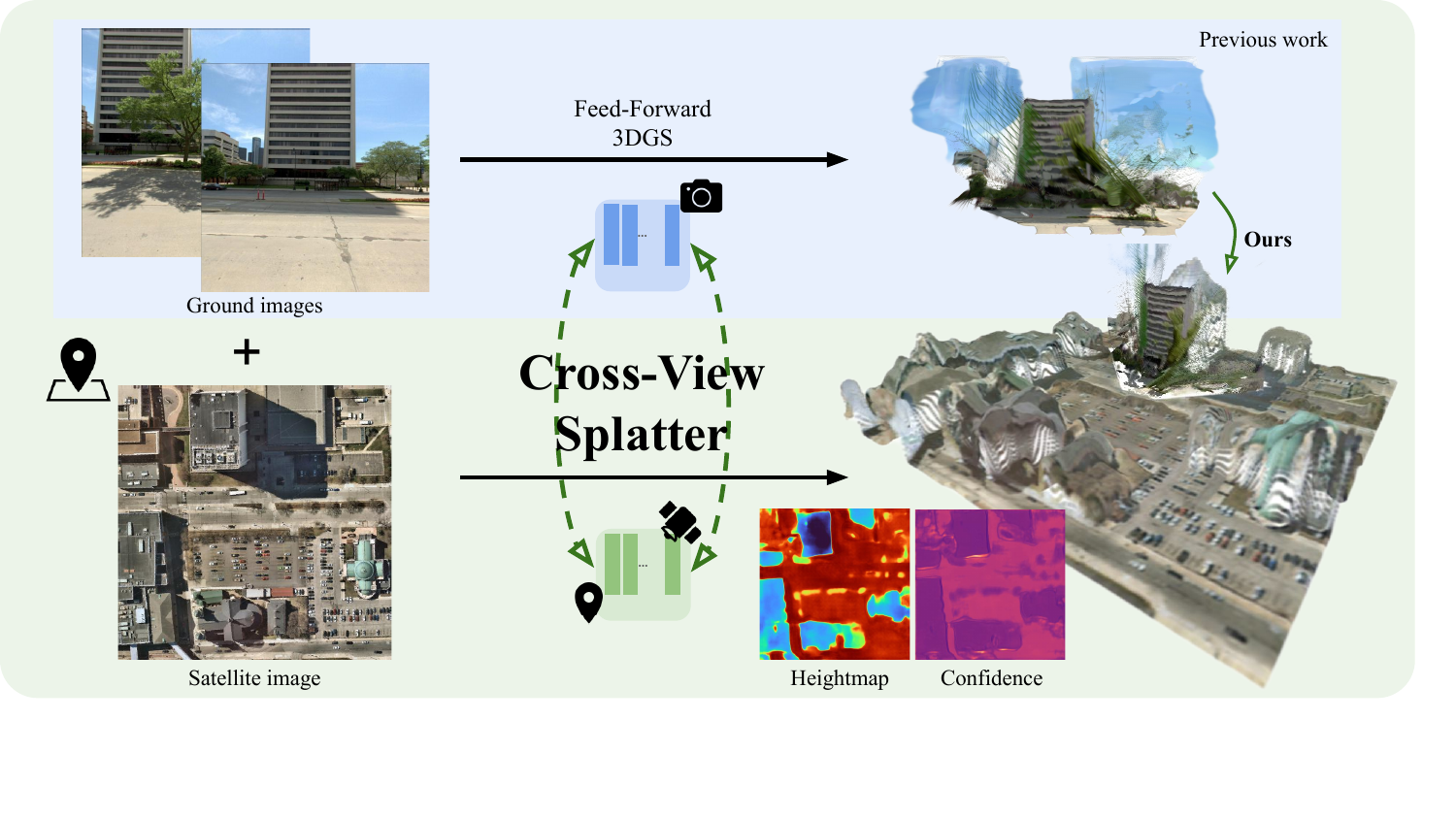}
    %\vspace*{-2em}
    %
    \captionof{figure}{\textbf{\ours} is a feed-forward model that predicts Gaussian splats for GPS-tagged ground level images and corresponding orthorectified satellite imagery from mapping services. It predicts Gaussian splats for both ground level and bird's-eye views in a unified coordinate system and supports multiple input images with unknown 6DoF poses. Only the GPS location of the ground level images is required.
    Our model improves scene coverage and novel-view synthesis compared to using ground imagery alone.}
    \label{fig:teaser}
    \vspace*{1em}
\end{center}%
}]
\renewcommand{\thefootnote}{*}
\footnotetext{Work done as an intern at Niantic Spatial.}

%%%%%%%%% ABSTRACT
\begin{abstract}
   We present \ours, a feed-forward method that predicts pixel-aligned Gaussian splats for outdoor scenes captured at ground level AND by satellite. Faithful reconstructions require good camera coverage, but ground imagery is time-consuming and hard to capture at scale for large outdoor scenes. Fortunately, satellite imagery can provide a global geometric prior that is easy to access via public APIs. \ours fuses orthorectified satellite views with GPS-tagged ground photos to predict Gaussian splats in a unified 3D coordinate frame. By aligning ground and bird's-eye feature representations, our model improves scene coverage and novel-view synthesis, compared to ground imagery alone. We train on curated georeferenced datasets and paired satellite--terrain data, mined from open mapping services. We evaluate our method on a new benchmark for novel-view synthesis with georeferenced imagery allowing comparison to prior state-of-the-art methods. Our code and data preparation will be available at \url{https://nianticspatial.github.io/cross-view-splatter/}.
\end{abstract}

%%%%%%%%% BODY TEXT
\section{Introduction}
\label{sec:intro}
Advances in 3D perception are transforming how visual systems model, interpret, and interact with the physical world. Generalized 3D reconstruction, in particular, has seen considerable success, with methods like \cite{wang2025vggt, Yang_2025_Fast3R, duisterhof2025mastrsfm, keetha2025mapanything, zhang2025flarefeedforwardgeometryappearance, wang2025continuous} allowing for accurate 3D point reconstruction using one or more images in a matter of seconds, with little to no prior information about camera calibrations. In addition, methods like \cite{xu2024depthsplat, zhang2025flarefeedforwardgeometryappearance, jiang2025anysplat, ye2024noposplat, chen2024mvsplat} have extended feed-forward 3D reconstruction to novel-view synthesis by predicting Gaussian attributes for pixel-aligned points, which can be rendered via 3D Gaussian splatting~\cite{kerbl3Dgaussians}, a desirable downstream task for 3D reconstruction. Despite this remarkable progress, a fundamental limitation still persists: these methods are designed, trained, and evaluated on ground level perspective imagery. This stems from the fact that widely used 3D foundation model training datasets consist predominantly of ground level calibrated views with aligned depth maps \cite{dai2017scannet, yeshwanth2023scannet++, kitti, Geiger2013IJRR, schoeps2017eth3d, reizenstein21co3d, ling2024dl3dv, cabon2020vkitti2, megadepth}. However, this reliance introduces significant scalability challenges, as modeling city-scale environments requires vast amounts of data that are both difficult to capture and computationally costly to process.

In contrast, satellite imagery provides a complementary perspective for outdoor captures. From a bird's-eye view (BEV), large-scale features such as roads and building footprints are more clearly delineated compared to ground-level imagery alone. High-quality satellite data is also readily available: mapping services such as Google Maps \cite{GoogleMaps}, Azure Maps \cite{AzureMaps}, and Esri World Imagery \cite{EsriWorldImagery2025} provide global surface imagery as fast, queryable tiled web maps (orthorectified satellite imagery).
Our core insight is that, for geo-localized ground-level captures, tiled web-map satellite imagery provides a strong prior for global scene structure and geometry beyond street-view observations. Nonetheless, compared with ground-level or UAV imagery \cite{dhaouadi2025ortholoc}, satellite imagery remains challenging due to its coarser spatial resolution and variation in weather, illumination, and season, making it impractical for conventional structure-from-motion and 3D Gaussian splatting frameworks. \looseness-1

We therefore leverage satellite imagery in a learned manner to capture low-resolution scene geometry. We introduce \ours, a feed-forward method that uses both ground-level imagery and orthographic satellite imagery from freely available mapping services. Given one or more ground-level images with GPS and heading metadata, our model queries a corresponding BEV image. Ground and satellite images are encoded into a unified feature space via cross-attention \cite{dosovitskiy2021image} with a strong pre-trained 3D reconstruction model \cite{wang2025vggt, jiang2025anysplat}. This feature space is then used to regress 3D Gaussian splat attributes. We backproject splats from ground images using perspective projection and from satellite images using orthographic projection, then merge them into a single 3D scene representation for novel-view rendering.

However, orthoimagery poses a fundamental challenge. Orthorectification removes perspective effects and parallax, making 3D reconstruction from tiled web map imagery alone impossible for classical MVS methods \cite{Leotta2019Urban, Qin2016RPC, lee2025skyfallgs, gao2024skyeyes}. Sparse overlap between surfaces visible in satellite and ground views also makes feature-based 3D reconstruction \cite{colmap} very challenging. To train our feed-forward method with suitable 3D data for ground and orthoimagery, we augment existing georeferenced ground-level datasets. We mine Digital Elevation Models (DEMs) and terrain data from public geological surveys~\cite{AzureMaps, USGSLidarExplorer, FlaiOpenLidarData} to create satellite-height map pairs, which provide a suitable 3D supervisory signal for satellite perspectives.

\boldparagraph{Contributions.}
To summarize, we provide the following contributions: \inc{} a novel feed-forward method capable of synthesizing Gaussian splats for both ground level perspective imagery and orthorectified satellite views, a first of its kind. We leverage satellite views alongside ground imagery to achieve greater coverage and a superior ability to extrapolate to novel views, an advantage over methods restricted to ground imagery alone. \inc{} Curated datasets with ground level imagery with georeferenced BEV satellite images and terrain height maps; and \inc{} a new challenging outdoor scenes benchmark of ground level images with aligned satellite views allowing comparison of state-of-the-art feed-forward 3D Gaussian splatting methods. 
%\fa{expand here}\gabe{+1}

\begin{figure*}[!t]
    \centering
    \includegraphics[width=0.85\linewidth]
    {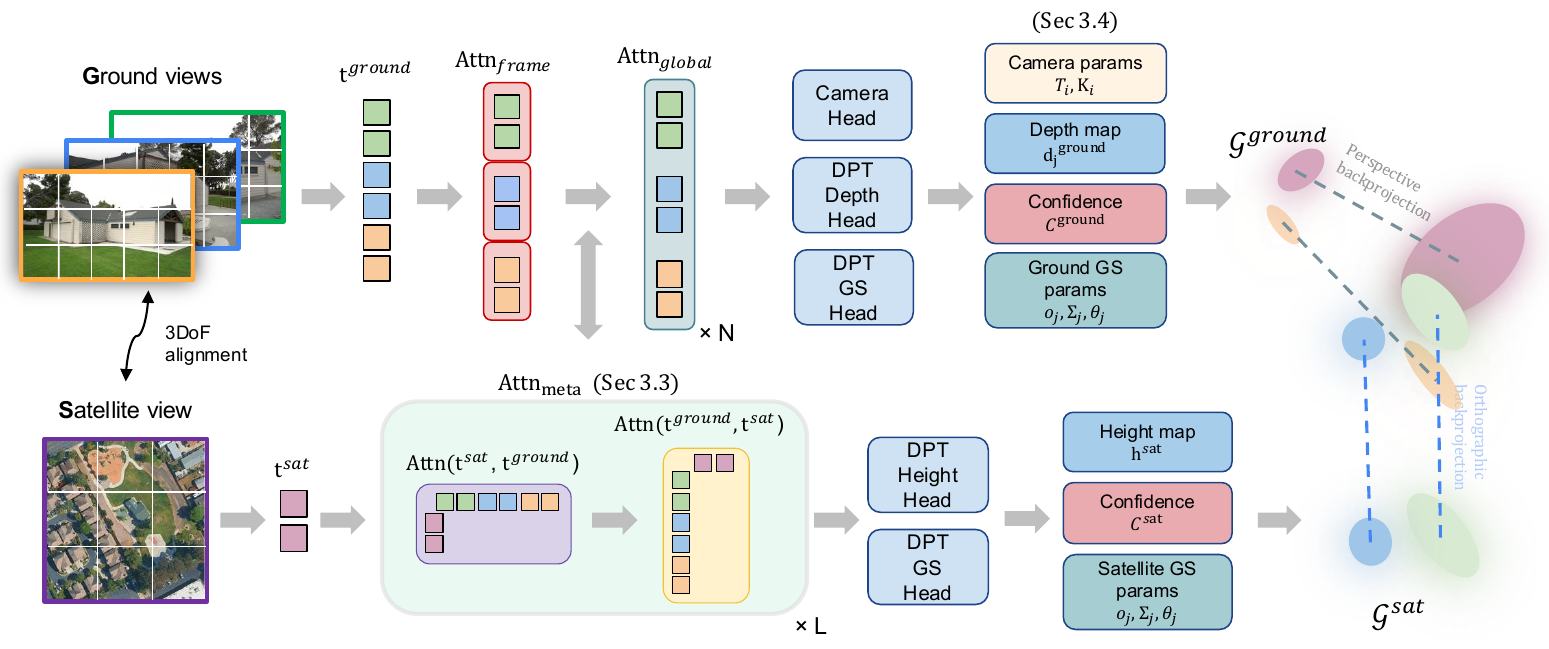} % left, bot, right, top
    \vspace{-1em}
    \caption{\textbf{Method overview}: Given geolocalized ground images and a single orthorectified satellite perspective, our model synthesizes 3D Gaussian splats in a shared coordinate frame. Ground views exchange information with satellite views within bidirectional cross-attention layers. Gaussians are predicted separately from ground and satellite branches, which are then combined into a unified coordinate frame. Although public elevation data is leveraged during training, only the satellite image and ground view(s) are necessary for inference.}
    \label{fig:main_figure}
    %\vspace*{-1em}
\end{figure*}

\section{Related Work}
\label{sec:related_work}

\boldparagraph{Feed-forward reconstruction.}
Feed-forward reconstruction refers to learning based methods for reconstructing the geometry of a scene via a single forward pass of a neural network. In contrast to classical structure-from-motion (SfM) \cite{colmap, pan2024glomap, SfMwithMRFs, Hartley2004Multiple} and multi-view stereo (MVS) \cite{schonberger2016pixelwise,galliani2015massively, gu2020cascade}  methods that reconstruct the scene with optimization schemes, feed-forward approaches are attractive for their speed and ability to generalize to challenging sparse inputs. The seminal work DUSt3R \cite{dust3r_cvpr24} introduced a ViT \cite{dosovitskiy2021image} encoder-decoder with self- and cross-attention that regresses two-view point maps in a canonical frame. Follow-up works \cite{Tang2025MVDust3R, Liu2025SLAM3R, Yang_2025_Fast3R, wang2025vggt, keetha2025mapanything, duisterhof2025mastrsfm, liu2025worldmirroruniversal3dworld} extended DUSt3R's two-view limited architecture to multi-view reconstruction. Methods like \cite{wang2025vggt, wang2025continuous, wang2025pi3, keetha2025mapanything, brachmann2024acezero} also regress 6DoF camera poses relative to an arbitrary reference image, with \cite{vuong2025aerialmegadepth} even using aerial perspective viewpoints. However, integrating orthoimagery remains an open and largely unexplored challenge.

\boldparagraph{Feed-forward novel-view synthesis.} Learning based methods for view synthesis have also been proposed. Early works adopted multi-plane images (MPI) \cite{zhou2018stereo} as a scene representation for small baseline novel-view synthesis. They included single-view methods \cite{single_view_mpi} and multi-view ones~\cite{srinivasan19, zhou2018stereo, wang2021ibrnet}, and utilized CNN based architectures for feature extraction and 3D regression. With the advent of NeRF's~\cite{mildenhall2020nerf}, methods like \cite{yu2021pixelnerf, du2023widerender, chen2021mvsnerf} use multi-view architectures with NeRF-like volume rendering for view-synthesis. 

The success of 3D Gaussian splatting \cite{kerbl3Dgaussians} motivated a series of works, notably \cite{chen2024mvsplat, xu2024depthsplat, charatan23pixelsplat, ye2024noposplat, jiang2025anysplat, wang2024freesplat, hong2024pf3plat, smart2024splatt3r}, that regress per-pixel Gaussian attributes rendered to novel views via splatting-based rasterization. Methods such as NoPoSplat~\cite{ye2024noposplat} and AnySplat~\cite{jiang2025anysplat} use strong pre-trained 3D foundation models \cite{dust3r_cvpr24, wang2025vggt} as backbones, distilling their priors for view synthesis. This is a scalable approach, especially for datasets without ground-truth geometry~\cite{zhou2018stereo,infinite_nature_2020}. However, 3D foundation models are not trained on satellite imagery and therefore struggle in our cross-view setting. We adopt a hybrid strategy: we use pre-trained foundation models \cite{wang2025vggt, jiang2025anysplat} for ground-level geometry tasks, and fine-tune additional satellite-specific layers to align satellite and ground-level features.

\boldparagraph{Satellite images for 3D reconstruction.} Prior work reconstructs city-scale 3D models from satellite imagery. Traditional methods \cite{Leotta2019Urban,Qin2016RPC,xu2024geospecificviewgeneration} use MVS on non-rectified satellite and aerial images to recover geometry and BEV camera models (RPCs). More recent methods \cite{lee2025skyfallgs, huang2025skysplat, gao2024skyeyes, horizongs} use Gaussian Splatting: SkySplat~\cite{huang2025skysplat} is a multi-image feed-forward method for satellite images, Horizon-GS~\cite{horizongs} jointly splats ground and aerial views, and Skyfall-GS~\cite{lee2025skyfallgs} jointly optimizes splats with diffusion-based refinement. SkyEyes~\cite{gao2024skyeyes} follows a similar setup and reconstructs meshes from multi-view satellite imagery with surface-aligned Gaussians~\cite{guedon2023sugar}, refined in ground-level views via denoising~\cite{zhang2023adding}. A key limitation is reliance on multi-view off-nadir, non-orthorectified imagery, which is hard to acquire and typically unavailable from mapping services such as Google Maps. To address this, \cite{Workman_2021_ICCV,hou2023enhancing} pair orthorectified satellite imagery with DEMs to predict satellite-aligned height maps, then convert them to voxel grids for ray-traced coarse ground-level depth.
We also estimate satellite-aligned height maps at inference time, but unlike methods that warp this geometry for depth prediction, we directly render it into ground-level views using Gaussian splatting.

%\todo{add Horizon-GS discussion}

\boldparagraph{Satellite-to-ground view-synthesis.} Generating ground-level novel views \textit{solely} from satellite imagery has also been studied. Shi \etal \cite{shi2022geometry} use relationships between orthographic views, height maps, and ground panoramas to warp satellite texture to ground level. Toker \etal \cite{Toker2021Coming} synthesize ground views with an intermediate polar transformation network. Xu \etal \cite{xu2024geospecificviewgeneration} reconstruct a scene mesh with MVS, project it to ground views, and texture it with a diffusion model. Sat2Density~\cite{sat2density, Sat2Density++} predicts a tri-plane representation~\cite{chan2022efficient} from top-down views that can be volume-rendered to known camera views, and unlike other methods can use a single satellite image. Inspired by these single-image methods, we adopt a Gaussian splat representation for top-down views that can be rasterized to novel views. Sky handling is also important: \cite{sat2density} regularizes sky regions in ground-level views to be opaque, whereas \cite{Sat2Density++} generates sky color with StyleGAN-2~\cite{Karras2019stylegan2}. As in \cite{sat2density}, we treat sky regions as part of the ground-level reconstruction model and promote sky Gaussians to be far away and opaque.

Generative satellite-to-ground methods have also been proposed. SatDreamer360 \cite{ze2025satdreamer360} predicts a tri-plane feature representation from BEV views, then synthesizes ground views by denoising a latent diffusion model \cite{rombach2022highresolution} with ray-based cross-attention. Sat2GroundScape \cite{xu2025groundscape} reconstructs a mesh from satellite views and textures it with satellite-guided denoising. Sat2Scene \cite{li2024sat2scene} and Sat2Vid \cite{li2021sat2vid} instead synthesize a point cloud from satellite views, then denoise and texture it with a 3D diffusion model. Streetscapes~\cite{deng2024streetscapes} generates ground-level depth and semantics from Google Maps height maps and semantic labels, then uses them to condition a video diffusion model that generates ground-level videos of city blocks. In contrast, feed-forward approaches like ours synthesize only visible regions in ground and satellite views without hallucinating unobserved areas.

\section{Model for Cross-View Splatter}
\subsection{Problem Formulation}
\boldparagraph{Goal.} 
Given an input sequence $(I_i^{\text{ground}})_{i=0}^N$ of ground level georeferenced imagery, we seek to reconstruct a 3D Gaussian splatting model in a feed-forward fashion with the aid of a \textit{single} satellite perspective. Georeferenced imagery refers to images tagged with 3DoF pose information, specifically GPS latitude, longitude, and heading information (available in most devices with an IMU~\cite{arnold2022mapfree}), which enable retrieval of a corresponding top-down orthographic view $I^{\text{sat}}$ with a known spatial resolution $r^{\text{sat}}$, with units of pixels per meter, from satellite mapping services~ \cite{GoogleMaps,AzureMaps,EsriWorldImagery2025}.

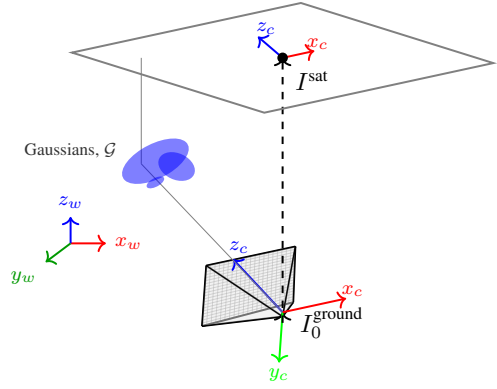
\begin{figure}[!t]
    \centering
    \resizebox{0.8\linewidth}{!}{%
        \begin{tikzpicture}[yscale=0.75]
    %Draw the camera frustum
   \tdplotsetmaincoords{70}{110}
    \usetikzlibrary{calc}
   \newcommand{\frustum}[3]{%
    % Use #3 for extra options like rotation
    \begin{scope}[scale=#1, shift={#2}, #3] 
        % Camera focal point
        \coordinate (F) at (0, .2);
    
        % Rectangle (image plane) coordinates
        \coordinate (A) at (-1.9, -.1);
        \coordinate (B) at (-1.8, 1.8);
        \coordinate (C) at (.3, 2.4);
        \coordinate (D) at (.25, .65);

        % Fill the plane with a pixel grid
        \begin{scope}
            % Create a clipping path so the grid is only drawn inside the plane
            \clip (A) -- (B) -- (C) -- (D) -- cycle;
            % Define the number of pixels/lines in the grid
            \def\numlines{20} 
            % Draw grid lines parallel to A--B and C--D
            \foreach \i in {0,...,\numlines} {
                \draw[gray, thin, opacity=0.5] ($(A)!\i/\numlines!(D)$) -- ($(B)!\i/\numlines!(C)$);
            }
            % Draw grid lines parallel to A--D and B--C
            \foreach \i in {0,...,\numlines} {
                \draw[gray, thin, opacity=0.5] ($(A)!\i/\numlines!(B)$) -- ($(D)!\i/\numlines!(C)$);
            }
        \end{scope}
    
        % Draw image plane
        \draw[black, thick] (A) -- (B);
        \draw[black, thick] (C) -- (D);
        \draw[black, thick] (B) -- (C);
        \draw[black, thick] (D) -- (A);
        
        % Draw lines connecting focal point to the image plane
        \draw[black, thick] (F) -- (A) (F) -- (B) (F) -- (C) (F) -- (D);
    
        % Fill
        \draw[fill=black!10,opacity=.5] (A) -- (B) -- (F) -- cycle;
        \draw[fill=black!10,opacity=.5] (B) -- (C) -- (F) -- cycle;
        \draw[fill=black!10,opacity=.5] (C) -- (D) -- (F) -- cycle;
    \end{scope}}
    
    % Place Cameras in scene
     \frustum{0.6}{(0,0)}{}
     \node[label=right:$I_0^{\text{ground}}$] at (0,0) {};

    % world point
    \coordinate (W) at (-2,3);
    % BEV camera
    \coordinate (B_bev) at (-2, 5);
    \coordinate (F) at (0, .2);
    
    % Draw camera coordinate directions
    \draw [thick, ->, every node/.style={font=\footnotesize, inner sep=0pt}]
      % z_c (Viewing axis, pointing from F toward the plane's center)
      (F) edge[blue] node [pos=1.1, anchor=south west] {$z_c$} ++(-0.7, 1.0)
      % y_c (Green axis, parallel to A--B)
      (F) edge[green] node [pos=1.1, anchor=north] {$y_c$} ++(-0.05, -0.95)
      % x_c (Red axis, parallel to B--C)
      (F) edge[red] node [pos=1.1, anchor=south] {$x_c$} ++(0.9, 0.26);

    \draw[gray] (F) -- (W);
    \draw[gray] (W) -- (B_bev);

    % second ground camera
    % \frustum{0.7}{(3, 2)}{rotate=8}

    % BEV CAMERA POINTING DOWN
    \coordinate (B_bev_center) at (0, 5);
    \draw[thick, dashed] (0,0) -- (B_bev_center);

    %% parallellogram
    \coordinate (X_vec) at (0.9, 0.26);
    \coordinate (Z_vec) at (-0.7, 0.8);
    \def\xscale{1.5*1.2}
    \def\zscale{1*1.2}
    %\path[draw=gray!70, fill=gray!10, opacity=0.6]
    %    ($(B_bev_center) + \xscale*(X_vec) + \zscale*(Z_vec)$) coordinate (C_top)
    % -- ($(B_bev_center) - \xscale*(X_vec) + \zscale*(Z_vec)$) coordinate (B_top)
    % -- ($(B_bev_center) - \xscale*(X_vec) - \zscale*(Z_vec)$) coordinate (A_top)
    % -- ($(B_bev_center) + \xscale*(X_vec) - \zscale*(Z_vec)$) coordinate (D_top)
    % -- cycle;

     %% axis
     % z_c axis
    \draw [thick, ->, every node/.style={font=\footnotesize, inner sep=0pt}]
      % z_c axis (half length)
      (B_bev_center) edge[blue] node [pos=1.1, anchor=south west] {$z_c$} ++(-0.35, 0.4)
      % x_c axis (half length)
      (B_bev_center) edge[red] node [pos=1.1, anchor=south] {$x_c$} ++(0.45, 0.13);
     
    \node[circle, fill, inner sep=1.5pt, label=south east:$I^{\text{sat}}$] at (B_bev_center) {};

    % --- 1. Pre-calculate the scaled vectors ---
    % We use \pgfmathsetmacro to do the math in your \def's
    \pgfmathsetmacro{\myxscale}{\xscale}
    \pgfmathsetmacro{\myzscale}{\zscale}
    
    % Calculate the final components of the new x-vector
    \pgfmathsetmacro{\xx}{\myxscale * 0.9}
    \pgfmathsetmacro{\yx}{\myxscale * 0.26}
    
    % Calculate the final components of the new y-vector (from your Z_vec)
    \pgfmathsetmacro{\xy}{\myzscale * -0.7}
    \pgfmathsetmacro{\yy}{\myzscale * 0.8}

    \coordinate (X_vec) at (0.9, 0.26);
    \coordinate (Z_vec) at (-0.7, 0.8);

    \begin{scope}[
    shift={(B_bev_center)}, % Move to the center point
        x={(\xx, \yx)},  % Set the new x-axis (your X_vec)
        y={(\xy, \yy)}   % Set the new y-axis (your Z_vec)
    ]
    \draw[gray!, thick] (-1,-1) -- (1,-1) -- (1,1) -- (-1,1) -- cycle;
    \end{scope}

    % SPLATS
    \newcommand{\blob}[6]{%
    \begin{scope}[xshift=#2cm, yshift=#3cm, fill opacity=0.5]
      \shade[rotate=#4, thick, inner color=#1, outer color=#1, scale=0.75, #6] (0, 0) ellipse (#5);
    \end{scope}
    }
    \begin{scope}[xshift=0cm,yshift=-0.75cm]
        \node[minimum width=3cm,minimum height=3cm,rounded corners=4pt,text width=3cm,align=center] (3d-scene) at (0,0) {};
        % blobs
        \blob{blue}{-1.8}{3.8}{40}{0.75cm and 0.4cm}{}{t1};
        \blob{blue}{-1.5}{3.7}{40}{0.3cm and 0.4cm}{}{t1};
        \blob{blue}{-1.8}{3.4}{40}{0.2cm and 0.1cm}{}{t1};

        \node[align=left, font=\scriptsize] at (-3,4) {\textcolor{black!80}{Gaussians, $\mathcal{G}$}};
    \end{scope}

    % world coordinate axis
    \def\axislen{0.5} % world coordinate axis
    \begin{scope}[shift={(-3, 1.5)}]
      % X (Red): Right
      \draw[red, thick, ->] (0,0) -- (\axislen,0) node[right] {\footnotesize $x_w$};
      % Z (Blue): Up
      \draw[blue, thick, ->] (0,0) -- (0,\axislen) node[above] {\footnotesize $z_w$};
      % Y (Green): Into the page (Slanted bottom-left based on typical perspective view)
      \draw[green!60!black, thick, ->] (0,0) -- ({cos(225)*\axislen}, {sin(225)*\axislen}) node[below left] {\footnotesize $y_w$};
    \end{scope}

    \end{tikzpicture}
    }
    \vspace{-4em}
    \caption{\textbf{Coordinate conventions}. We consider camera $I_0^{\text{ground}}$ to define the origin of the world coordinates, \ie\ the identity pose, as well as the spatial location of the BEV satellite image $I^{\text{sat}}$. The BEV $I^{\text{sat}}$ frame is aligned with the heading of $I_0^{\text{ground}}$ such that the camera look-at direction $z_c$ is pointing up in the satellite view. Gaussian splats $\mathcal{G}$ are projected via perspective projection to ground views, and orthographic projection into BEV views. Other ground level images $I_{i>0}^\text{ground}$ are expressed relative to $I_0^{\text{ground}}$.
    %\mt{TODO: 1) illustrate the Splats are rendered via perspective projection to ground views but orthographic projection to BEV views. 2) Add $\mathcal{G^\text{sat}}$ and $\mathcal{G^\text{ground}}$ into the visual, such that the reader understands where ground and satellite gaussians come from. 3) Show world coordinates x,y,z where Z is up (parallel to -y plane).}
    }
    \label{fig:coordinate-conventions}
    \vspace{-1em}
\end{figure}

\boldparagraph{Coordinate conventions.}
We construct a coordinate frame where $I_0^{\text{ground}}$ serves as the reference, with the queried satellite image $I^{\text{sat}}$ centered at this location. We consider $I_0^{\text{ground}}$ as the reference frame and express images $I_{i>0}^\text{ground}$ relative to it. $I_0^{\text{ground}}$ also defines the zero-altitude location. See \cref{fig:coordinate-conventions} for more details. We use $i \in \{0, \ldots, N-1\}$ to index images and $j \in \{0, \ldots, M-1\}$ to index per-pixel attributes.

\subsection{Preliminaries}
\boldparagraph{Geometry transformer.} We seek to encode ground level perspective imagery and orthoimagery into a similar feature space for 3D reconstruction. For ground level images, we adopt the VGGT \cite{wang2025vggt} architecture as a feature extractor. Images $I_i^{\text{ground}} \in \mathbb{R}^{H \times W}$ are encoded using DINOv2 \cite{oquab2023dinov2} to extract a sequence of patch tokens $t_i^{\text{img}} \in \mathbb{R}^{l_i \times d} $ with $d=1024$ being the embedding dimension and $l_i = \frac{H \times W}{p^2}$ for a patch size of $p=14$. A camera token $t_i^\text{cam} \in \mathbb{R}^{1 \times d}$ and four register tokens $t_i^\text{reg} \in \mathbb{R}^{1 \times d}$ \cite{darcet2023vitneedreg} are prepended to each image token sequence. These concatenated token sequences $t_{i}^\text{ground} = (t_i^\text{cam}, t_i^\text{reg}, t_i^\text{img})$ are processed by a $N$-layer alternating-attention ViT transformer \cite{dosovitskiy2021image}. Each layer applies the attention mechanism to either its own image token sequence, referred to as frame attention $\operatorname{Attn}_{\text{frame}}(t_i, t_i)$, or jointly with all image tokens referred to as global attention $\operatorname{Attn}_{\text{global}}\left(\{t_i\}_{i=0}^N, \{t_i\}_{i=0}^N\right)$. Tokens $t_i^\text{ground}$ are trained to regress 6DoF relative camera poses and intrinsics via the camera tokens $t_i^{\text{cam}}$ and dense depth and point maps with DPT \cite{Ranftl2021ICCV} heads.

\boldparagraph{Gaussian splatting.} We express our 3D scene representation as a set $\mathcal{G} = \{ (\bm \mu_j, \bm \Sigma_j, o_j, \bm \theta_j ) \}_j$ of Gaussian distributions $(\bm \mu_j, \bm \Sigma_j)$, with opacities $o_j$ and view-dependent colors $\bm \theta_j \in \mathcal C^{N_{\rm sh}}$ represented via order $N_{\rm sh}=1$ spherical harmonics. We regress Gaussian parameters for both ground and satellite views, represented by $\mathcal{G}^\text{ground}$ and $\mathcal{G}^\text{sat}$ respectively.
% We set the spherical harmonics degree to 1 such that the Gaussian color $\bm \theta_j$ is represented by $N_{\rm sh} = 3 \cdot(\rm deg + 1)^2 = 12$ learnable parameters.
To render Gaussian splats we follow~\cite{kerbl3Dgaussians} and use depth-sorted alpha-blending of colors based on accumulated transmittance at each pixel location.
% Rendering a view from a Gaussian set is done via depth sorted alpha-blending:
% \begin{equation}\label{eq:volume-rendering}
%   {\bf \hat{C}_{\text{3DGS}}} = \sum_{j \in N} {\bf{c}}_{j}\alpha_j T_j, 
% \end{equation}
% \noindent where $\bm c_j$ is the RGB value after resolving for SH coefficients, $T_j = \prod_{k = 1}^{k - 1} (1- \alpha_k)$ is the accumulated transmittance at a pixel location, and $\alpha_j = o_j \cdot \exp{\left(-\frac{1}{2}(\Delta_i)^\intercal \boldsymbol{\Sigma}_j^{-1}(\Delta_i)\right)}$. Similarly, depth renders can be approximated from a Gaussian set using the z-component of view-space means $d_j = \tilde \mu_{j,z}$:
% \begin{align}\label{eq:gaussian-depth}
%   {\hat{D}_{\text{3DGS}}}  & = \frac{\sum_{j \in N} {{d}}_{j}\alpha_j T_j}{\sum_{j \in N} \alpha_j T_j}.
%  \end{align}

\begin{figure*}[!t]
    \centering
    \setlength{\figurewidth}{0.125\textwidth}
    
    \begin{tikzpicture}[
        image/.style = {inner sep=0pt, outer sep=0pt, minimum width=\figurewidth, anchor=north west, text width=\figurewidth},
        node distance = 1.5pt and 1.5pt,
        every node/.style={font={\tiny}\strut},
        label/.style = {font={\footnotesize\bf\vphantom{p}}, anchor=south, inner sep=0pt}
    ]
    %% Row 1
    \node [image] (stacked_inputs_1) {
        \begin{minipage}[t]{\figurewidth}
            \centering
            \includegraphics[height=0.5\figurewidth, width=\figurewidth, keepaspectratio]{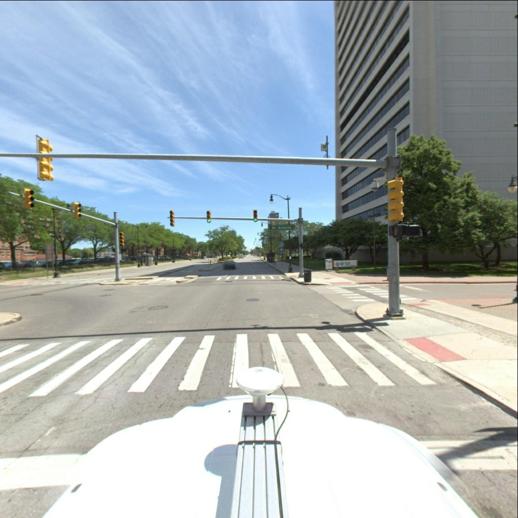}
            \\[0pt]
            \includegraphics[height=0.5\figurewidth, width=\figurewidth, keepaspectratio]{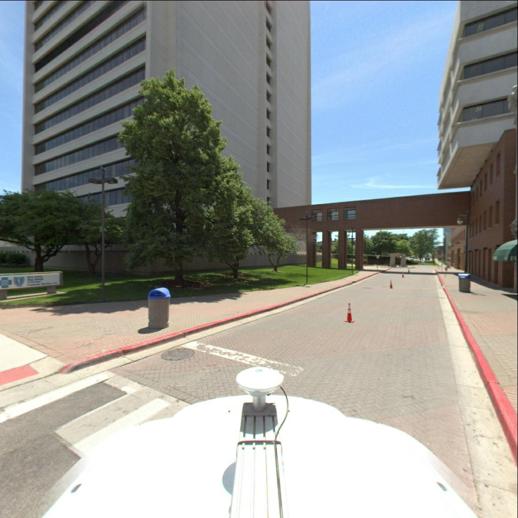}
        \end{minipage}
    };
    \node [image, right=-0.3cm of stacked_inputs_1] (img-00) {\includegraphics[width=\figurewidth]{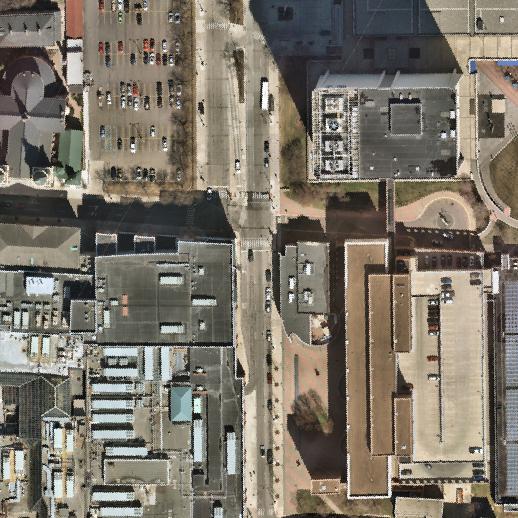}};
    \node [image, right=0.35cm of img-00] (img-01){\includegraphics[width=\figurewidth]{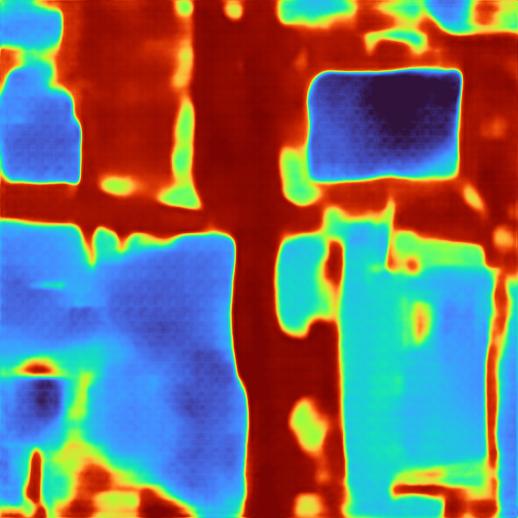}};
    \node [image, right=0.02cm of img-01] (img-02){\includegraphics[width=\figurewidth]{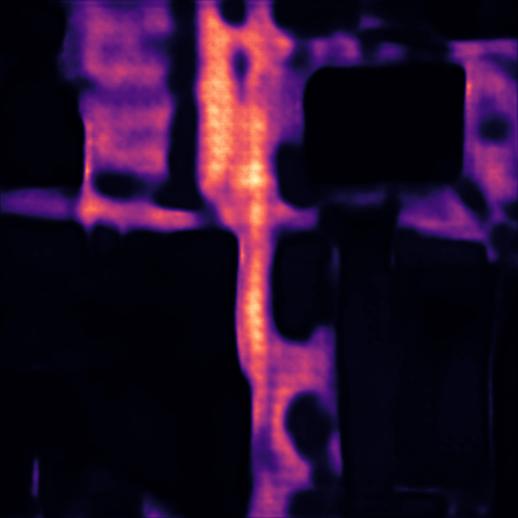}};
    \node [image, right=0.02cm of img-02] (img-03){\includegraphics[width=\figurewidth]{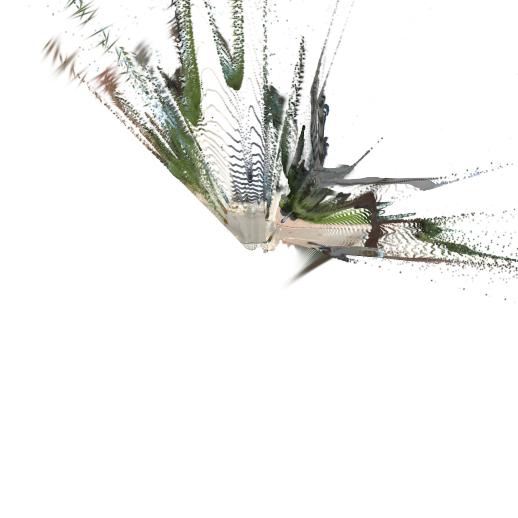}};
    \node [image, right=0.02cm of img-03] (img-04)
    {\includegraphics[width=\figurewidth]{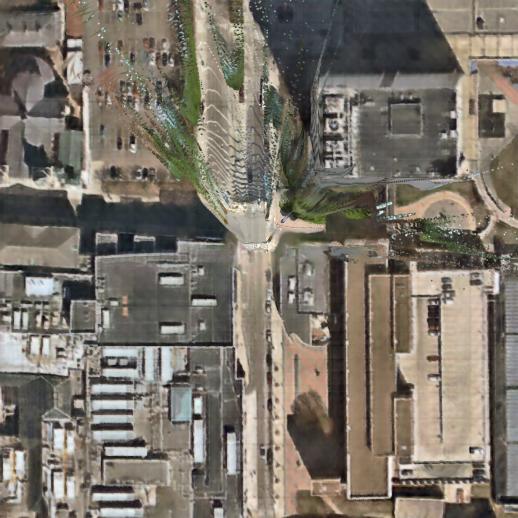}};
    \node [image, right=0.12cm of img-04] (img-05)
    {\includegraphics[width=\figurewidth]{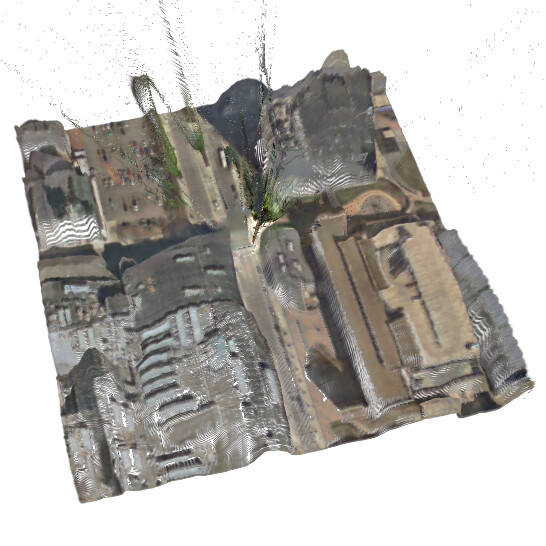}};

    %% Row 2
    \node [image, below=0.2cm of stacked_inputs_1] (stacked_inputs_2) {
        \begin{minipage}[t]{\figurewidth}
            \centering
            \includegraphics[height=0.5\figurewidth, width=\figurewidth, keepaspectratio]{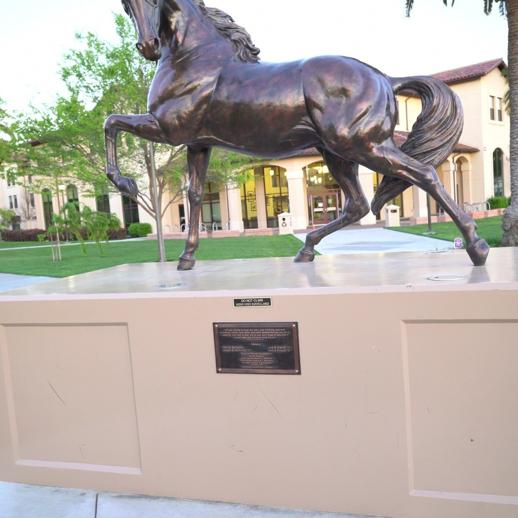}
            \\[0pt]
            \includegraphics[height=0.5\figurewidth, width=\figurewidth, keepaspectratio]{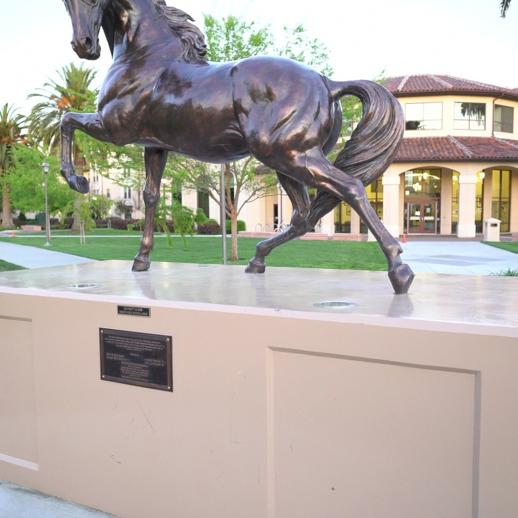}
        \end{minipage}
    };
    \node [image, right=-0.3cm of stacked_inputs_2] (img-10) {\includegraphics[width=\figurewidth]{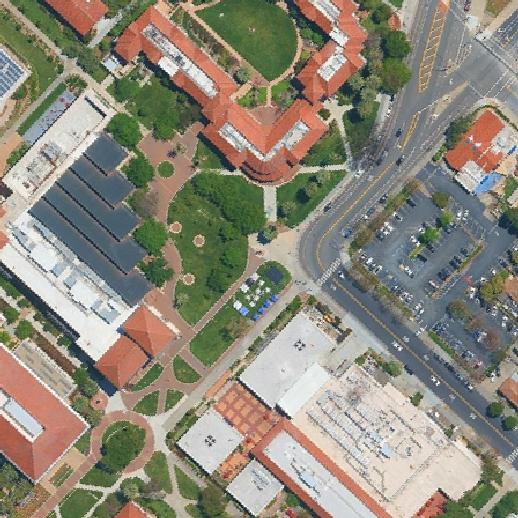}};
    \node [image, right=0.35cm of img-10] (img-11){\includegraphics[width=\figurewidth]{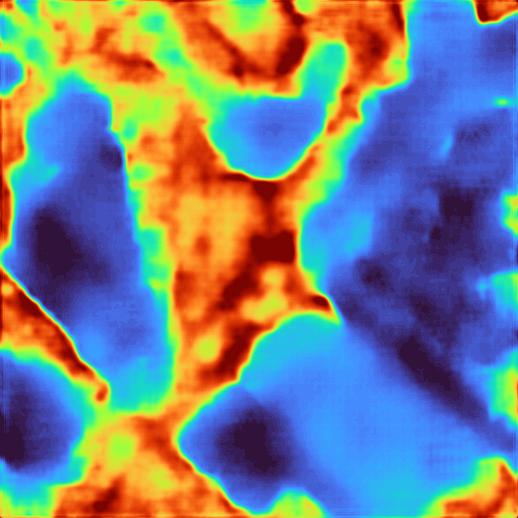}};
    \node [image, right=0.02cm of img-11] (img-12){\includegraphics[width=\figurewidth]{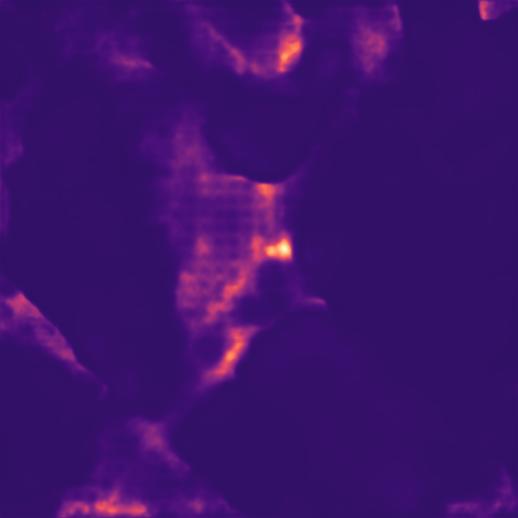}};
    \node [image, right=0.02cm of img-12] (img-13){\includegraphics[width=\figurewidth]{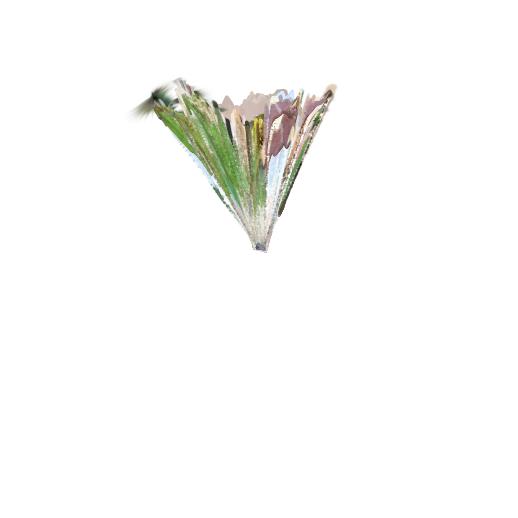}};
    \node [image, right=0.02cm of img-13] (img-14)
    {\includegraphics[width=\figurewidth]{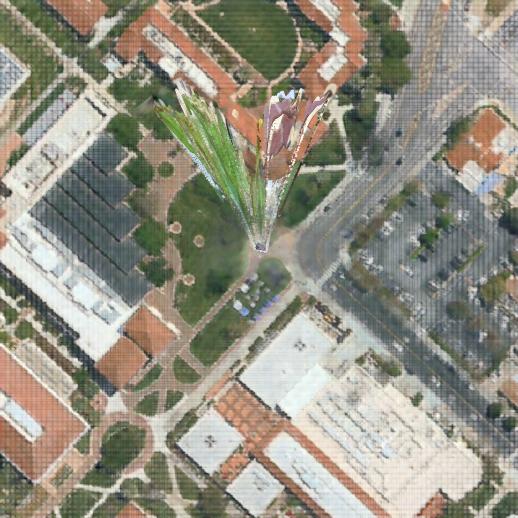}};
    \node [image, right=0.12cm of img-14] (img-15)
    {\includegraphics[width=\figurewidth]{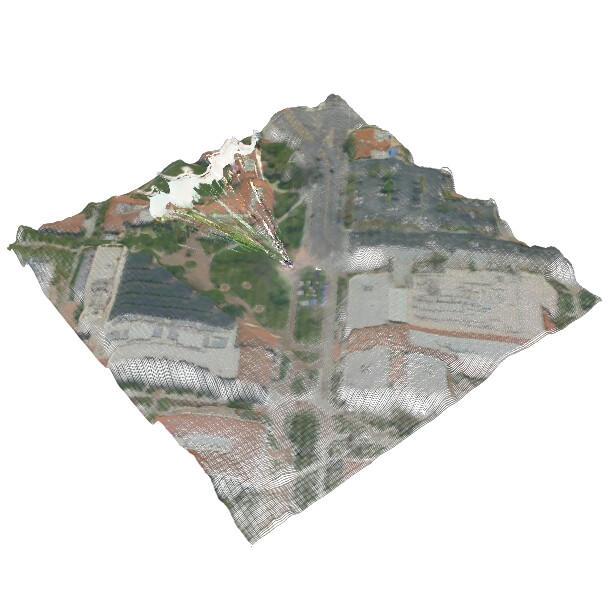}};

    % % % Labels
    \node[label, yshift=3pt, xshift=0pt] at ($(stacked_inputs_1.north west)!0.5!(stacked_inputs_1.north east)$) {Input Imgs};
    \node[label, yshift=2pt] at (img-00.north) {Input Sat};
    \node[label, yshift=2pt] at (img-01.north) {$h^{\text{sat}}$};
    \node[label, yshift=2pt] at (img-02.north) {$C^{\text{sat}}$};
    \node[label, yshift=2pt] at (img-03.north) {$\mathcal{G}^{\text{ground}}$};
    \node[label, yshift=2pt] at ($(img-04.north)!0.5!(img-05.north)$) {$\mathcal{G}^{\text{combined}}$};

    % scene labels
    \node[label, rotate=90, xshift=0pt, yshift=-12pt] at (stacked_inputs_1.west) {\itshape Metropolis \cite{MapillaryMetropolis}};
    \node[label, rotate=90, xshift=0pt, yshift=-12pt] at (stacked_inputs_2.west) {\itshape T \& T \cite{Knapitsch2017}};

    % arrows
    \draw[-{Latex[length=2mm]}, thick] ([xshift=2pt]img-00.east) -- ([xshift=-2pt]img-01.west);
    \draw[-{Latex[length=2mm]}, thick] ([xshift=2pt]img-10.east) -- ([xshift=-2pt]img-11.west);
    
    \end{tikzpicture}
    \vspace{-1em}
    \caption{\textbf{Example reconstruction outputs} on scenes not seen during training. Left to right: input ground images, input satellite image, predicted height map, predicted height confidence (black: low, red: high), predicted ground Gaussians, predicted combined Gaussians.}
    \label{fig:model_outputs}
    %\vspace{-1em}
\end{figure*}
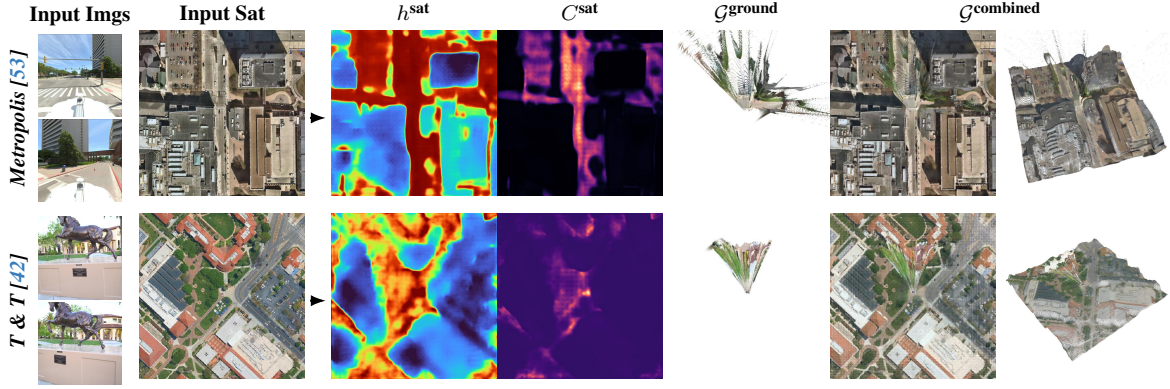

\subsection{Joint Ground and Satellite Aware Transformer}
The direct use of VGGT for orthorectified imagery is problematic: projections are not perspective, so they lack 6DoF poses and camera intrinsics and, more fundamentally, depth estimation is not usable without altitude information. Instead, we consider the transformation between ground and satellite views as 3DoF.
% Specifically, we assume that the reference frame $I_0^\text{ground}$ and orthoimage $I^\text{sat}$ are related by a planar translation and yaw about the orthorectified principal axis.
We assume this transformation is known such that alignment as described in \cref{fig:coordinate-conventions} holds. Similarly, we assume that the reference image $I_0^\text{ground}$ defines the zero-altitude location. We thus formulate BEV geometry prediction as a height-map regression problem
% , disregarding perspective camera models by 
and estimating heights relative to the $I_0^\text{ground}$ frame.

To achieve joint predictions with ground and satellite imagery, we inject cross-attention layers into the alternating-attention backbone of \cite{wang2025vggt}. Specifically, we encode a BEV image $I^{\text{sat}} \in \mathbb{R}^{H_{\text{sat}}\times W_{\text{sat}}}$ as patch tokens $t^{\text{sat}} \in \mathbb{R}^{\frac{H_{\text{sat}} \times W_{\text{sat}}}{p^2} \times d}$ and introduce a 
bidirectional attention mechanism between ground tokens $t^{\text{ground}}_i$ and $t^{\text{sat}}$, referred to as $\operatorname{Attn}_{\text{meta}}$. This consists of two residual cross-attention layers
% , where the only difference is in interchanging the key, query, and value pairs with either
that mix signals from satellite tokens $t^{\text{sat}}$ and ground tokens $t^{\text{ground}}$, so 
% Formally, we define it as follows:
\begin{equation}
\operatorname{Attn}_{\text{meta}}(t^{\text{sat}},t^{\text{ground}}) = \mathcal{A}_2(t^{\text{sat}},\mathcal{A}_1(t^{\text{ground}}, t^{\text{sat}}, t^{\text{sat}})),
\end{equation}
where $\mathcal{A}(Q, K, V)$
% = \mathrm{softmax}\!\left(\frac{Q K^\top}{\sqrt{d}}\right) V$
is the multi-head attention mechanism \cite{dosovitskiy2021image, vaswani2017attention}. $\operatorname{Attn}_{\text{meta}}$ is performed $L=12$ times.
% This design allows for the exchange of information between satellite and ground image tokens, enabling both token sets to be updated. 
% We hypothesize that such interaction aligns satellite patch tokens with VGGT’s expressive feature space.

%\todo{Mention almost 90 degree shift in pose. Vastly different resolutions and texture.}

\subsection{Geometry and Gaussian Splats Prediction}
\label{sec:geometry_and_gaussian_splats_prediction}
% \todo{explain rgb image shortcut in 3DGS DPT heads.}

\boldparagraph{Ground views.} Ground tokens $t^{\text{ground}}_i$ flow through N-layers of alternating % frame 
$\operatorname{Attn}_{\text{frame}}$
and % global
$\operatorname{Attn}_{\text{global}}$
attention as well as $L$-layers of  $\operatorname{Attn}_{\text{meta}}$. Tokens $t^{\text{ground}}_i$ are used to predict ground level depth maps and confidences with a DPT head:
\begin{equation}
    d_j^{\text{ground}}, C_j^{\text{ground}} = \operatorname{DPT_{depth}}(t^{\text{ground}}_i).
\end{equation}
Additionally, a camera-head is used to regress the 6DoF relative pose $\bm T_i$ and perspective camera intrinsics $\bm K_i$ for each input ground image using the camera tokens $t_i^\text{cam}$, which are used to define the means of ground Gaussians $\bm \mu_j^\text{ground} = \operatorname{backproject}(d^{\text{ground}}_j, \hat{\bm K_j}, \hat{\bm T_j})$. The remaining ground Gaussian parameters are predicted with a DPT head  $\Sigma_j^{\text{ground}}, o_j^{\text{ground}}, \theta_j^{\text{ground}}  = \operatorname{DPT_{3DGS}}(t^{\text{ground}}_i)$ that also has a skip connection from the input images to allow color information to propagate to the outputs~\cite{jiang2025anysplat}.

\boldparagraph{Satellite views.} Satellite tokens $t^\text{sat}$ % exchange information with ground level views within the L-layers of bidirectional cross-attention $\operatorname{Attn}_{\text{meta}}$. We then 
are used to regress a height map relative to $I_0^\text{ground}$ and per-pixel confidence values $C^{\text{sat}}$, the latter of which is crucial for mitigating the impact of inaccurate or noisy ground-truth terrain height data:
\begin{equation}\label{eq:height_reg}
    h^{\text{sat}}, C^{\text{sat}} = \operatorname{DPT_{height}}(t^{\text{sat}}).
\end{equation}
Height map $h^\text{sat}$ is transformed into 3D Gaussian locations $\mu_j^{\text{sat}}$ utilizing the known spatial resolution $r^{\text{sat}}$ of the satellite image, so
\begin{equation}
    \bm \mu_j^{\text{sat}} = \begin{pmatrix} \mu_x & \mu_y & \mu_z \end{pmatrix}^\top = \begin{pmatrix} \frac{u}{r^{\text{sat}}} & \frac{v}{r^{\text{sat}}} & h^{\text{sat}}(u, v) \end{pmatrix}^\top,
\end{equation}
where $u, v$ are pixel locations in $I^{\text{sat}}$. This conversion directly assumes an orthographic projection model for the satellite image. Similarly, the remaining satellite Gaussian $\mathcal{G^\text{sat}}$ attributes are regressed with a DPT head.

\boldparagraph{Scene scale normalization.}
% The choice of scene normalization for optimal training supervision is a critical design choice. Approaches vary, including scaling by ground truth depths \cite{dust3r_cvpr24, wang2025vggt, wang2025moge} or camera baselines \cite{ye2024noposplat}, or adopting a fully metric coordinate system \cite{duisterhof2025mastrsfm, wang2025moge2}.
% We adopt the per-batch $\ell_2$-scaling derived from backprojected depths, as established in \cite{dust3r_cvpr24,wang2025vggt}, to normalize depths and poses of ground imagery. 
We adopt the per-batch $\ell_2$-scaling of the scene derived from backprojected depths, as established in \cite{dust3r_cvpr24,wang2025vggt}, to normalize depths, poses of ground imagery, and we also integrate $\mathbf{h}^{\text{sat}}$ and $r^{\text{sat}}$ into the normalization scheme. See the Supplemental for further details. Notably, at inference time, all network predictions are expressed in the same normalized coordinate frame.

% The scaling impacts the regressed height maps $\mathbf{h}^{\text{sat}}$ relative to $I_0^{\text{ground}}$ from orthoimages with a known spatial resolution $r^{\text{sat}}$ (expressed in pixels per meter). The spatial resolution $r^{\text{sat}}$ is used to map between satellite pixel space and world coordinates. Although we regress a single per-pixel scalar value for height maps in \cref{eq:height_reg}, its spatial consistency with the ground level depth and camera poses is paramount. Therefore, we integrate $\mathbf{h}^{\text{sat}}$ and $r^{\text{sat}}$ into the same normalization scheme. Specifically, we compute a scalar value $s$ with:
% \begin{equation}
% s = \frac{1}{M} \sum_{j=1}^{M} \| \bm \mu_j \|_2, \quad \text{where } \bm \mu_j = \operatorname{backproject}(d_j, \bm K_j, \bm T_j).
% \end{equation}
% $s$ is then used to normalize all metric quantities during training: camera poses,  depth maps, height maps, and the satellite spatial resolution factor: $
% \hat{\bm T} = \frac{\bm T}{s}, \quad
% \hat{d} = \frac{d}{s}, \quad
% \hat{\bm h}^{\text{sat}} = \frac{\bm h^{\text{sat}}}{s}, \quad
% \hat{r}^{\text{sat}} = s \cdot r^{\text{sat}}.
% $ We then train our network to regress values in this normalized space. Notably, at inference time, all network predictions are expressed in the same normalized coordinate frame.

\subsection{Losses}
\boldparagraph{Ground views.} We supervise our ground Gaussians $\mathcal{G}^\text{ground} = \{ (\bm \mu_j, \bm \Sigma_j, o_j, \bm \theta_j ) \}_j$ with the following losses. $\bm \mu_j^{\text{ground}}$ are regularized implicitly via depths $d_j^{\text{ground}}$ with a confidence weighted depth loss
\begin{equation}
\mathcal{L}_{\text{depth}} = \sum_{j=1}^{M} \left( \left\| \hat{d}^{\text{ground}}_{j} - d^{\text{ground}}_{j} \right\|_2 - \alpha \log C_{j} \right).
\end{equation}
During training, we utilize ground truth camera parameters, whereas at inference time we utilize predicted parameters. Camera poses and intrinsics are regularized with an L1 loss: $\mathcal{L}_{\text{cam}} = \|\hat{\bm T} - \bm T \|_1 + \|\hat{\bm K} - \bm K \|_1$. $\bm \Sigma_j$, i.e. the Gaussian sizes, are regularized with a depth consistency loss between predicted depths $\bm d_j^{\text{ground}}$ and Gaussian rendered depths, so %, utilizing \ref{eq:gaussian-depth}:
\begin{equation}
    \mathcal{L}_{\text{const}} = \sum_{i=1}^{M} \left( \left\|d_j^{\text{ground}} - \hat{D}_{j,\text{3DGS}}^{\text{ground}} \right\| \right).
\end{equation}
The Gaussian splat colors $\theta_j^{\text{ground}}$ and opacities $o_j^{\text{ground}}$ are regularized with a mean-squared error and perceptual error \cite{zhang2018perceptual} between ground-truth input images and Gaussian splats rendered from input image views, so this loss is
\begin{equation}\label{eq:rgb_ground}
    \mathcal{L}_\text{RGB}^{\text{ground}} = \sum_{i=1}^{N} \left( \|I_i - C^{\text{ground}}_{i,\text{3DGS}} \| + \gamma \cdot \operatorname{LPIPS}(I_i, C^{\text{ground}}_{i,\text{3DGS}}) \right).
\end{equation}

\boldparagraph{Satellite views.} In a similar fashion, we regularize height predictions $h^{\text{sat}}$ with a confidence weighted loss: $\mathcal{L}_{\text{height}} =\left\| \hat{h}^{\text{sat}} - h^{\text{sat}} \right\|_2 - \alpha \log C^{\text{sat}}.$ Differently from ground views, for color supervision, we render satellite Gaussians to both input views as well as novel views, so 
\begin{equation}\label{eq:rgb_sat}
    \mathcal{L}_{\text{RGB}}^{\text{sat}} = \sum_{i=1}^{N} \|I_i - C^{\text{sat}}_{i,\text{3DGS}} \| + \sum_{k=1}^{K}(\|I_{k,\text{nvs}} - C^{\text{sat}}_{k,\text{3DGS}} \|),
\end{equation}
\noindent where $I_{\text{nvs}}$ are selected as interpolating views between the bounds of the input images $N$.

\boldparagraph{Combined ground and BEV views.} Lastly, we promote the synergy between the two Gaussian sets $\mathcal{G}^{\text{ground}}$ and $\mathcal{G}^{\text{sat}}$ by computing a loss on $\mathcal{G}^{\text{combined}} = \mathcal{G}^{\text{ground}} \cup \mathcal{G}^{\text{sat}}$, namely 
\begin{equation}\label{eq:rgb_combined}
    \mathcal{L}_\text{RGB}^{\text{combined}} = \sum_{i=1}^{N} \|I_i - C^{\text{combined}}_{i,\text{3DGS}} \|.
\end{equation}

\boldparagraph{BEV rendering.}
In \cref{eq:rgb_ground} and \cref{eq:rgb_sat}, we regularize $\mathcal{G}^{\text{ground}}$ and $\mathcal{G}^{\text{sat}}$ with ground level views. To introduce a similar BEV constraint, we reproject Gaussian splats via orthographic projection onto the satellite plane. A key difference lies in the Gaussian size $\bm \Sigma_j$: while perspective Gaussian splatting scales $\bm \Sigma_j$ proportionally to the inverse $z$-depth (distance between camera $\mathbf{T}_i$ and Gaussian mean $\bm {\mu}_j$), orthographic projection omits this depth-dependent scaling. Therefore, when transforming $\mathcal{G}^{\text{ground}}$ and $\mathcal{G}^{\text{sat}}$ into the BEV perspective, we directly project normalized world coordinates into satellite pixel space using the spatial resolution $r^{\text{sat}}$ factor of the satellite image. % and alpha-blend using \cref{eq:volume-rendering}.
We compute a combined Gaussian rendering loss in the satellite view, so 
\begin{equation}\label{eq:l_bev}
    \mathcal{L}_{BEV} = \|I^{\text{sat}} - C^{\text{combined}}_{\text{3DGS}} \|,
\end{equation}
\noindent where $C^{\text{combined}}_{\text{3DGS}}$ denotes renders from combined Gaussians $\mathcal{G}^{\text{combined}} = \mathcal{G}^{\text{ground}} \cup \mathcal{G}^{\text{sat}}$.

\boldparagraph{Sky regularization.}
Sky pixels lack reliable depth cues and often produce erroneous depth estimates. To handle such regions, we identify sky pixels using an off-the-shelf segmentation model~\cite{skyseg1,skyseg2,skyseg3} and penalize implausibly close depth estimates using
\begin{equation}
    \mathcal{L}_{\text{sky\_depth}} = \sum_{j=0}^M \bm M_j \cdot \operatorname{ReLU}(\tau - d_j^{\text{ground}}),
\end{equation}
\noindent where $\tau$ is a threshold distance and $$\bm {M}_{j} = \begin{cases} 1 & \text{if sky} \\ 0 & \text{otherwise.} \end{cases}$$
We also observe that explicitly promoting opaqueness of the sky pixels is necessary, so
\begin{equation}
    \mathcal{L}_{\text{sky\_alpha}} = \sum_{j=0}^M  \bm M_j \cdot \|1 - o_j\|_1.
\end{equation}
We denote the total sky loss with $\mathcal{L}_{\text{sky}} = \mathcal{L}_{\text{sky\_depth}} + \mathcal{L}_{\text{sky\_alpha}}$.

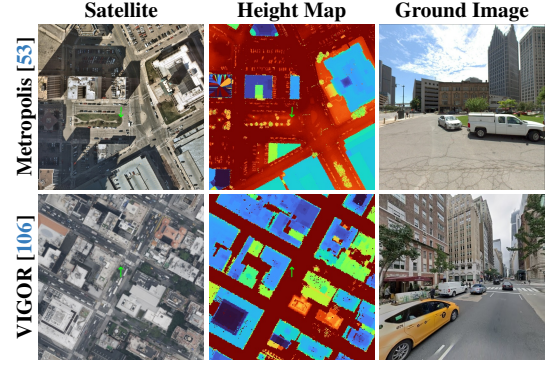
\begin{figure}[!t]
    \centering
    \setlength{\figurewidth}{0.125\textwidth}
    \begin{tikzpicture}[
        image/.style = {inner sep=0pt, outer sep=1pt, minimum width=\figurewidth, anchor=north west, text width=\figurewidth}, 
        node distance = 1pt and 1pt, every node/.style={font= {\tiny}}, 
        label/.style = {font={\footnotesize\bf\vphantom{p}},anchor=south,inner sep=0pt},
    ]

        %% Row 1
        \node [image] (img-00) {\includegraphics[width=\figurewidth]{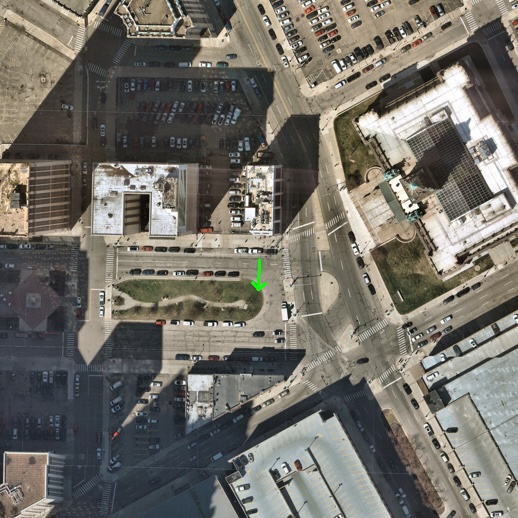}};
        \node [image, right=0.0cm of img-00] (img-01) {\includegraphics[width=\figurewidth]{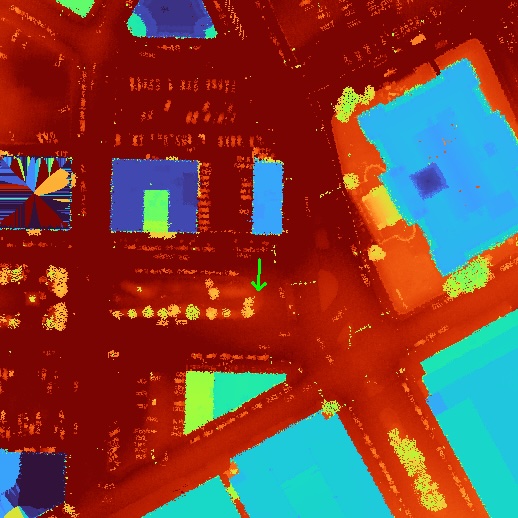}};
        \node [image, right=0.0cm of img-01] (img-02) {\includegraphics[width=\figurewidth]{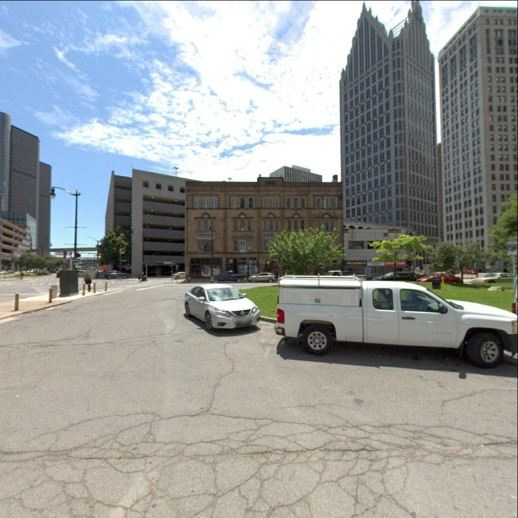}};

        %% Row 2
        \node [image, below=0.0cm of img-00] (img-10) {\includegraphics[width=\figurewidth]{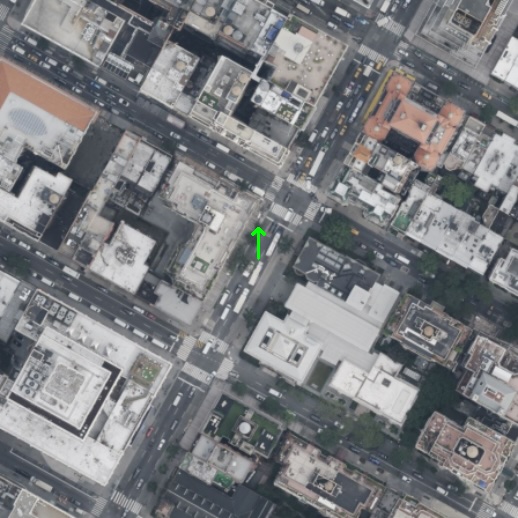}};
        \node [image, right=0.0cm of img-10] (img-11) {\includegraphics[width=\figurewidth]{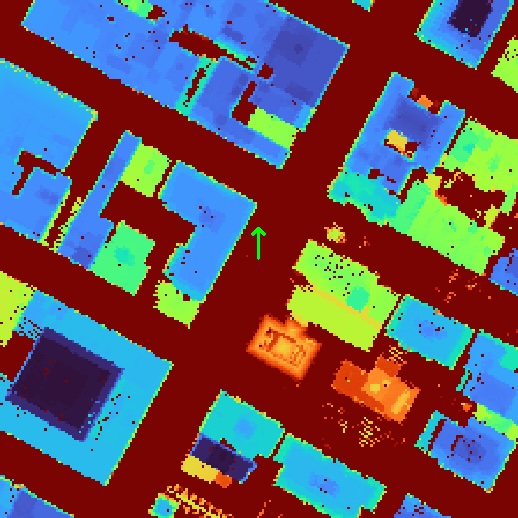}};
        \node [image, right=0.0cm of img-11] (img-12) {\includegraphics[width=\figurewidth]{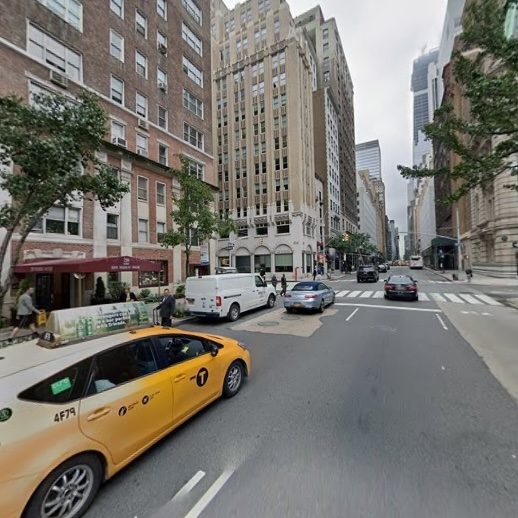
        }};

        %% Row 3
        %\node [image, below=0.0cm of img-10] (img-20) {\includegraphics[width=\figurewidth]{images/dataset_fig/test_holicity_satellite_pose.jpg}};
        %\node [image, right=0.0cm of img-20] (img-21) {\includegraphics[width=\figurewidth]{images/dataset_fig/test_holicity_terrain_pose.jpg}};
        %\node [image, right=0.0cm of img-21] (img-22) {\includegraphics[width=\figurewidth]{images/dataset_fig/test_holicity_image.jpg}
        %};

        % layer labels
        \node[label,rotate=90] (scene1) at (img-00.west) {Metropolis \cite{MapillaryMetropolis}};
        \node[label,rotate=90] (scene2) at (img-10.west) {VIGOR \cite{zhu2021vigor}};
        %\node[label,rotate=90] (scene2) at (img-20.west) {HoliCity \cite{zhou2020holicity}};
        
         % % % Labels
        \node[label] (label1) at (img-00.north) {Satellite};
        \node[label] (label1) at (img-01.north) {Height Map};
        \node[label] (label2) at (img-02.north) {Ground Image};
        
    \end{tikzpicture}
    \vspace*{-.1in}
    \caption{Illustration of training samples from our BEV augmented data. Ground 3DoF pose is denoted with green \textcolor{green}{$\uparrow$} arrow.}
    \label{fig:dataset_figure_main}
    %\vspace*{-1em}
\end{figure}

\begin{table}[!t]
\centering\footnotesize
\caption{\textbf{Our training data} consists of georeferenced outdoor images that include 3DoF pose information, satellite imagery, terrain height maps, and/or ground level depth data. Our driving scenes are sourced from Metropolis and VIGOR. VIGOR contains panoramas, we create perspective cutouts with $90^\circ$ FoV. \textcolor{teal}{$\text{*}$} we generate pseudo-ground truth depths using UniK3D \cite{piccinelli2025unik3d}.
}
\setlength{\tabcolsep}{3pt}
\vspace*{-.1in}
\resizebox{1\columnwidth}{!}{%
\begin{tabular}{lccccc}
\toprule
\textbf{Name} & \textbf{Satellite} & \textbf{Terrain Maps} & \textbf{Depth Maps} & \textbf{\# scenes} & \textbf{Static scenes?} \\
\midrule
Metropolis \cite{MapillaryMetropolis} & \textcolor{teal}{$\checkmark$} & \textcolor{teal}{$\checkmark$} & \textcolor{teal}{$\checkmark$} & 75 & \textcolor{red}{-}\\
%HoliCity \cite{zhou2020holicity} & \textcolor{teal}{$\checkmark$} & \textcolor{teal}{$\checkmark$} & \textcolor{teal}{$\checkmark$} & 6.3K & \textcolor{teal}{$\checkmark$}\\
VIGOR \cite{zhu2021vigor} & \textcolor{teal}{$\checkmark$} & \textcolor{teal}{$\checkmark$} & \textcolor{teal}{$\text{*}$} & 52K & \textcolor{teal}{$\checkmark$} \\
MapFree \cite{arnold2022mapfree} & \textcolor{red}{-} & \textcolor{red}{-} & \textcolor{teal}{$\checkmark$} & 655 & \textcolor{teal}{$\checkmark$} \\
VKITTI2~\cite{cabon2020vkitti2,gaidon2016virtual} & \textcolor{red}{-} & \textcolor{red}{-} & \textcolor{teal}{$\checkmark$} & 21 & \textcolor{red}{-} \\
DL3DV\cite{ling2024dl3dv} & \textcolor{red}{-} & \textcolor{red}{-} & \textcolor{teal}{$\checkmark$} & 10K & \textcolor{teal}{$\checkmark$} \\
\bottomrule
\end{tabular}
}
\vspace{-5pt}
\label{tab:training-datasets}
\end{table}

\subsection{Georeferenced Data Curation}\label{sec:georef_data}
To train our model, we collected a set of geolocalizable ground level datasets, then augment those with BEV satellite images and terrain height data. For orthoimages, we use tiled web map providers Google Maps~\cite{GoogleMaps}, Azure Maps~\cite{AzureMaps}, and Esri World Imagery~\cite{EsriWorldImagery2025}. For access, we  used~\cite{tiledwebmaps1,tiledwebmaps2,tiledwebmaps3} and query views at ground level GPS locations at a sampling density of 2 pixels per meter, at resolution $512 {\times} 512$. For terrain data, we leverage government and public lidar data, namely Geological Survey data~\cite{USGSLidarExplorer} and various data from \cite{FlaiOpenLidarData}. We illustrate training samples in \cref{fig:dataset_figure_main} and give a breakdown of datasets in Table~\ref{tab:training-datasets}. We give more examples and details in the Supplemental. Due to licensing restrictions (\eg Google Maps, Azure Maps) we host only our additional data, and will provide code to query for the satellite images and reproduce the training datasets.

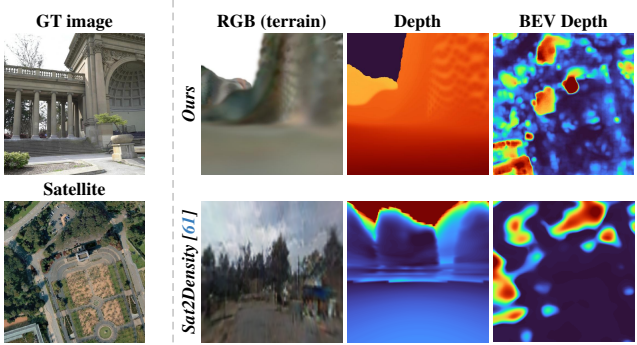
\begin{figure}[!t]
    \centering
    \setlength{\figurewidth}{0.11\textwidth}
    \resizebox{\columnwidth}{!}{%
    \begin{tikzpicture}[
        image/.style = {inner sep=0pt, outer sep=1pt, minimum width=\figurewidth, anchor=north west, text width=\figurewidth}, 
        node distance = 1pt and 1pt, every node/.style={font= {\tiny}}, 
        label/.style = {font={\scriptsize\bfseries\vphantom{p}},anchor=south,inner sep=0pt},
    ]

        %% Column 1 (GT) — SWAPPED ORDER
        \node [image] (input-00) {\includegraphics[width=\figurewidth]{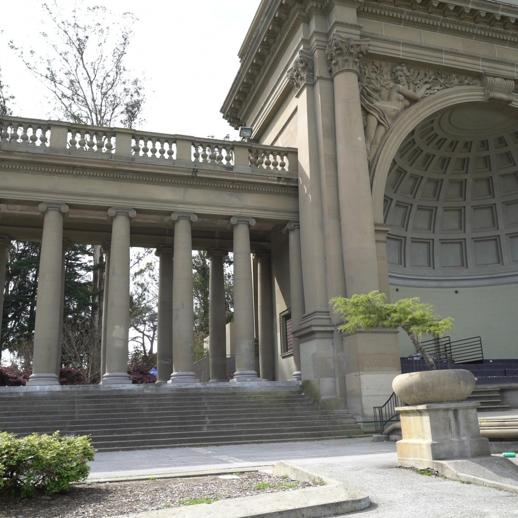}};  % GT on top
        \node [image, below=0.3cm of input-00] (input-01) {\includegraphics[width=\figurewidth]{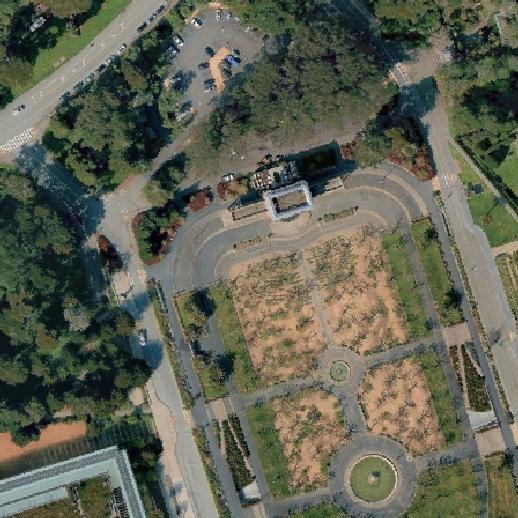}}; % Satellite below

        %% Our method (top row) — extra separation from GT column
        \node [image, right=0.7cm of input-00] (img-01) {\includegraphics[width=\figurewidth]{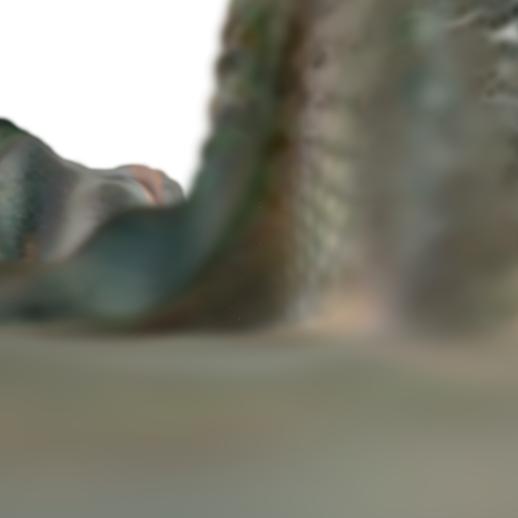}};
        \node [image, right=0.0cm of img-01] (img-02) {\includegraphics[width=\figurewidth]{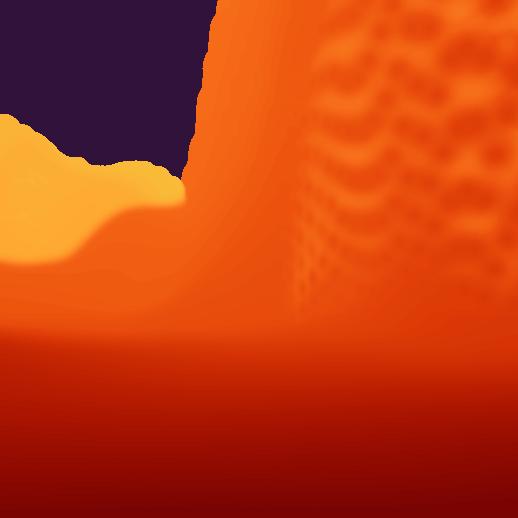}};
        \node [image, right=0.0cm of img-02] (img-03) {\includegraphics[width=\figurewidth]{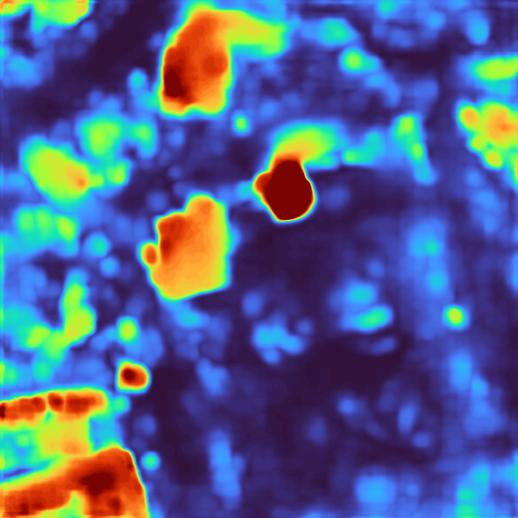}};

        %% Sat2Density (bottom row)
        \node[image, below=0.3cm of img-01] (img-20)
        {\includegraphics[width=\figurewidth]{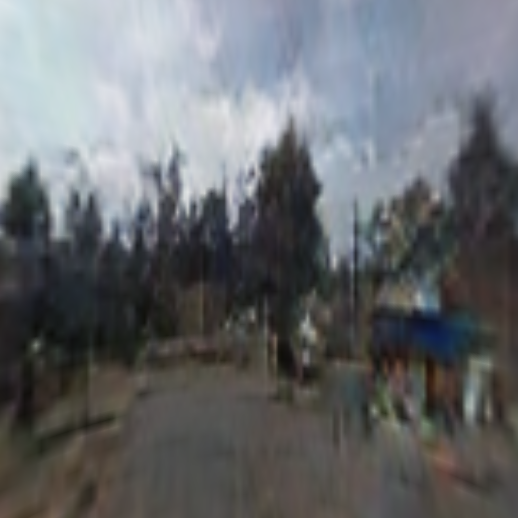}};
        \node [image, right=0.0cm of img-20] (img-21) {\includegraphics[width=\figurewidth]{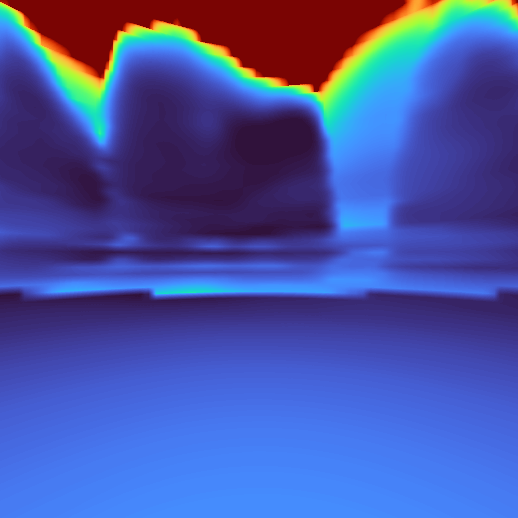}};
        \node [image, right=0.0cm of img-21] (img-22) {\includegraphics[width=\figurewidth,
        viewport=21 21 234 232, clip]{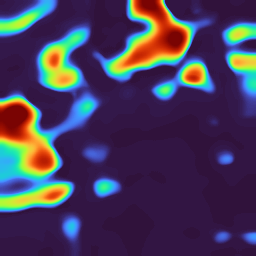}};
        
        %% Labels
        \node[label] (label0) at (input-00.north) {GT image};
        \node[label] (label1) at (input-01.north) {Satellite};
        \node[label] (label2a) at (img-01.north) {RGB (terrain)};
        \node[label] (label2b) at (img-02.north) {Depth};
        \node[label] (label2c) at (img-03.north) {BEV Depth};

        \node[label,rotate=90] (scene1) at (img-01.west) {\itshape Ours};
        \node[label,rotate=90] (scene2) at (img-20.west) {\itshape Sat2Density \cite{sat2density}};

        %% Vertical dashed separator between GT column and predictions
        \coordinate (sep) at ($ (input-00.east)!0.5!(img-01.west) $);
        \draw[densely dashed, gray!70, line width=0.4pt]
            ([yshift=1.2ex]sep |- label0.north) --
            ([yshift=-1.2ex]sep |- img-22.south);

        % (Optional) subtle background tint for the GT column
        %\begin{scope}[on background layer]
        %    \fill[gray!6] ([xshift=-2pt,yshift=2pt]input-00.north west) rectangle
        %                   ([xshift=2pt,yshift=-2pt]input-01.south east);
        %\end{scope}

    \end{tikzpicture}}
    \vspace*{-.3in}
    \caption{\textbf{Satellite-to-ground qualitative results.} \textbf{Column~1:} target ground image and satellite image input. \textbf{Columns~{(2-4)}: Predictions.} \textit{(Top)} our $\mathcal{G}^{\text{sat}}$ rendered to ground-view RGB, ground-view depth, and BEV depth. \textit{(Bottom)} \textbf{Sat2Density}~\cite{sat2density} with RGB and volume-rendered ground/BEV depths from predicted density. Our method produces sharper, more accurate depth maps. The scene is challenging for Sat2Density~\cite{sat2density}, as it was predominantly trained on US country-side images.}
    \vspace{-1em}
    \label{fig:terrain_qualitative}
\end{figure}

%% TANKS AND TEMPLES
\begin{table*}[!htbp]
\centering\scriptsize
\setlength{\tabcolsep}{3pt}
\renewcommand*{\arraystretch}{1}

\caption{\textbf{Outdoor Tanks and Temples sparse-view synthesis.} Metrics are averaged over 10 scenes. For \ours, we report ground-only, terrain-only, and combined (ground+terrain) reconstructions. Methods marked with \textcolor{teal}{{$\text{*}$}} use ground-truth intrinsics. Methods marked with \textcolor{red}{$\text{*}$} require multi-view input and were given one additional adjacent frame during testing. Sat2Density\textsuperscript{\textdagger} takes a single satellite image stylized with one context image (see \cref{fig:terrain_qualitative}).}
\label{tab:tandt_nvs}

\vspace*{-1em}
\setlength{\tabcolsep}{9.3pt} % <- hspacing between columns
\begin{tabular}{llc|ccc|ccc|ccc}
\toprule
& & & \multicolumn{3}{c}{1 context view} & \multicolumn{3}{c}{2 context views} & \multicolumn{3}{c}{3 context views}\\
\textbf{Method} & & GT Pose? & 
\tiny{PSNR$\uparrow$} & \tiny{SSIM$\uparrow$} & \tiny{LPIPS$\downarrow$} & \tiny{PSNR$\uparrow$} & \tiny{SSIM$\uparrow$} & \tiny{LPIPS$\downarrow$} & \tiny{PSNR$\uparrow$} & \tiny{SSIM$\uparrow$} & \tiny{LPIPS$\downarrow$}\\
\midrule
\texttt{Splatfacto} & & \textcolor{teal}{$\checkmark$}  & - & - & -
& 11.53 & 0.2611 & 0.6436
& 11.72 & 0.2888 & 0.6267 \\
\midrule
\texttt{MVSplat} & & \textcolor{teal}{$\checkmark$} 
& - & - & - 
& 6.93 & 0.1252 & 0.6997 
& 7.58 & 0.1631 & 0.6941 \\

\texttt{DepthSplat} & & \textcolor{teal}{$\checkmark$} 
& - & - &- 
&  9.61 & 0.3146 & 0.6077 
& 10.72 & \textit{0.3557} & 0.5873 \\

\texttt{NoPoSplat} & & \textcolor{teal}{$\text{*}$} 
& 6.43\textcolor{red}{$\text{*}$} & 0.1062\textcolor{red}{$\text{*}$} & 0.7040\textcolor{red}{$\text{*}$}
& 8.97 & 0.2197 & 0.6830 
& 8.82 & 0.2359 & 0.6825 \\

% \texttt{FLARE} & & - & & \\
\texttt{Long-LRM} & & - & 8.53 & 0.3392 & 0.7054 & 8.53 & 0.3392 & 0.7054 & 10.54 & 0.3253 & 0.6477\\
\texttt{AnySplat} & & - 
& 7.48 & \textit{0.3572} & \textit{0.6482} 
& \textit{9.85} & 0.3483 & \textbf{0.5773} 
& \textit{10.93} & \textit{0.3775} & \textbf{0.5331} \\
\midrule
\texttt{Sat2Density\textsuperscript{\textdagger}} & & \checkmark 
& \textit{8.81} & 0.3557 & 0.8172 
& 8.90 & \textit{0.3507}& 0.8097 
& 8.85 & 0.3508 & 0.8037 \\
\midrule
\rowcolor{gray!10} \texttt{Ours} & \textit{Combined} & - 
& \textbf{11.13} & \textbf{0.3764} & \textbf{0.6286}
& \textbf{11.67} & \textbf{0.3725} & \textit{0.5984}
& \textbf{12.00} & \textbf{0.3855} & \textit{0.5699} \\

\rowcolor{gray!10} & \textit{Ground} & - 
& 8.92 & 0.3621 & 0.6066
& 9.94 & 0.3615 & 0.5877
& 10.61 & 0.3763 & 0.5631 \\

\rowcolor{gray!10} & \textit{Terrain} & - 
& 8.39 & 0.3783 & 0.6257
& 9.82 & 0.4341 & 0.7474
& 9.63 & 0.4301 & 0.7472\\
\bottomrule
\end{tabular}
\vspace*{-1em}

\end{table*}

%%% DL3DV
\begin{table*}[!htbp]
\centering\scriptsize
\setlength{\tabcolsep}{3pt}
\renewcommand*{\arraystretch}{1}
\caption{\textbf{Outdoor DL3DV sparse-view synthesis results.} Results are averaged over 40 scenes.}
\label{tab:dl3dv_nvs_updated}
\label{tab:dl3dv_nvs}
\vspace*{-1em}
\setlength{\tabcolsep}{9.3pt} % <- hspacing between columns
\begin{tabular}{llc|ccc|ccc|ccc}
\toprule
& & & \multicolumn{3}{c}{1 context view} & \multicolumn{3}{c}{2 context views} & \multicolumn{3}{c}{3 context views} \\
\textbf{Method} & & GT Pose? &  
\tiny{PSNR$\uparrow$} & \tiny{SSIM$\uparrow$} & \tiny{LPIPS$\downarrow$} &  
\tiny{PSNR$\uparrow$} & \tiny{SSIM$\uparrow$} & \tiny{LPIPS$\downarrow$} & \tiny{PSNR$\uparrow$} & \tiny{SSIM$\uparrow$} & \tiny{LPIPS$\downarrow$}\\

%HARDER SPLIT
\midrule
\texttt{Splatfacto} & & \textcolor{teal}{$\checkmark$}  & - & - & -
& 13.46 & 0.2962 & 0.6158
& 13.61 & 0.3018 & 0.6026 \\
\midrule
\texttt{MVSplat } & & \textcolor{teal}{$\checkmark$}
 & - & - & - %& 5.807 & 0.0056 & 0.7378
 & 6.27 & 0.0413 & 0.7174
 & 6.29 & 0.0474 & 0.7158 \\

\texttt{DepthSplat}  & & \textcolor{teal}{$\checkmark$}
 & - & - & - % 8.06 & 0.0900 & 0.6848
 & 8.58 & 0.1569 & 0.6774
 & 9.04 & 0.1817 & 0.6761 \\

\texttt{NoPoSplat}   & & \textcolor{teal}{$\text{*}$}
 & 6.89\textcolor{red}{$\text{*}$} & 0.0669\textcolor{red}{$\text{*}$} & 0.7019\textcolor{red}{$\text{*}$}
 & \textit{11.01} & 0.2670 & 0.6665
 & \textit{11.10} & 0.2731 & 0.6687 \\

\texttt{Long-LRM} & & - 
& 4.78 & \textbf{0.3153} & 0.7196
& 9.74 & 0.2842 & 0.6813
& 10.93 & 0.2890 & 0.6149\\

\texttt{AnySplat}    & & -
 & \textit{8.37} & 0.2639 & \textit{0.6498}
 & 10.37 & \textbf{0.3014} & \textbf{0.5702} 
 & 10.88 & \textit{0.3122} & \textbf{0.5557} \\

\midrule
\rowcolor{gray!10} \texttt{Ours}
 & \textit{Combined} & -
& \textbf{11.33}  & \textit{0.2741} & \textbf{0.6307}
& \textbf{12.10} & \textit{0.2976} & \textit{0.5940}
& \textbf{12.61} & \textbf{0.3204} & \textit{0.5683}\\
 
\rowcolor{gray!10} & \textit{Ground} & - 
& 9.00  & 0.2592  & 0.6191
& 10.05 & 0.2878 & 0.5842
& 10.65 & 0.3103 & 0.5606 \\

\rowcolor{gray!10} & \textit{Terrain} & - 
& 8.24 & 0.2790 & 0.6884
& 8.41 & 0.2801 & 0.6932
& 8.30 & 0.2834 & 0.6928 \\
\bottomrule
\end{tabular}
\vspace*{-1em}
\end{table*}

\begin{figure*}[h]
    \centering
     \includegraphics[width=0.95\textwidth]{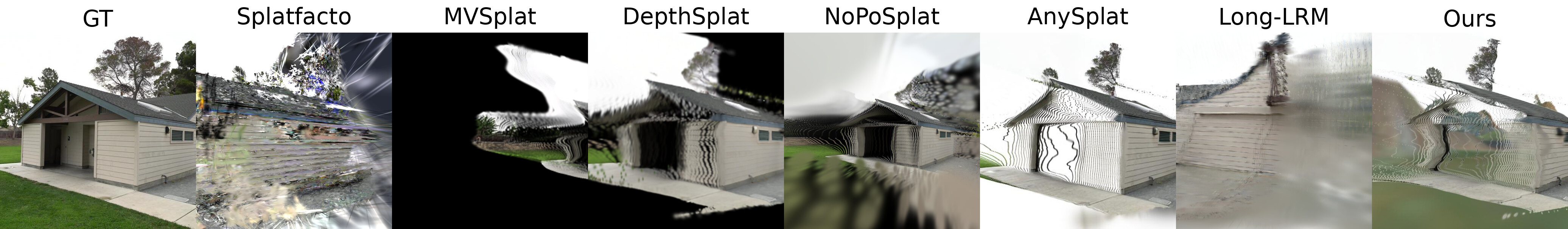}
    \vspace{-0.7em}
    \caption{\textbf{Qualitative results.} We show results for a target image from the Outdoor Tanks and Temples dataset under the sparse-view synthesis setting. Our method extrapolates into unobserved regions by leveraging visual cues from the corresponding satellite image.}
    \label{fig:qual_comparison}
    \vspace{-1.5em}
\end{figure*}

\section{Experiments}
We evaluate \ours on novel view synthesis.

\boldparagraph{Datasets.} We introduce a new task: \textit{view synthesis with geolocalized outdoor images}.
Because no benchmark currently exists for this setting, we construct one by augmenting the well known Tanks and Temples dataset \cite{Knapitsch2017} and selected outdoor scenes from the DL3DV-Benchmark~\cite{ling2024dl3dv}. Specifically, we manually align and scale the COLMAP \cite{colmap} reconstructions of 10 outdoor scenes from Tanks and Temples and 40 scenes from DL3DV to satellite imagery. This alignment process is detailed in the Supplemental. We then evaluate our method and competing baselines in a sparse view setting with varying numbers of input views. % , where we sample the context and target views uniformly in the number of shared 3D feature tracks recovered by COLMAP. % \mt{UPDATE THIS MOHAMED with details how eval split was created}
% We follow \cite{ye2024noposplat} and generate an evaluation set by selecting input images based on their feature overlap ratio, calculated via dense feature matching using ROMA \cite{edstedt2024roma}.
%Performance is analyzed across two distinct difficulty splits: easy split (average overlap score of 57.2\%) and hard split (overlap score of 43.4\%).

\boldparagraph{Training Data.} We train with a combination of ground-view only datasets and those that we augment with satellite imagery, see Table~\ref{tab:training-datasets} and Sec~\ref{sec:georef_data}. Note that for DL3DV\cite{ling2024dl3dv}, we remove benchmark scenes from our training data.

\boldparagraph{Baselines.} We consider various feed-forward novel-view methods. We compare against a) MVSplat \cite{chen2024mvsplat} and b) DepthSplat \cite{xu2024depthsplat} for cost-volume based multi-view 3DGS methods c) AnySplat \cite{jiang2025anysplat}, a feed-forward method initialized by the \cite{wang2025vggt} model weights, d) NoPoSplat \cite{ye2024noposplat}, a similar feed-forward method initialized by \cite{dust3r_cvpr24} weights, 
% e) FLARE \mt{TODO what to say here} \cite{zhang2025flarefeedforwardgeometryappearance} a generalized feed-forward method for 3D regression
and the e) Splatfacto \cite{nerfstudio} method, our baseline that performs per-scene optimization using input views taking 5+ minutes to optimize each scene. We also provide qualitative results comparing satellite-to-ground-only view-synthesis using Sat2Density \cite{sat2density}.

As DL3DV-10K includes DL3DV-Benchmark scenes and NoPoSplat trains on the entire set, in Table~\ref{tab:dl3dv_nvs} we evaluate the RealEstate10K~\cite{zhou2018stereo} variant of NoPoSplat. We do not compare against FLARE~\cite{zhang2025flarefeedforwardgeometryappearance}, as their training data contains scenes of DL3DV-Benchmark and scenes of Tanks and Temples within the Megadepth dataset. For fairness, we center-crop input and target images to square images and rescale to each method's training resolution.

\boldparagraph{View Selection.} First, we compute frame overlaps based on the number of co-visible sparse points recovered by Colmap across every view pair. For context views, we start by selecting the first index in each scene and then we select each subsequent required context view by optimizing for a target overlap. For each combination of context views, we select four target views, each at varying overlap targets to the context views. See the Supplemental for more details.

\boldparagraph{Evaluation metrics.} We report PSNR, SSIM~\cite{ssim}, and LPIPS (VGG-net)~\cite{zhang2018perceptual} metrics.
For our method, we report metrics for ground level Gaussians (\textit{ground}), satellite-to-ground level Gaussians (\textit{terrain}), and combined ground and satellite Gaussians (\textit{combined} (Ours)).
% Alternative names: $\mathcal{G}^{\text{ground}}$, $\mathcal{G}^{\text{satellite}}$, $\mathcal{G}^{\text{combined}}$
%We refer to these as \textit{ground}, \textit{satellite}, and \textit{combined} (Ours) for simplicity.

\boldparagraph{Implementation details.} We use PyTorch \cite{paszke2019pytorch} with \texttt{gsplat} (\texttt{v1.5}) \cite{ye2025gsplat} for Gaussian  rasterization. We initialize our model from AnySplat~\cite{jiang2025anysplat}. We train for 4 days on 2{$\times$}-A100 GPUs with a batch size of 10, using FlashAttention-v2 \cite{dao2022flashattention, dao2023flashattention2} and mixed-precision. % Satellite imagery is queried at 2 pixels per meter resolution from public mapping APIs \cite{EsriWorldImagery2025, GoogleMaps, AzureMaps}.
We resize satellite images and terrain heights such that their spatial extent is 244 meters. Our input resolution is 518{$\times$}518.

\subsection{Results and Ablations}

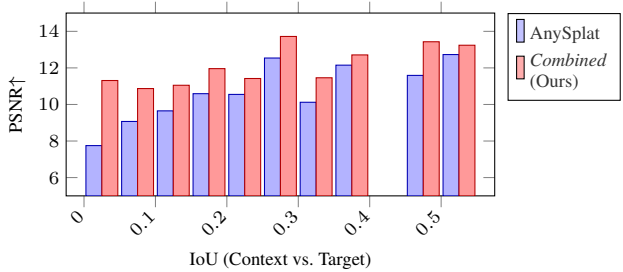
\begin{figure}[t]
    \centering\scriptsize
    \begin{tikzpicture}
        \begin{axis}[
            ybar=0pt,
            bar width=6pt,
            width=0.87\linewidth,
            height=4cm,
            xlabel={IoU (Context vs.\ Target)},
            ylabel={PSNR$\uparrow$},
            ymin=5, ymax=15,
            xtick={0, 0.1, 0.2, 0.3, 0.4, 0.5},
            xticklabel style={
                font=\scriptsize, 
                rotate=45, 
                anchor=north east,
                yshift=4pt,
                /pgf/number format/fixed, 
                /pgf/number format/precision=1
            },
            % 2. Move legend down further to avoid overlap with x-axis labels
            legend style={
                at={(1.03, 1)}, % Position: 3% from left, 95% from bottom
                anchor=north west, 
                legend columns=1,   % Stacked vertically for the side
                font=\scriptsize,
                fill=white,         % Ensures background is opaque
                fill opacity=0.8,   % Slight transparency looks professional
                draw opacity=1,
                cells={anchor=west}, % Align text to the left
                align=left,
            },
            legend image code/.code={
                \draw[#1, draw=black] (0cm,-0.1cm) rectangle (0.1cm,0.2cm);
            },
        ]
            % AnySplat Data
            \addplot[fill=blue!30, draw=blue!70!black] coordinates {
                (0.025, 7.75) (0.075, 9.07) (0.125, 9.65) (0.175, 10.59) 
                (0.225, 10.55) (0.275, 12.54) (0.325, 10.12) (0.375, 12.15) 
                (0.475, 11.59) (0.525, 12.73)
            };
            % OURS Data
            \addplot[fill=red!40, draw=red!70!black] coordinates {
                (0.025, 11.31) (0.075, 10.87) (0.125, 11.05) (0.175, 11.96) 
                (0.225, 11.42) (0.275, 13.72) (0.325, 11.46) (0.375, 12.71) 
                (0.475, 13.43) (0.525, 13.24)
            };
            \legend{AnySplat, \textit{Combined}\\(Ours)}
        \end{axis}
    \end{tikzpicture}
    \vspace*{-8pt}
    \caption{\textbf{Stratified evaluation}. Bucketed PSNR performance ($5\%$ bins) vs. image overlap on our geolocalized Tanks \& Temples dataset.}
    \label{fig:overlap_histogram}    
    \vspace{-1em}
\end{figure}

We report novel view synthesis results in Tables~\ref{tab:tandt_nvs} and \ref{tab:dl3dv_nvs}. Our method matches or outperforms baselines, with the satellite head (\textit{Combined}) improving scores across all context coverage levels. Although overall PSNR values are low, this reflects the challenging setup—input–target pairs have low overlap (IoU $\approx 0.05$--$0.5$), making many cases difficult. Cost-volume baselines like MVSplat and DepthSplat require multiple inputs and tend to fail under 1-view or small-baseline conditions, where our satellite-conditioned model performs more robustly. Qualitative comparisons are shown in Fig.~\ref{fig:qual_comparison} and Fig.~\ref{fig:terrain_qualitative}.

\begin{figure}[h]
    \centering
    % 1. FIGURE CONTENT 
    \setlength{\figurewidth}{0.11\textwidth}
    \begin{tikzpicture}[
        image/.style = {inner sep=0pt, outer sep=1pt, minimum width=\figurewidth, anchor=north west, text width=\figurewidth}, 
        node distance = 1pt and 1pt, every node/.style={font= {\tiny}}, 
        label/.style = {font={\footnotesize\bf\vphantom{p}},anchor=south,inner sep=0pt},
    ]
        %% Row 1
        \node [image] (img-00) {\includegraphics[width=\figurewidth]{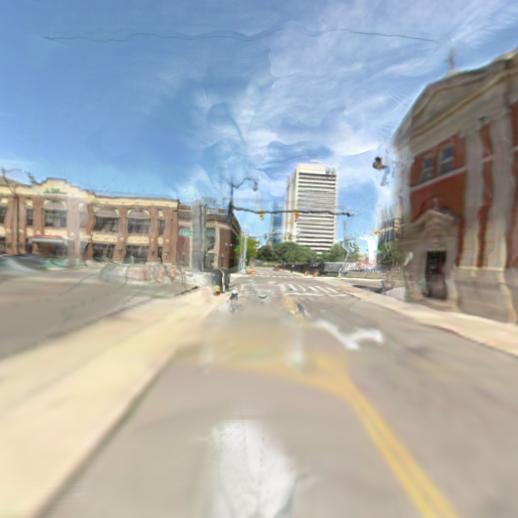}};
        \node [image, right=0.2cm of img-00] (img-01) {\includegraphics[width=\figurewidth]{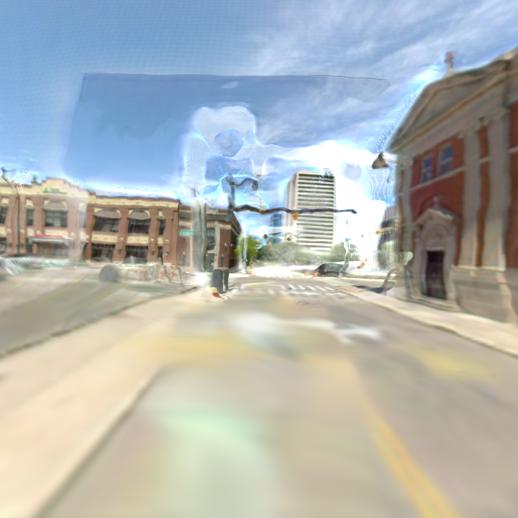}};
        \node [image, right=0.2cm of img-01] (img-02) {\includegraphics[width=\figurewidth]{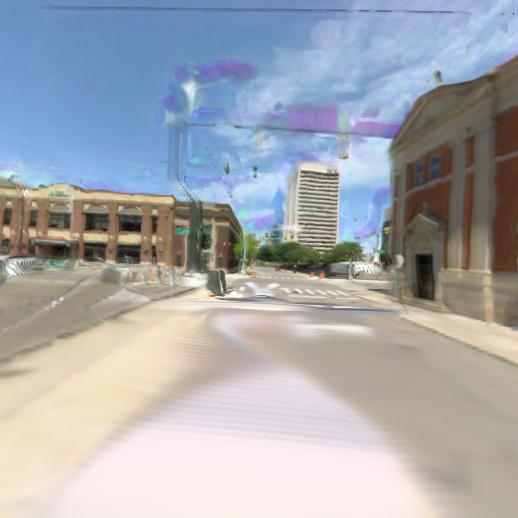}};
        \node [image, right=0.2cm of img-02] (img-03) {\includegraphics[width=\figurewidth]{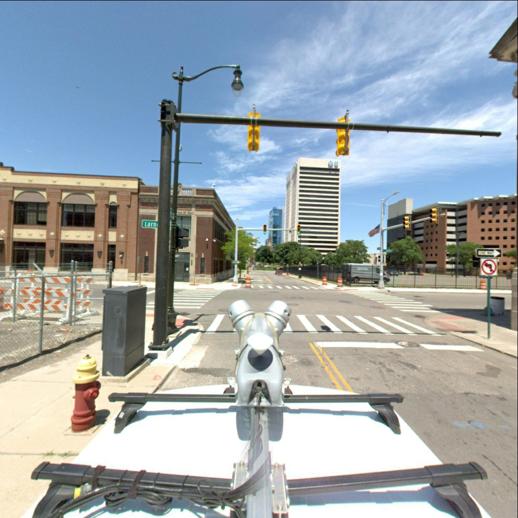}};

        %% Row 2
        \node [image, below=0.27cm of img-00] (img-10) {\includegraphics[width=\figurewidth]{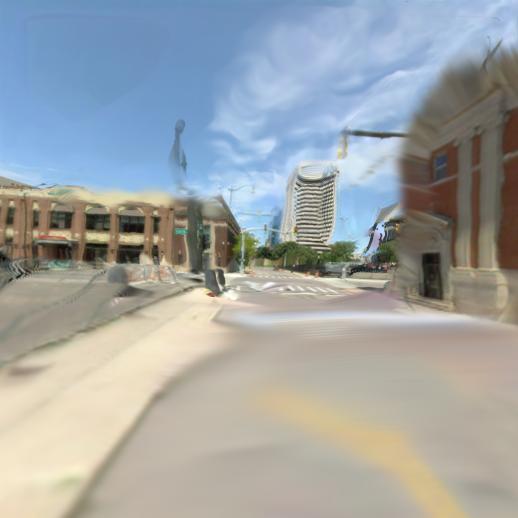}};
        \node [image, right=0.22cm of img-10] (img-11) {\includegraphics[width=\figurewidth]{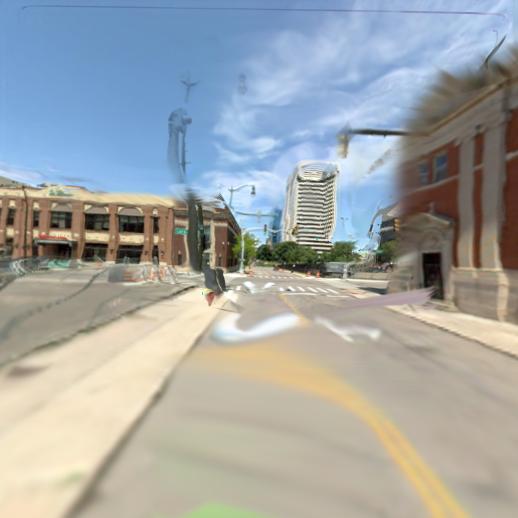}};
        \node [image, right=0.22cm of img-11] (img-12) {\includegraphics[width=\figurewidth]{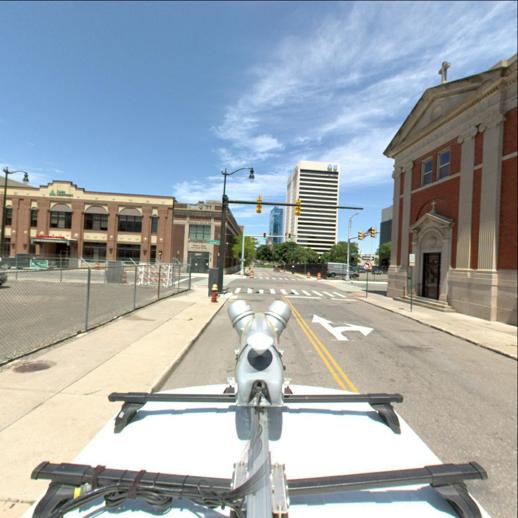}};
        \node [image, right=0.22cm of img-12] (img-13) {\includegraphics[width=\figurewidth]{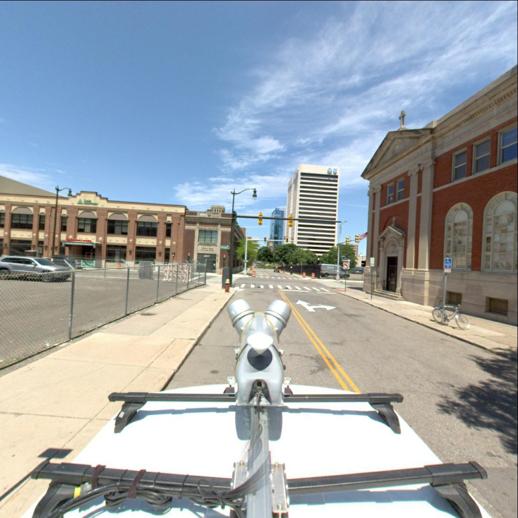}};
        
        % % % Labels
        \node[label] (label0) at (img-00.north) {Baseline};
        \node[label] (label1) at (img-01.north) {w/ $\mathcal{L}_{\text{const}}$};
        \node[label] (label2) at (img-02.north) {w/ $\mathcal{L}_{\text{sky}}$};
        \node[label] (label3) at (img-10.north) { w/ sat};
        \node[label] (label4) at (img-11.north) { w/ all losses};
        \node[label] (label5) at (img-12.north) {GT Image};
        \node[label] (label6) at (img-13.north) {Input Image 2};
        \node[label] (label5) at (img-03.north) {Input Image 1};

    %\node[label, rotate=270] (scene2) at ([xshift=-0.5cm]stacked_inputs.east) {\itshape Input Images};
    
    \end{tikzpicture}
    \vspace*{-1.5em}
    \caption{\textbf{Qualitative ablation} of model with 2 input images.}
    \label{fig:ablation_qualitative}
    
    % *** CRITICAL STEP: Remove space after the first caption ***
    % 2. TABLE CONTENT
    \newcommand{\rowindent}{\hspace{6pt}}
    \centering\scriptsize
    \setlength{\tabcolsep}{3pt}
    \renewcommand*{\arraystretch}{1}
    \captionof{table}{\textbf{Quantitative ablation} on the Metropolis dataset. %Models are trained for 50k iterations on an A100 GPU, lasting $\sim$12h. 
    Metrics are averaged over 36 test scenes, covering 2 input and 2 interpolated novel-views. Our satellite variant achieves superior results due to its larger BEV coverage, allowing for extrapolation into occluded and unseen regions in large baseline driving scenes.}
    \label{tab:ablations}
    \begin{tabular}{l|ccc}
    \toprule
    \textbf{Method} & \textbf{Ground} & \textbf{Terrain} & \textbf{Combined} \\
    & PSNR & PSNR & PSNR \\
    \midrule
    \multicolumn{4}{l}{VGGT w/ 3DGS:} \\
    \rowindent $\mathcal{L}_{\text{cam}} + \mathcal{L}_{\text{depth}} + \mathcal{L}_\text{RGB}^{\text{ground}}$ & 15.26 & - & - \\
    \rowindent $+\mathcal{L}_{\text{const}}$ & 16.99 & - & - \\
    \rowindent $+\mathcal{L}_{\text{sky}}$ & 17.10 & - & -\\
    \midrule
    \multicolumn{4}{l}{VGGT w/ 3DGS w/ SAT:} \\
    \rowindent $\mathcal{L}_{\text{cam}} + \mathcal{L}_{\text{depth}} +\mathcal{L}_{\text{const}} +\mathcal{L}_{\text{sky}} + \mathcal{L}_\text{RGB}^{\text{combined}}$ & 16.99 & 5.24 & 17.17 \\
    \rowindent $+ \mathcal{L}_\text{RGB}^{\text{ground}}$ & 16.61 & 5.36 & 16.87 \\
    \rowcolor{gray!10} \rowindent $+ \mathcal{L}_\text{RGB}^{\text{sat}}$ & 17.59 & 12.25 & 18.63 \\
    \bottomrule
    \end{tabular}
    \vspace*{-1em}
\end{figure}

In \cref{fig:overlap_histogram}, we stratify Tanks and Temples by context--target IoU (co-visible COLMAP features at $5\%$ bins) and plot PSNR. The satellite model outperforms the baseline across bins, with the largest gains at low overlap ($\leq 0.15\,\mathrm{IoU}$), indicating stronger extrapolation to novel views. Ablations in \cref{tab:ablations} and \cref{fig:ablation_qualitative} show that consistency and sky regularization improve VGGT w/ 3DGS, while VGGT w/ 3DGS w/ SAT performs best, especially with $\mathcal{L}_\text{RGB}^{\text{sat}}$, likely due to better BEV coverage of occluded and unseen regions.

%% ABLATIONS

\section{Conclusions}
We introduced \ours, a novel feed-forward method that predicts Gaussian splats for ground \textit{and} satellite images. Unlike previous approaches, our BEV Gaussians increase scene coverage and improves view-synthesis in difficult outdoor scenes. To train our model with the new cross-view setting, we curated a collection of existing datasets and supplemented them with satellite views and height maps. We also introduce a new benchmark for novel-view synthesis with the aid of satellite images using public outdoor scenes of Tanks and Temples and DL3DV.

\section{Acknowledgments}
We thank Zawar Qureshi and Jakub Powierza for compute infrastructure support and Alan Paul for help in generating terrain data. MT, JK, and AS acknowledge funding from the Research Council of Finland (362408, 339730).

% \boldparagraph{Limitations.}

\clearpage

%%%%%%%%% REFERENCES
{\small
\bibliographystyle{ieee_fullname}
\bibliography{egbib}
}
\clearpage

% CVPR 2026 Paper Template; see https://github.com/cvpr-org/author-kit

%\documentclass[10pt,twocolumn,letterpaper]{article}
%
%\PassOptionsToPackage{nameinlink}{cleveref}

%%%%%%%%% PAPER TYPE  - PLEASE UPDATE FOR FINAL VERSION
%\usepackage{cvpr}              % To produce the CAMERA-READY version
% \usepackage[review]{cvpr}      % To produce the REVIEW version
% \usepackage[pagenumbers]{cvpr} % To force page numbers, e.g. for an arXiv version

% It is strongly recommended to use hyperref, especially for the review version.
% hyperref with option pagebackref eases the reviewers' job.
% Please disable hyperref *only* if you encounter grave issues, 
% e.g. with the file validation for the camera-ready version.
%
% If you comment hyperref and then uncomment it, you should delete *.aux before re-running LaTeX.
% (Or just hit 'q' on the first LaTeX run, let it finish, and you should be clear).
\definecolor{cvprblue}{rgb}{0.21,0.49,0.74}

\maketitlesupplementary
\appendix

In this supplementary material, we provide additional implementation details omitted from the main paper. We describe our custom benchmark datasets, outline the evaluation protocol, present further experimental analysis of our method, and illustrate more qualitative results of our model’s performance alongside comparisons to baseline methods. We further discuss the limitations of the proposed \ours approach.

\section{More Implementation Details}
\boldparagraph{Training.}
We train our method with an initial learning rate of $1 \times 10^{-4}$ using the AdamW~\cite{kingma2015adam} optimizer (weight decay $0.05$, $\beta_1 = 0.9$, $\beta_2 = 0.95$), and apply a cosine annealing schedule for $70K$ iterations. For training, we initialize with \cite{jiang2025anysplat} weights and freeze ground-level patch embedding, $\operatorname{Attn}_{\text{frame}}$ and $\operatorname{Attn}_{\text{global}}$ layers and fine-tune all output heads which include the camera, depth, and Gaussian prediction heads as well as our $\operatorname{Attn}_{\text{meta}}$ layers. Our total supervision loss is given by:
{\small%
\begin{equation}
\begin{split}
\mathcal{L}_{\text{total}} =\;&
    \lambda_{\text{cam}} \mathcal{L}_{\text{cam}}
  + \lambda_{\text{depth}} \mathcal{L}_{\text{depth}}
  + \lambda_{\text{const}} \mathcal{L}_{\text{const}} \\
  &+ \lambda_{\text{height}} \mathcal{L}_{\text{height}} \\
  &+ \lambda_{\text{ground}} \mathcal{L}_{\text{RGB}}^{\text{ground}}
  + \lambda_{\text{combined}} \mathcal{L}_{\text{RGB}}^{\text{combined}}
  + \lambda_{\text{sat}} \mathcal{L}_{\text{RGB}}^{\text{sat}} \\
  &+ \lambda_{\text{sky}} \mathcal{L}_{\text{sky}}
  + \lambda_{\text{bev}} \mathcal{L}_{\text{BEV}},
\end{split}
\end{equation}
}%

\noindent where $ \mathcal{L}_{\text{sky}} = \mathcal{L}_{\text{sky\_depth}} + \mathcal{L}_{\text{sky\_alpha}}$. 
We set $\lambda_{\text{cam}} = 1.0$,
$\lambda_{\text{depth}} = 1.0$,
$\lambda_{\text{const}} = 1.0$,
$\lambda_{\text{height}} = 1.0$,
$\lambda_{\text{ground}} = 1.0$, 
$\lambda_{\text{combined}} = 1.0$, 
$\lambda_{\text{sat}} = 1.0$,
$\lambda_{\text{sky}} = 0.1$, and
$\lambda_{\text{BEV}} = 0.5$. Note, we utilize ground truth terrain heights \textit{only} during training for our height regression loss $\mathcal{L}_\text{height}$. At inference time, we utilize satellite RGB images.

\boldparagraph{Gaussian rendering.} For \ours, we are able to render images from ground Gaussians $\mathcal{G}^{\text{ground}}$, satellite Gaussians $\mathcal{G}^{\text{sat}}$, and combined Gaussians $\mathcal{G}^{\text{combined}}$. In the worst case, to apply our ground level RGB losses $\mathcal{L}_{\text{RGB}}^{\text{ground}}$, $\mathcal{L}_{\text{RGB}}^{\text{sat}}$, and $\mathcal{L}_{\text{RGB}}^{\text{combined}}$ we require three forward calls to the \texttt{gsplat} \cite{ye2025gsplat} rasterizer. Instead, we do two forward calls, one for $\mathcal{G}^{\text{ground}}$ and $\mathcal{G}^{\text{sat}}$ and alpha-blend to obtain:
\begin{equation}\label{eq:gsplat_double_pass}
    C_{\text{3DGS}}^{\text{combined}} \approx C_{\text{3DGS}}^{\text{ground}}+ (1 {-} \alpha_{\text{ground}}) C_{\text{3DGS}}^{\text{sat}},
\end{equation}
\noindent where $\alpha_{\text{ground}} = \!\sum_{i=1}^{M} \!\alpha_{\text{ground},i} \prod_{j=1}^{i-1} (1 {-} \alpha_{\text{ground},j})$ is the accumulated transparency for ground Gaussians. This formulation is not strictly equivalent to rendering the unified set $\mathcal{G}^{\text{combined}} = \mathcal{G}^{\text{ground}} \cup \mathcal{G}^{\text{sat}}$, because satellite Gaussians that would occlude ground Gaussians along the camera ray are implicitly omitted in \cref{eq:gsplat_double_pass}.

\boldparagraph{Scene normalization.}
The choice of scene normalization during training is a critical design choice. Approaches vary, including scaling by ground truth depth maps \cite{dust3r_cvpr24, wang2025vggt, wang2025moge} or camera baseline distances \cite{ye2024noposplat}, or adopting a fully metric coordinate system \cite{duisterhof2025mastrsfm, wang2025moge2}. As mentioned in Sec. 3.4, we adopt the per-batch $\ell_2$-norm scaling derived from back-projected depths \cite{dust3r_cvpr24,wang2025vggt} to normalize the depth and pose of the ground-level imagery. A crucial difference is that we also regress height maps $\mathbf{h}^{\text{sat}}$ relative to $I_0^{\text{ground}}$ from orthoimages with a known spatial resolution $r^{\text{sat}}$ (expressed in pixels per meter). The spatial resolution $r^{\text{sat}}$ is used to map between satellite pixel space and world coordinates. Although we regress a single per-pixel scalar value for height maps, its spatial consistency with the ground level depth and camera poses is paramount. Therefore, we integrate $\mathbf{h}^{\text{sat}}$ and $r^{\text{sat}}$ into the same normalization scheme. Specifically, we compute a scalar value $s$ with:
\begin{equation}
s = \frac{1}{M} \sum_{j=1}^{M} \| \bm \mu_j \|_2, \quad \text{where } \bm \mu_j = \operatorname{backproject}(d_j, \bm K_j, \bm T_j).
\end{equation}
$s$ is then used to normalize all metric quantities during training: camera poses,  depth maps, height maps, and the satellite spatial resolution factor: $
\hat{\bm T} = \frac{\bm T}{s}, \quad
\hat{d} = \frac{d}{s}, \quad
\hat{\bm h}^{\text{sat}} = \frac{\bm h^{\text{sat}}}{s}, \quad
\hat{r}^{\text{sat}} = s \cdot r^{\text{sat}}.
$ We then train our network to regress values in this normalized space.

\boldparagraph{Height ambiguity.} We have one unknown degree of freedom in our training data, mainly the height of the ground level camera with respect to the BEV height maps. We follow prior works \cite{sat2density, Sat2Density++} and set the ground level height to 2 meters off the ground for all datasets. Due to this ambiguity, it is possible that the satellite-to-ground level renders are not perfectly aligned with ground perspectives.

\begin{figure*}[!t]
    \centering
    \includegraphics[width=0.99\linewidth]{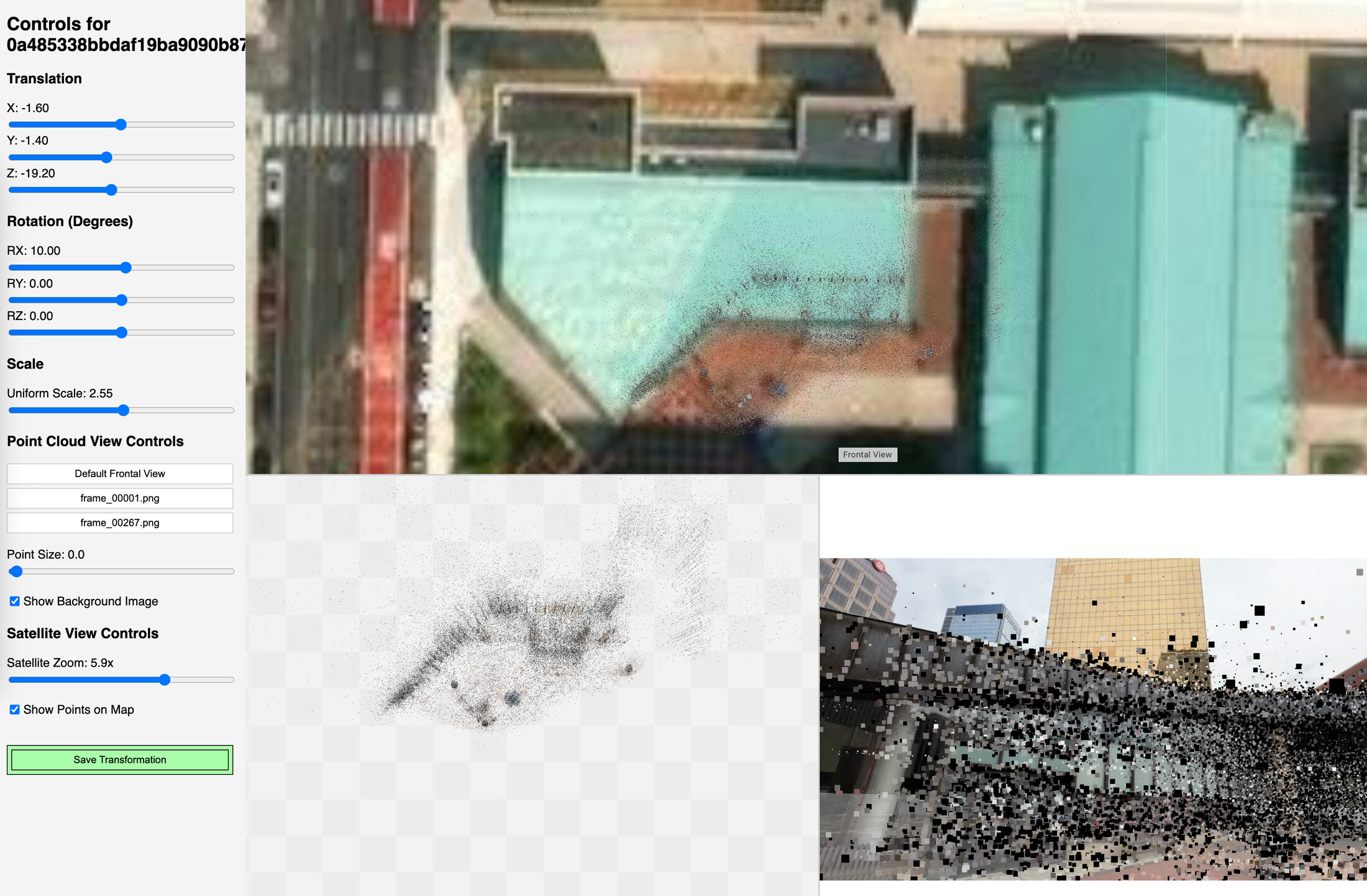}
    \caption{\textbf{Benchmark geoalignment tool.} We manually align COLMAP reconstructions to satellite imagery for 10 scenes from Tanks and Temples and 40 scenes from DL3DV-Benchmark datasets. Top: satellite image. Bottom-left: aligned COLMAP pointcloud. Bottom-right: visualization of points projected to a scene image.}
    \label{fig:geoalignment_tool}
\end{figure*}
\section{Geoaligned Benchmark Dataset} 

\boldparagraph{Benchmark alignment.} We introduce a new task, \textit{novel-view synthesis with geolocalized images}, and construct an evaluation dataset that remains in-domain with prior work to ensure fair comparison. Neither the Tanks and Temples (Table 2) nor the DL3DV-Benchmark (Table 3) datasets provide GPS metadata for geolocating ground-level images. We thus perform manual alignment to localize scenes. We first identify cues in the input images, such as street signs, building names, or distinctive landmarks (e.g., statues or monuments) and use these to obtain approximate GPS coordinates via Google Maps.

After establishing a coarse estimate, we refine the localization by projecting sparse COLMAP reconstruction points to satellite imagery. We used COLMAP reconstrutions provided by DL3DV. For Tanks and Temples, we used camera intrinsics and camera poses provided by the dataset and ran `colmap point\_triangulator' command to generate the sparse reconstruction.

\begin{figure*}
    \centering
    \input{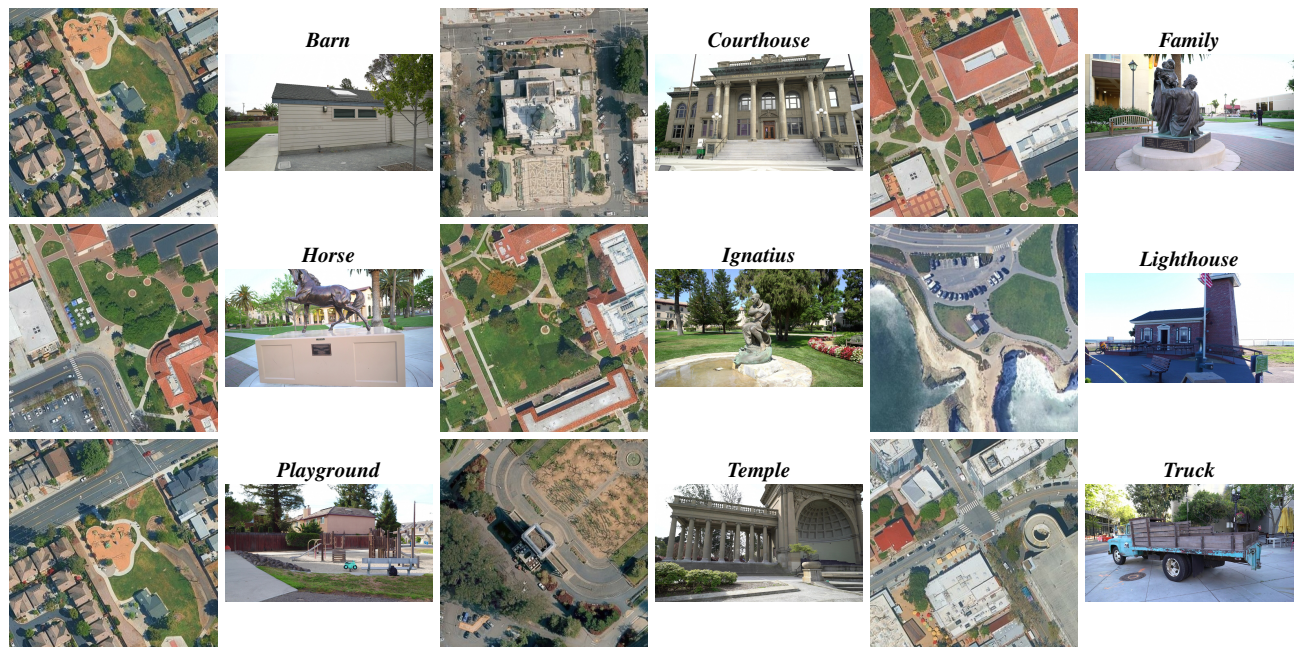}
\caption{\textbf{Our Tanks and Temples benchmark.} Visualization of satellite and ground level images.}
\label{fig:geoaligner_tandt}
\end{figure*}

\begin{figure*}
    \centering
    \input{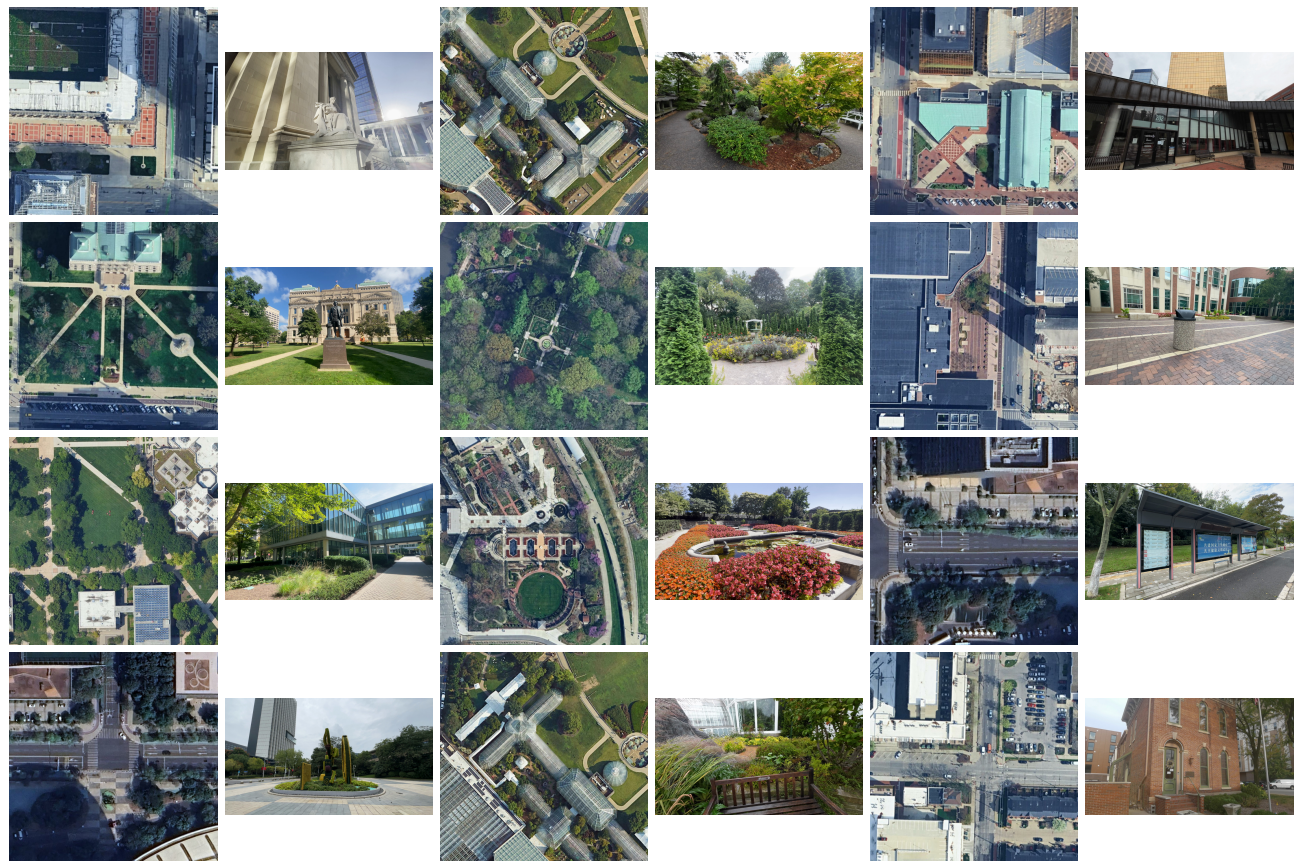}
\caption{\textbf{Our DL3DV benchmark.} Visualization of satellite and ground level images.}
\label{fig:geoaligner_dl3dv}
\end{figure*}

We then manually find translation, rotation, and scaling factor that aligns the point cloud to satellite imagery with known spatial resolution, transforming scenes to metric space. In \cref{fig:geoalignment_tool} we visualization the alignment process with a scene from the DL3DV-Benchmark dataset.
Since the COLMAP reconstructions for these scenes are rather dense, we hypothesize that this manual alignment is accurate within a few meters, but not pixel-perfect.
We will release the aligned COLMAP poses and location information for our benchmark scenes to facilitate further research in this area. We visualize satellite imagery and ground images for Tanks and Temples samples in \cref{fig:geoaligner_tandt} and DL3DV-Benchmark in \cref{fig:geoaligner_dl3dv}.


\begin{figure*}[t]
    \centering
    \begin{tikzpicture}[
        image/.style = {
            inner sep=0pt,
            outer sep=0pt,
            anchor=north west
        },
        node distance = 1pt and 1pt
    ]
    \setlength{\figurewidth}{0.135\textwidth}
    %% Row 1
    \node[image] (img-00) {\includegraphics[width=\figurewidth]{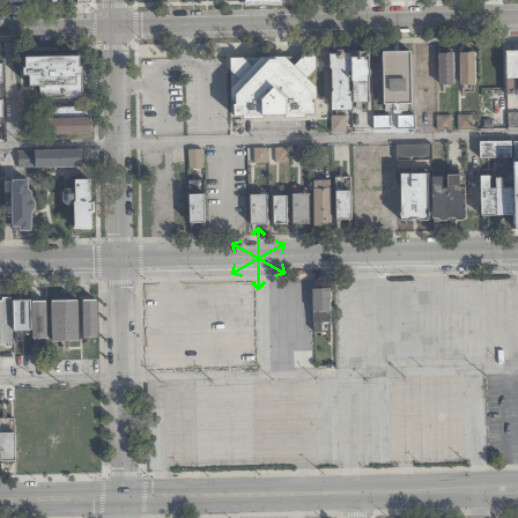}};
    \node[image, right=of img-00] (img-01) {\includegraphics[width=\figurewidth]{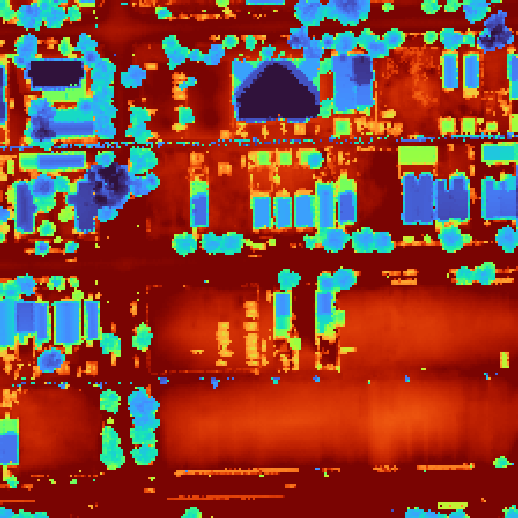}};
    \node[image, right=of img-01] (img-02) {\includegraphics[width=2\figurewidth]{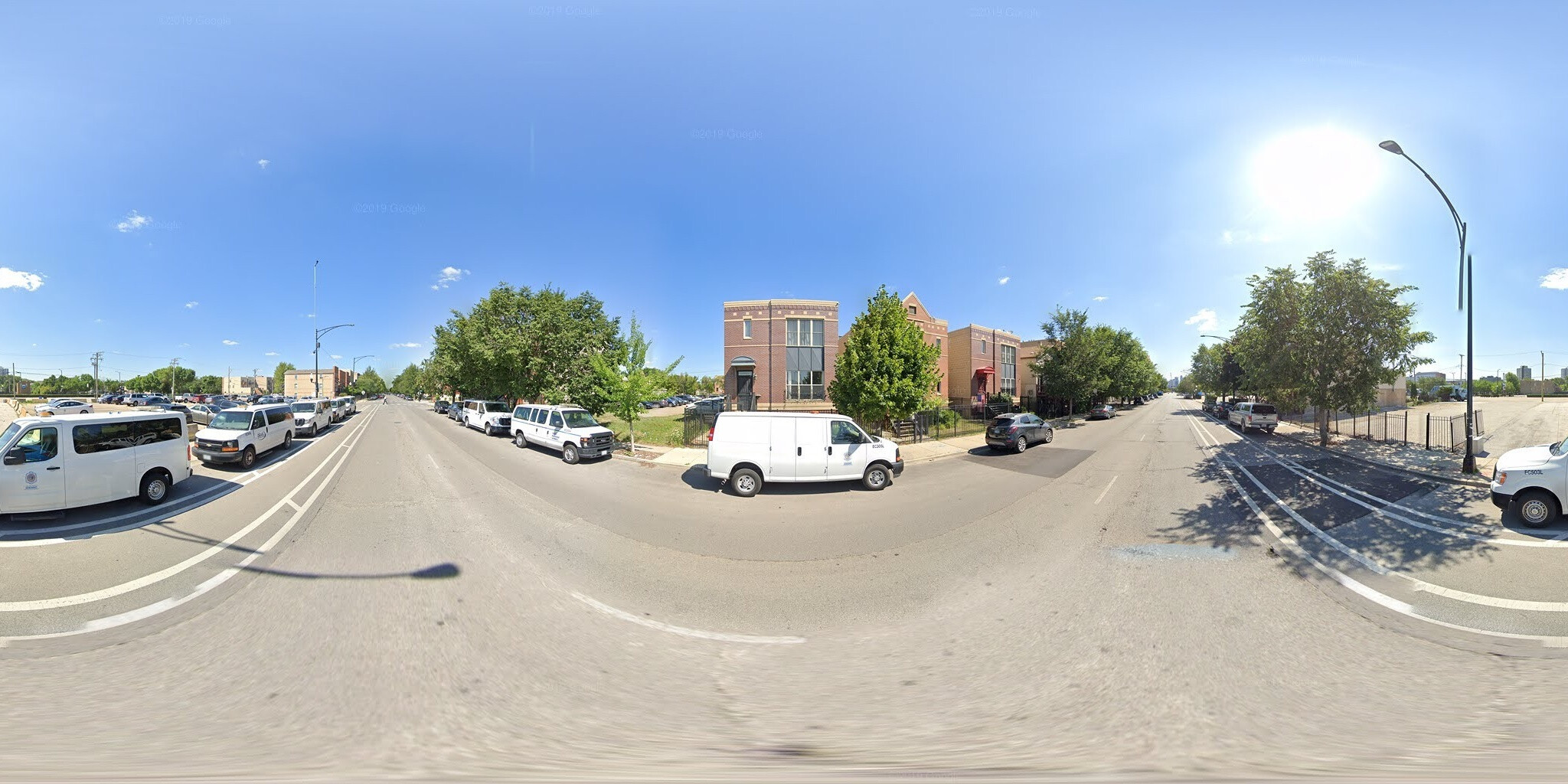}};
    \node[image, right=of img-02] (img-03) {\includegraphics[width=2\figurewidth]{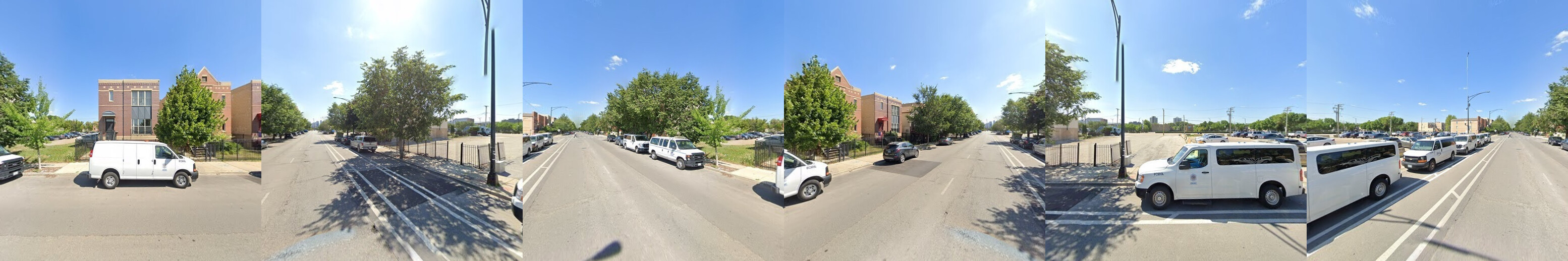}};
    
    %% Row 2
    \node[image, below=of img-00] (img-10) {\includegraphics[width=\figurewidth]{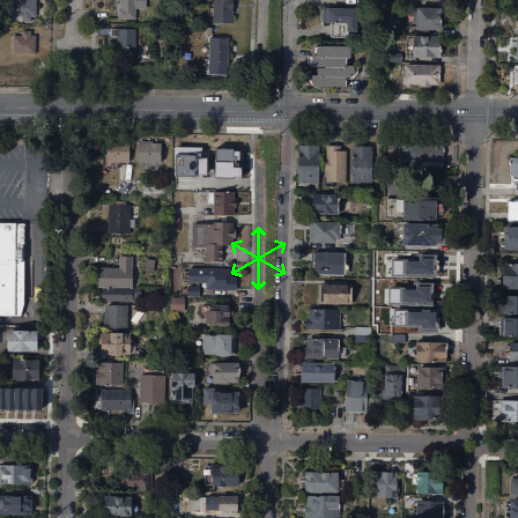}};
    \node[image, right=of img-10] (img-11) {\includegraphics[width=\figurewidth]{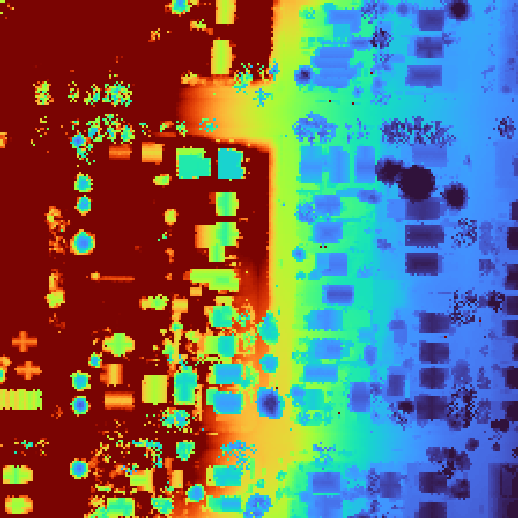}};
    \node[image, right=of img-11] (img-12) {\includegraphics[width=2\figurewidth]{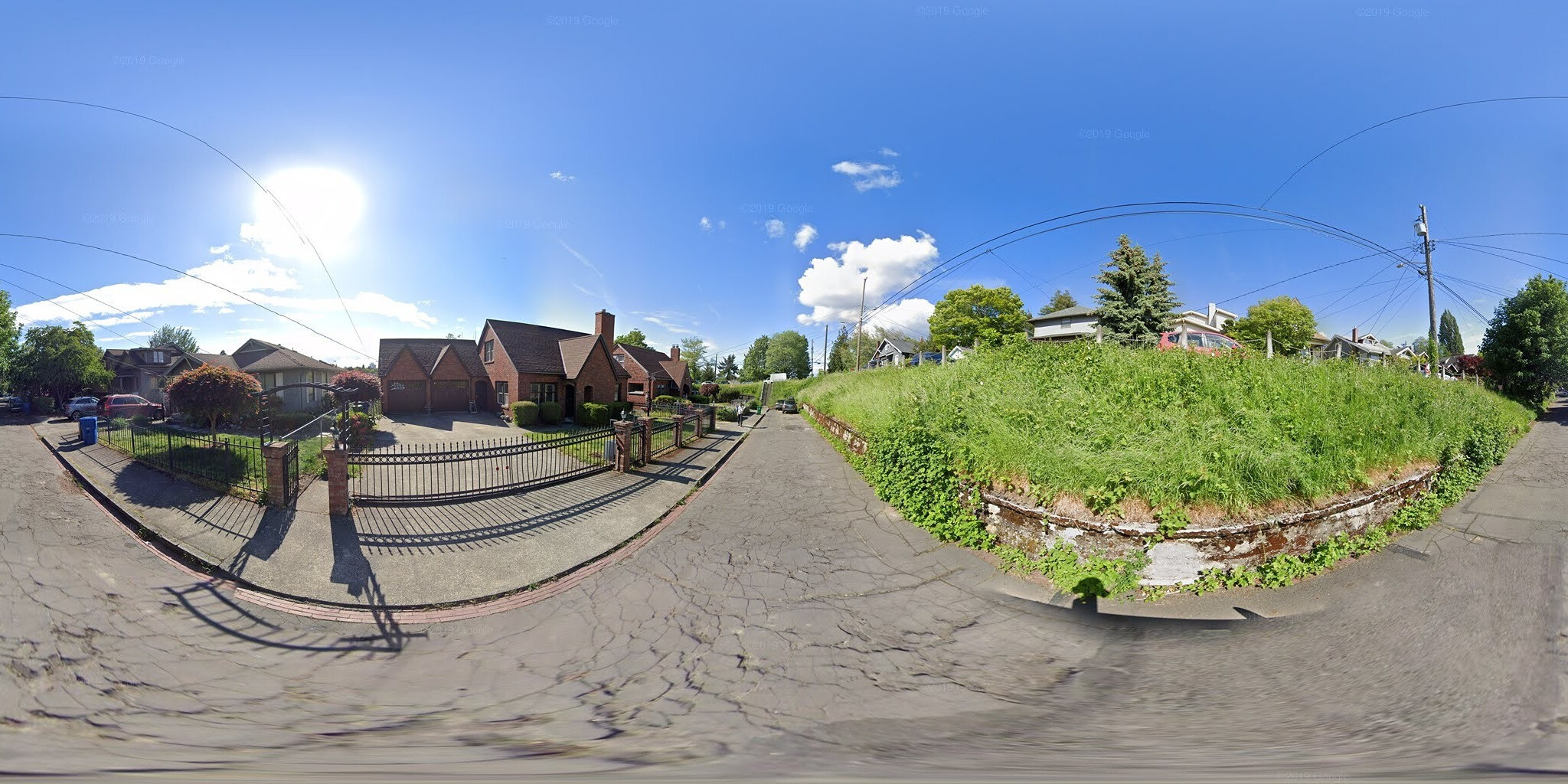}};
    \node[image, right=of img-12] (img-13) {\includegraphics[width=2\figurewidth]{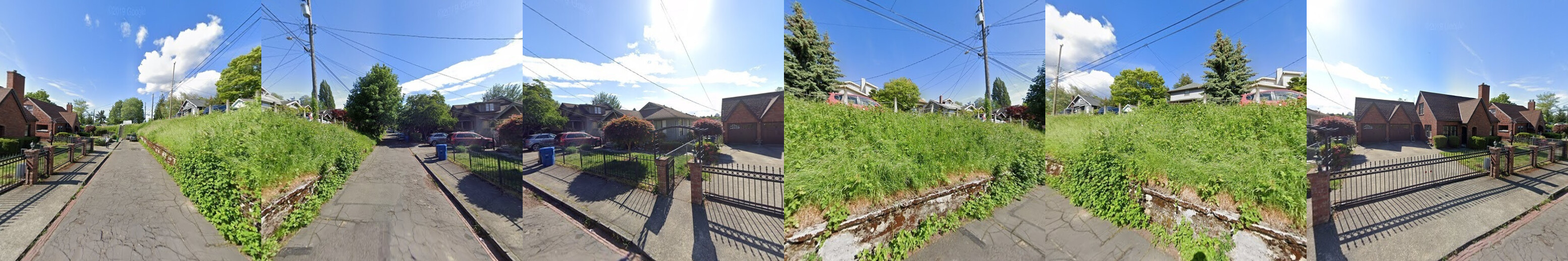}};

    %% Row 3
    \node[image, below=of img-10] (img-20) {\includegraphics[width=\figurewidth]{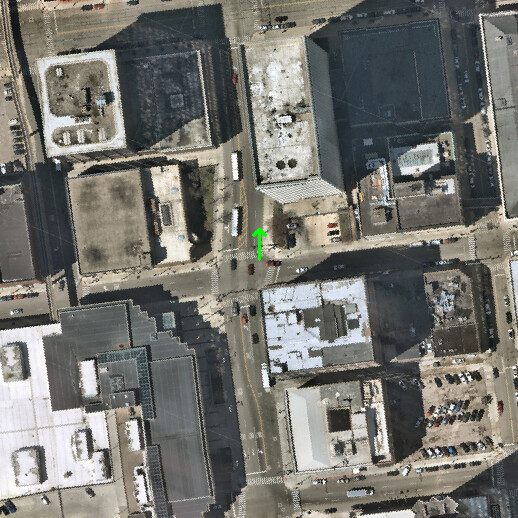}};
    \node[image, right=of img-20] (img-21) {\includegraphics[width=\figurewidth]{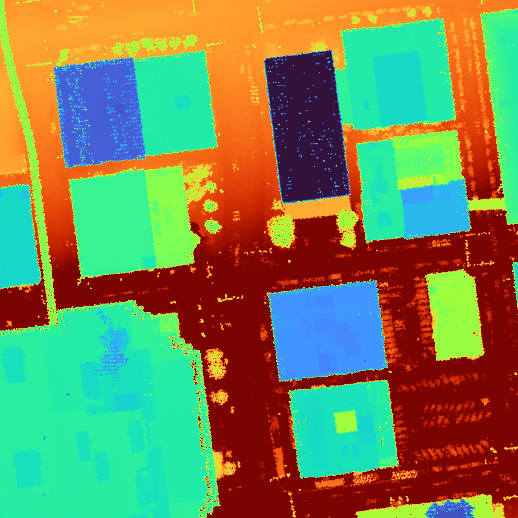}};
    \node[image, right=of img-21] (img-22) {\includegraphics[width=2\figurewidth]{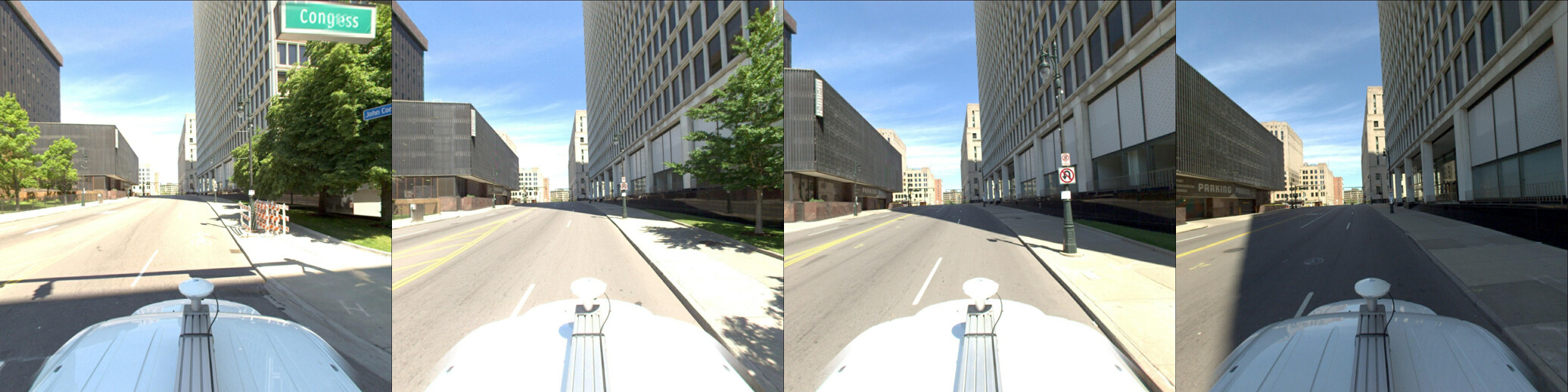}};
    \node[image, right=of img-22] (img-23) {\includegraphics[width=2\figurewidth]{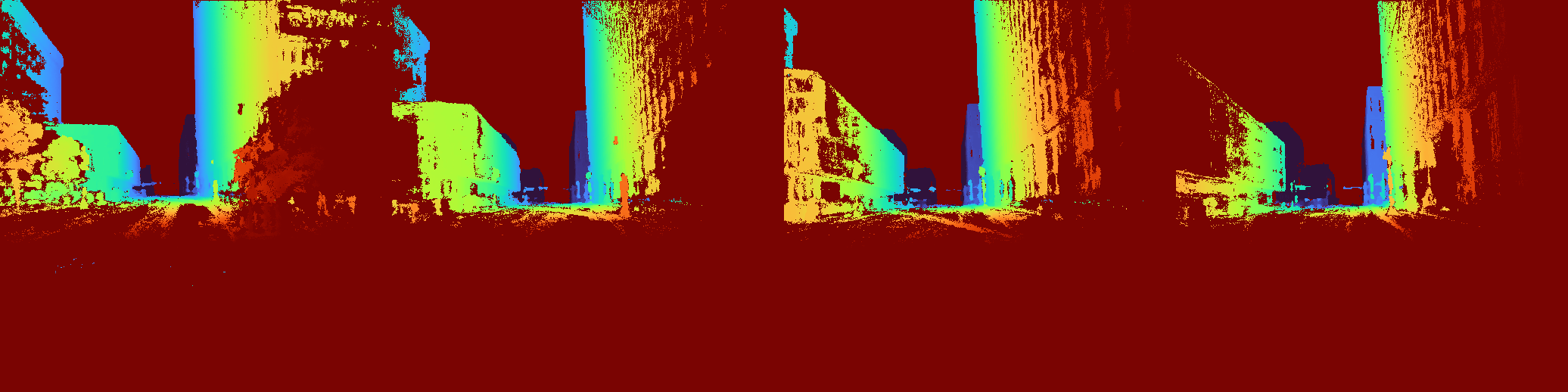}};

    %% Row 4
    \node[image, below=of img-20] (img-30) {\includegraphics[width=\figurewidth]{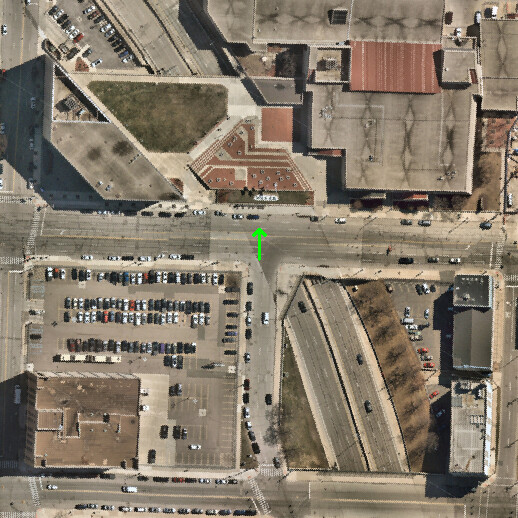}};
    \node[image, right=of img-30] (img-31) {\includegraphics[width=\figurewidth]{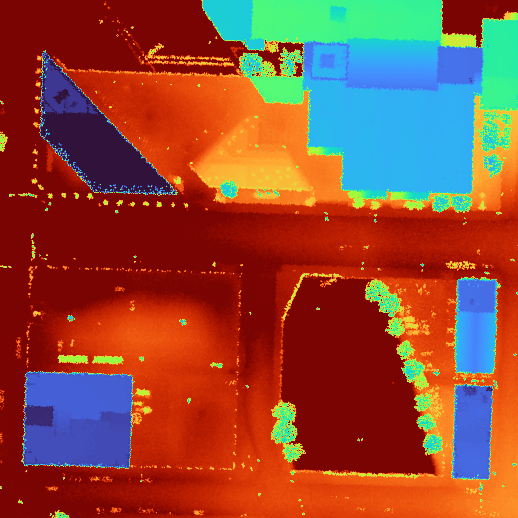}};
    \node[image, right=of img-31] (img-32) {\includegraphics[width=2\figurewidth]{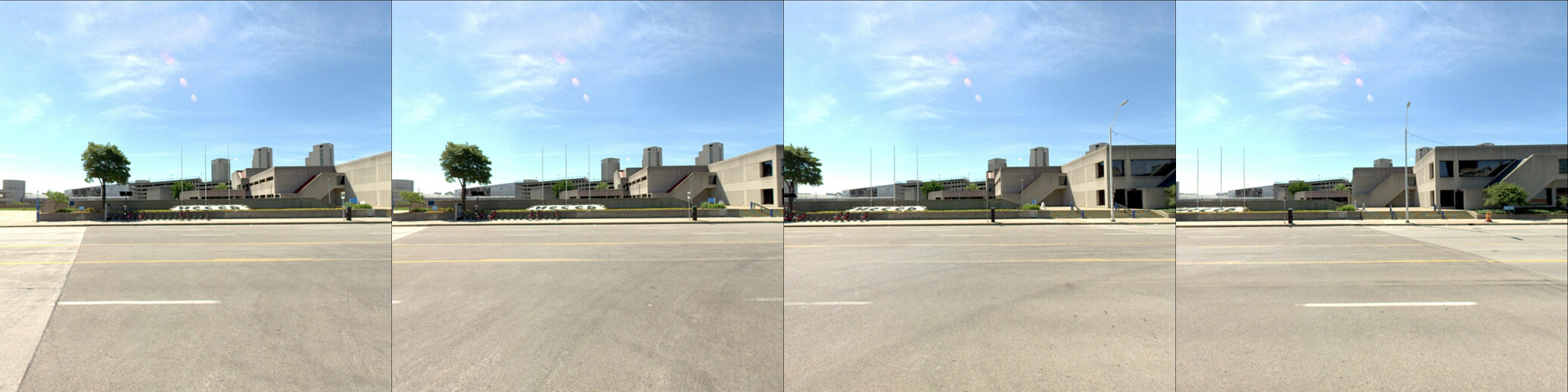}};
    \node[image, right=of img-32] (img-33) {\includegraphics[width=2\figurewidth]{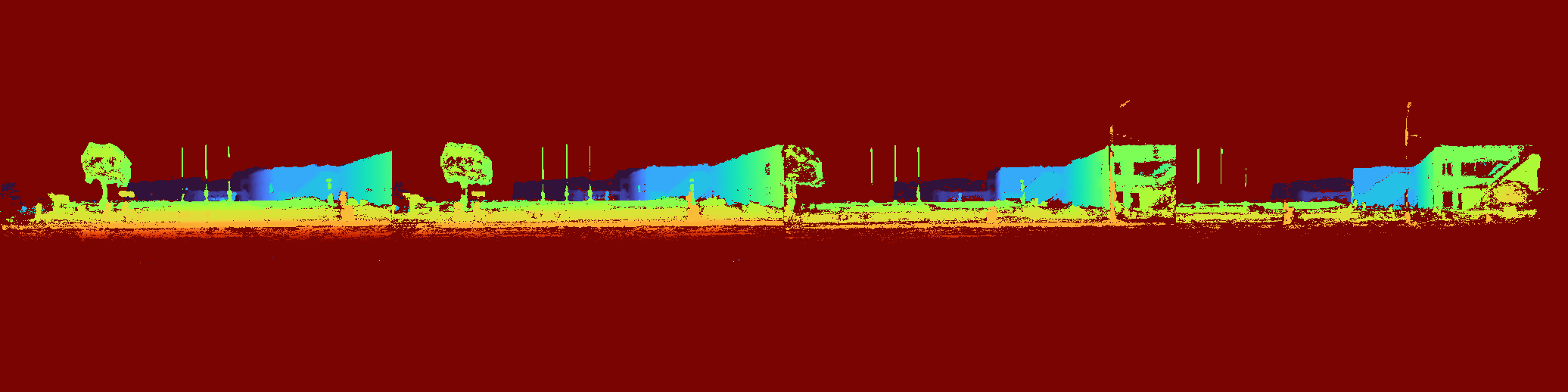}};

    % headings
    \node[label, yshift=0.15cm] (label1) at (img-00.north) {Satellite};
    \node[label, yshift=0.15cm] (label2) at (img-01.north) {GT Height Map};
    \node[label, yshift=0.15cm] (label3) at (img-02.north) {Panorama};
    \node[label, yshift=0.15cm] (label4) at (img-03.north) {Perspective Images};

    \node[label, yshift=0.15cm] (label3) at (img-22.north) {Perspective Images};
    \node[label, yshift=0.15cm] (label4) at (img-23.north) {Depths};
    
    % scene labels
    \node[label, xshift=-0.15cm, rotate=90,] (scene1) at (img-00.west) {VIGOR\cite{zhu2021vigor}};
    \node[label, xshift=-0.15cm, rotate=90] (scene2) at (img-10.west) {VIGOR\cite{zhu2021vigor}};
    \node[label,xshift=-0.15cm,rotate=90] (scene3) at (img-20.west) {Metropolis\cite{MapillaryMetropolis}};
    \node[label,xshift=-0.15cm,rotate=90] (scene3) at (img-30.west) {Metropolis\cite{MapillaryMetropolis}};

    \end{tikzpicture}

    \caption{\textbf{Training data visualization}. We showcase our training data that consists of satellite images and terrain height maps aligned with ground level images.}
    \label{fig:training_data}
\end{figure*}

\boldparagraph{Test split creation.} We construct our context and target view splits such that we have a range of increasingly challenging and representative scenarios where overlap is reduced. 

To compute frame overlap for a frame pair, we compute the IOU of visible COLMAP tracked points. We count the number of COLMAP tracked points visible in both frames and divide by the union of points across frames. 

For all splits, we pick the first image in the sequence as a context view. For the 2 and 3 context-view splits, we greedily select context frames that most closely satisfy a target IOU overlap to the first context image ($0.15$ for DL3DV and $0.25$ for Tanks and Temples). We then select four target frames that each satisfy an average IOU to selected context frames. Those targets are $0.02$, $0.05$, $0.07$, and $0.1$ for DL3DV and $0.03$, $0.07$, $0.1$, and $0.15$ for Tanks and Temples. 

We've found that the same targets IOUs don't yield the desired behavior across datasets given differences in how COLMAP reconstructions were created and the image resolution affecting the number and coverage of tracked points. Therefore, we select a different set of target IOU values.

\boldparagraph{Evaluating baselines.} We evaluate all baselines using their publicly available code and pretrained weights, strictly following each method’s recommended evaluation protocol. MVSplat and DepthSplat require ground-truth camera poses for novel-view synthesis. NoPoSplat first reconstructs a splat and then refines each novel camera pose for 200 iterations to align it with the reconstruction. Long-LRM uses Plucker rays for the target-views. AnySplat performs two forward passes: one using only the context views, and another using both context and target views; the latter provides estimated target poses that are then used to evaluate the context-only model. We adopt the same evaluation settings as AnySplat for our \ours evaluation protocol.

\section{Training Data}
\boldparagraph{Dataset reproducibility.} Our satellite API sources (Google, Azure, Esri) have licensing constraints on sharing. Therefore, we are unable to directly share raw satellite images; however, code to query data for georeferenced locations can be provided; although, exact replication of training data is uncertain due to the black-box nature of the APIs that can change with time. For our terrain height data, we are able to release the full raw training data. 

\boldparagraph{VIGOR\cite{zhu2021vigor}.} The original VIGOR dataset contains panorama images with non-centered satellite images with large zoom levels. We regenerate the dataset satellite images such that they are centered at the panorama latitude, longitude location. We also generate height maps for these locations. We create perspective images with $90^\circ$ FOV from the panoramas and sample these as our context and target images during training.

\boldparagraph{Metropolis~\cite{MapillaryMetropolis}.} The Mapillary Metropolis dataset provides high-resolution satellite imagery, perspective driving images (captured from forward, backward, left, and right cameras), panoramas, and MVS depth maps. We use the original satellite images and extract centered crops at ground-level positions. We also project Lidar depths to satellite images using the GDAL~\cite{gdal2024} library and these serve as our height maps. The forward and backward perspective images suffer from severe occlusions caused by the vehicle itself, and we observed that training directly on this data leads to degraded reconstruction quality due to multi-view inconsistencies. To mitigate this, we mask out thge vehicle using binary masks generated by SAM2.

We visualize various training data samples in \cref{fig:training_data}.

\section{Discussion of baselines} 

\boldparagraph{AnySplat~\cite{jiang2025anysplat}.}
We use AnySplat model at \url{https://github.com/InternRobotics/AnySplat} for our model initialisation and comparisons. The model was trained on DL3DV-10K~\cite{ling2024dl3dv} dataset, however the 140 scenes used in DL3DV-Benchmark split were removed from the training split, see \url{https://github.com/InternRobotics/AnySplat/issues/9}.

\boldparagraph{Long-LRM~\cite{ziwen2025llrm}.}
Long-LRM also removes the 140 scenes used in DL3DV-Benchmark split from the training set, see \url{https://github.com/arthurhero/Long-LRM?tab=readme-ov-file#long-lrm-evaluation-results}.

\boldparagraph{FLARE~\cite{zhang2025flarefeedforwardgeometryappearance}.}
Unfortunately, FLARE uses a custom train/test split of original DL3DV-10K, see
\url{https://github.com/ant-research/FLARE/blob/main/assets/DL3DV.json}. Thus, their training set includes scenes from DL3DV-Benchmark making comparison unfair. 

Furthermore, FLARE trains on Megadepth dataset~\cite{megadepth}, which contains scenes of Tanks and Temples dataset~\cite{Knapitsch2017}, \eg scenes in folders 5000, 5001, 5002, 5003, 5004, 5005, 5006, 5007, 5008, 5009, 5010, 5011, 5012, 5013. 

As a result, we omit direct evaluation of FLARE on our benchmarks.

\boldparagraph{Non-public baselines.}
The paper does not show comparisons to some closely related methods. This is due to the fact that our evaluation method requires GPS locations for input images, so we cannot compare scores on existing benchmarks. We thus re-run baselines on our geolocalized evaluation scenes. For some methods, there was no code available to run at the time of the submission and re-implementation is non-trivial, see answer to ``Can reviews request comparison to closed source?'' on CVPR Reviewer-Guidelines page\footnote{\url{https://cvpr.thecvf.com/Conferences/2025/ReviewerGuidelines}}.

Below we list some of the competing methods that we aim to add to the evaluation table when official implementations are available.

% Per a passed 2024 PAMI-TC motion: whenever a comparison of published research without publicly available code / data / pretrained models is requested (i.e., requiring re-implementation), it should be appropriately justified if used as a basis for a paper decision. Exceptions apply when the change is minor to an already implemented method with available code / data, or re-implementing a method based on details provided in a publication is common practice in a sub-field. In any case, comparisons should only be requested if the publication and / or code has been available sufficiently ahead of the submission deadline.

\boldparagraph{GS-LRM~\cite{gslrm}.}
At the time of submission no code or model is available on the project web page \url{https://sai-bi.github.io/project/gs-lrm/}.

\boldparagraph{Bolt-3D~\cite{bolt3d}.}
At the time of submission no code or model is available on the project web page \url{https://szymanowiczs.github.io/bolt3d}.

\boldparagraph{Sat2Density++~\cite{Sat2Density++}.}
At the time of submission a pretrained model is not available on the project web page \url{https://qianmingduowan.github.io/sat2density-pp/}.

\section{More Experimental Analysis}
Here we provide more experimental analysis of the design choices of our proposed \ours.

\subsection{Comparison to Diffusion Based Method}
\begin{figure}[!tb]
    \centering
    \centering
    \begin{tikzpicture}[
        image/.style = {
            inner sep=0pt,
            outer sep=0pt,
            anchor=north west
        },
        node distance = 1pt and 1pt
    ]
    \setlength{\figurewidth}{0.157\textwidth}
    %% Row 1
    \node[image] (img-00) {\includegraphics[width=\figurewidth]{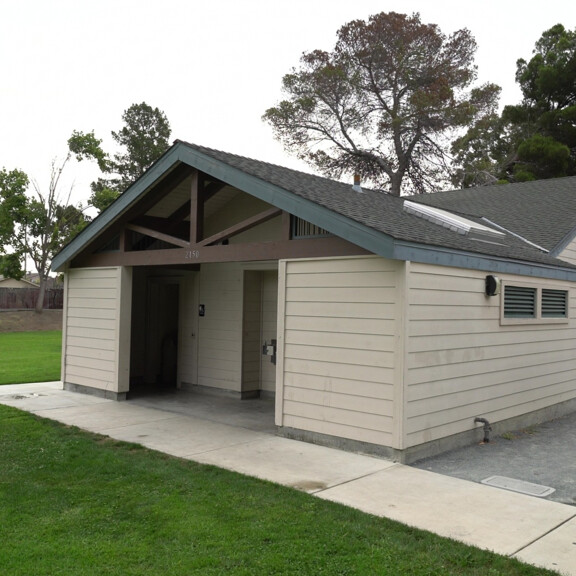}};
    \node[image, right=of img-00] (img-01) {\includegraphics[width=\figurewidth]{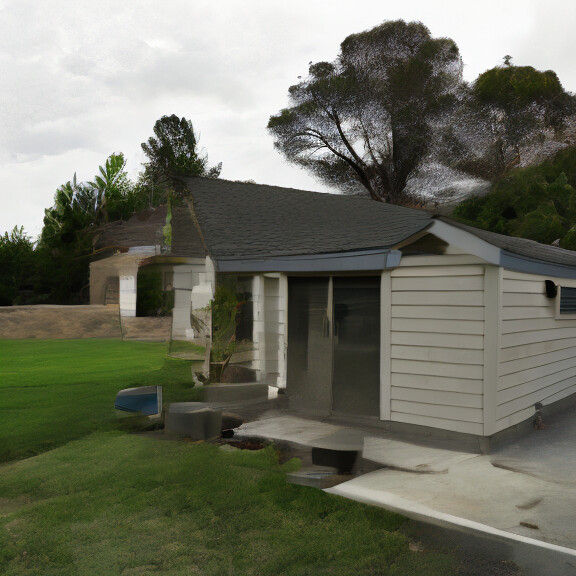}};
    \node[image, right=of img-01] (img-02) {\includegraphics[width=1\figurewidth]{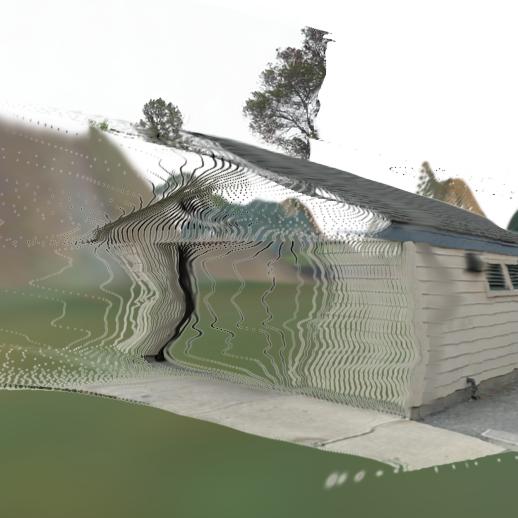}};

    %% Row 2
    \node[image, below=of img-00] (img-10) {\includegraphics[width=\figurewidth]{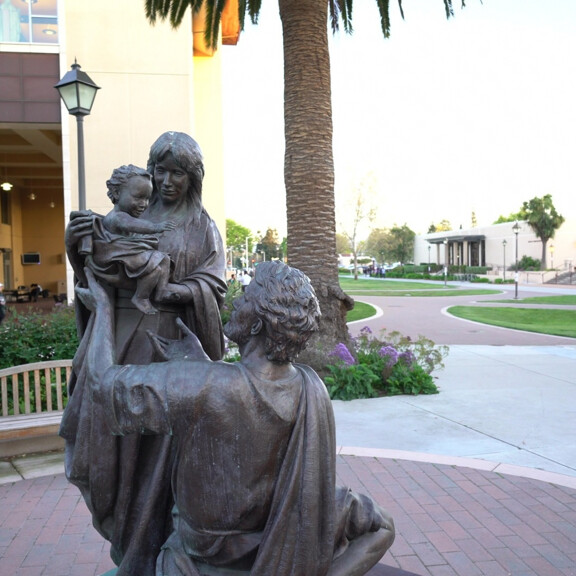}};
    \node[image, right=of img-10] (img-11) {\includegraphics[width=\figurewidth]{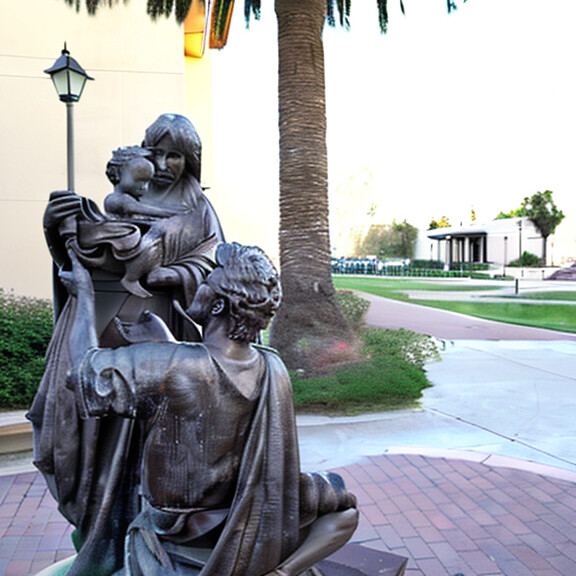}};
    \node[image, right=of img-11] (img-12) {\includegraphics[width=1\figurewidth]{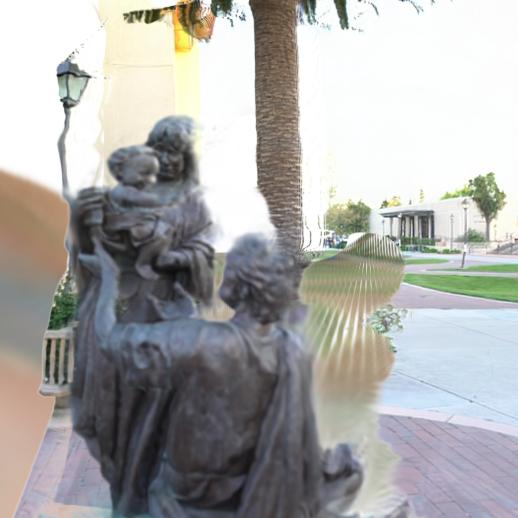}};

    % labels
    \node[label, yshift=0.15cm] (label1) at (img-00.north) {GT};
    \node[label, yshift=0.15cm] (label2) at (img-01.north) {SEVA};
    \node[label, yshift=0.15cm] (label3) at (img-02.north) {Ours};

    \end{tikzpicture}
    \caption{\textbf{Qualitative comparison to SEVA.} SEVA is a generative based model capable of hallucinating unseen areas whereas our \ours is a feed-forward approach that predicts geometry \textit{only for visible regions} in ground images and satellite image.}
    \label{fig:seva_qualitative}
    \vspace{-1em}
\end{figure}

We compare our method to Stable Virtual Camera (SEVA) \cite{zhou2025stable}, which is a state-of-the-art diffusion model for view-synthesis that takes in ground-level imagery and target render poses. We show qualitative renders in \cref{fig:seva_qualitative} and quantitative results in \cref{tab:seva}.

\begin{table*}[!t]
    \centering
    \caption{\textbf{Comparison to diffusion based SEVA model} on our geoaligned Tanks and Temples benchmark.}
    \label{tab:seva}
    \setlength{\tabcolsep}{5.0pt}
    \begin{tabular}{llc|ccc|ccc|ccc}
    \toprule
    & & & \multicolumn{3}{c}{1 context view} & \multicolumn{3}{c}{2 context views} & \multicolumn{3}{c}{3 context views}\\
    \textbf{Method} & & GT Pose? & 
    \tiny{PSNR$\uparrow$} & \tiny{SSIM$\uparrow$} & \tiny{LPIPS$\downarrow$} & 
    \tiny{PSNR$\uparrow$} & \tiny{SSIM$\uparrow$} & \tiny{LPIPS$\downarrow$} & 
    \tiny{PSNR$\uparrow$} & \tiny{SSIM$\uparrow$} & \tiny{LPIPS$\downarrow$}\\
    \midrule
    \texttt{SEVA} & & \checkmark 
    & 10.39 & 0.3024 & \textbf{0.6066}
    & \textbf{12.09} & 0.3614 & \textbf{0.5284}
    & \textbf{12.65} & 0.3723 & \textbf{0.5034} \\

    \midrule
    \rowcolor{gray!10} \texttt{Ours} & \textit{Combined} & - &
    \textbf{11.13} & \textbf{0.3764} & 0.6286
    & 11.67 & \textbf{0.3725} & 0.5984
    & 12.00 & \textbf{0.3855} & 0.5699 \\
    \bottomrule
    \end{tabular}
\end{table*}

\subsection{GPS Sensitivity Analysis}
We conduct satellite alignment sensitivity analysis in \cref{tab:sensitivity}, simulating GPS noise and showing that \textit{Ours} is robust to noise. Specifically, we add Gaussian noise to the 3DoF translation and rotation at increasing intervals ($\sigma$) and report mean and variance after five runs (random seeds). As satellite images have a lower sampling density compared to ground views, minor coordinate shifts do not significantly impact ground NVS renders.

\begin{table}[th]
    \centering\scriptsize
    \setlength{\tabcolsep}{1.2pt}
    \renewcommand*{\arraystretch}{1}
    \caption{\textbf{GPS sensitivity analysis} results for the 1-context view setting for \textit{Combined} (\ours) method on Tanks \& Temples.}
    \label{tab:sensitivity}
    \begin{tabular}{lc|lc}
        \toprule
        \textbf{Trans. Noise ($\sigma$)} & \textbf{\textit{Combined} PSNR$\uparrow$} & \textbf{Rot. Noise ($\sigma$)} &  \textbf{\textit{Combined} PSNR$\uparrow$} \\
        \midrule
        0m (Manual aligned) & 11.13 & 0$^\circ$ (Manual aligned) & 11.13 \\
        1m  & 11.09 $\pm$ 0.04 & 5$^\circ$   & 11.14 $\pm$ 0.03 \\
        3m  & 11.13 $\pm$ 0.06 & 10$^\circ$  & 11.16 $\pm$ 0.18 \\
        5m  & 11.12 $\pm$ 0.04 & 15$^\circ$  & 11.11 $\pm$ 0.12 \\
        %10m & 11.12 $\pm$ 0.16 & 20$^\circ$  & 11.03 $\pm$ 0.07 \\
        \bottomrule
    \end{tabular}
\end{table}

\section{More Qualitative Analysis}
\begin{figure*}[!th]
    \centering
    \begin{tikzpicture}[
        image/.style = {
            inner sep=0pt,
            outer sep=0pt,
            anchor=north west
        },
        node distance = 2pt and 1pt
    ]
    \setlength{\figurewidth}{0.98\linewidth}

    %% One image per row
    \node[image] (img-00) {\includegraphics[width=\figurewidth]{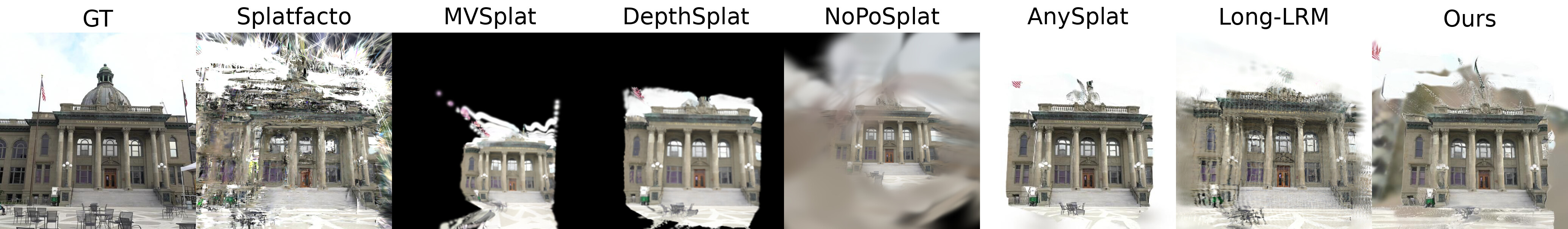}};
    \node[image, below=of img-00] (img-01) {\includegraphics[width=\figurewidth]{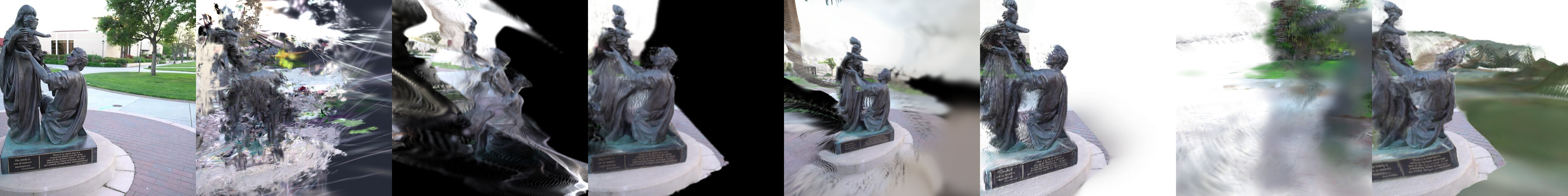}};
    \node[image, below=of img-01] (img-02) {\includegraphics[width=\figurewidth]{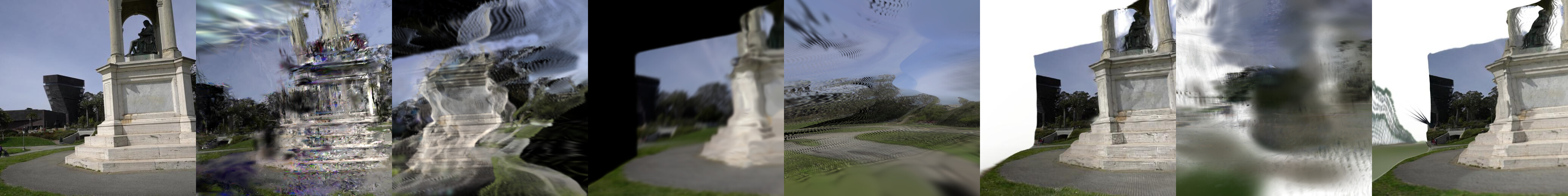}};
    \node[image, below=of img-02] (img-03) {\includegraphics[width=\figurewidth]{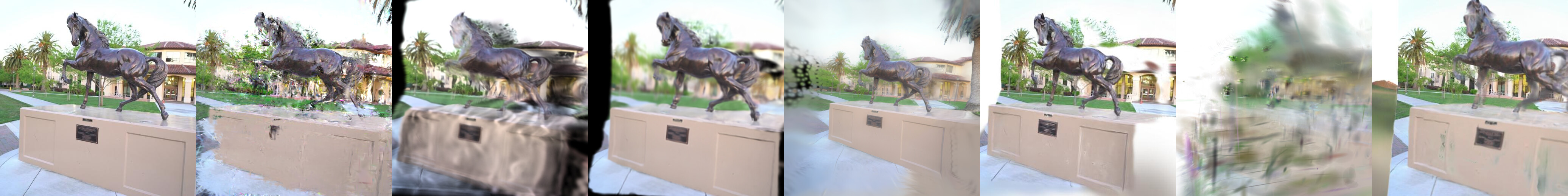}};
    \node[image, below=of img-03] (img-04) {\includegraphics[width=\figurewidth]{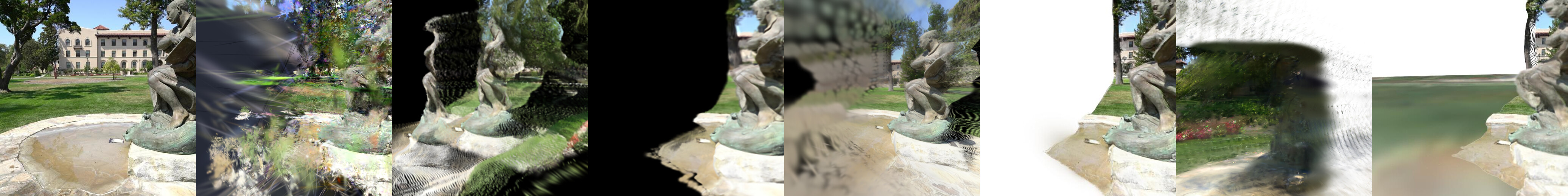}};
    \node[image, below=of img-04] (img-05) {\includegraphics[width=\figurewidth]{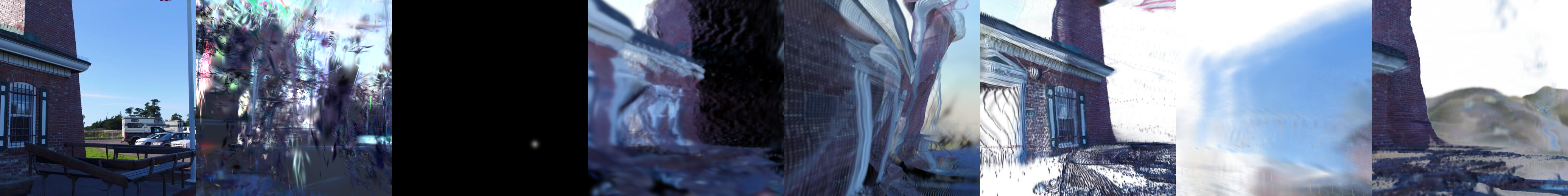}};
    \node[image, below=of img-05] (img-06) {\includegraphics[width=\figurewidth]{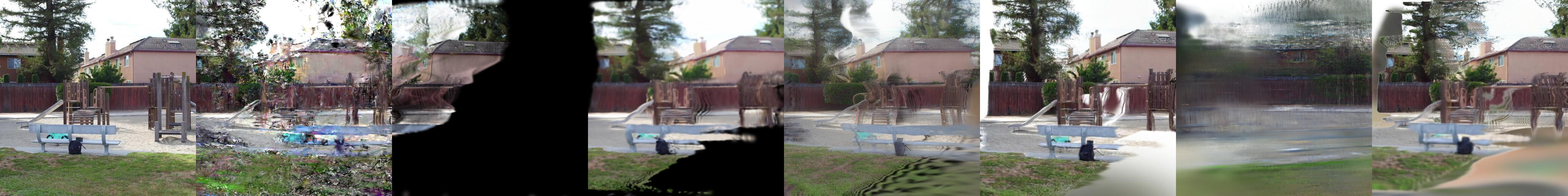}};
    \node[image, below=of img-06] (img-07) {\includegraphics[width=\figurewidth]{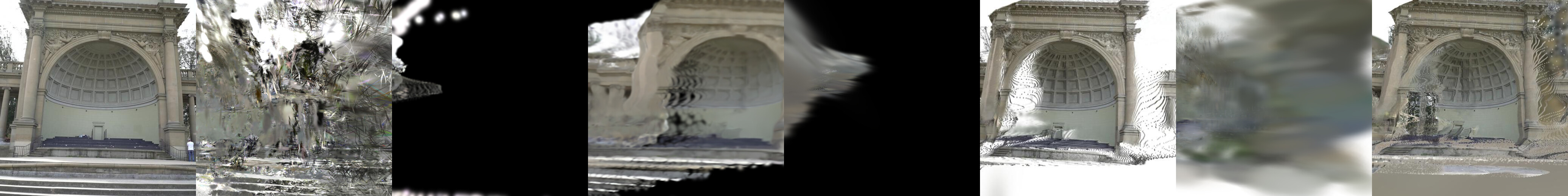}};
    \end{tikzpicture}
    \caption{\textbf{Qualitative baseline comparisons.} Additional qualitative comparisons of various baselines on our georeferenced Tanks and Temples dataset.}
    \label{fig:baselines_qualitative}
\end{figure*}

\begin{figure*}[!t]
    \centering
    \begin{tikzpicture}[
        image/.style = {
            inner sep=0pt,
            outer sep=0pt,
            anchor=north west
        },
        node distance = 1pt and 1pt
    ]
    \setlength{\figurewidth}{0.157\textwidth}
    %% Row 1
    \node[image] (img-00) {\includegraphics[width=\figurewidth]{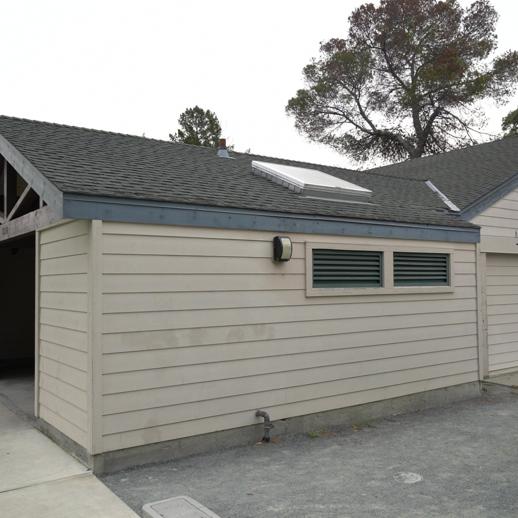}};
    \node[image, right=of img-00] (img-01) {\includegraphics[width=\figurewidth]{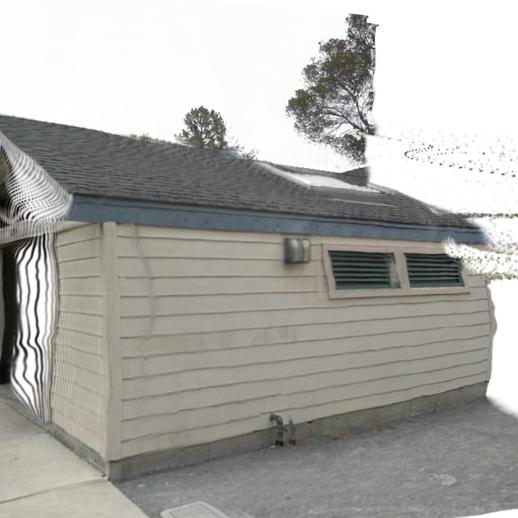}};
    \node[image, right=of img-01] (img-02) {\includegraphics[width=1\figurewidth]{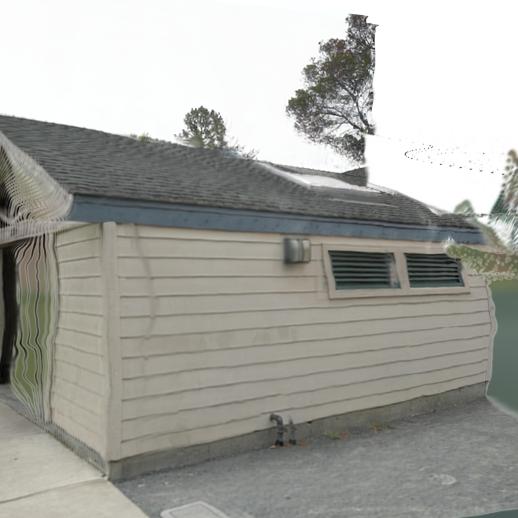}};
    \node[image, right=0.2cm of img-02] (img-03) {\includegraphics[width=\figurewidth]{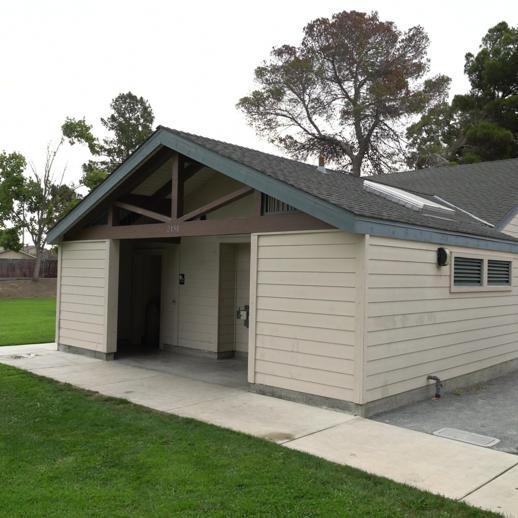}};
    \node[image, right=of img-03] (img-04) {\includegraphics[width=\figurewidth]{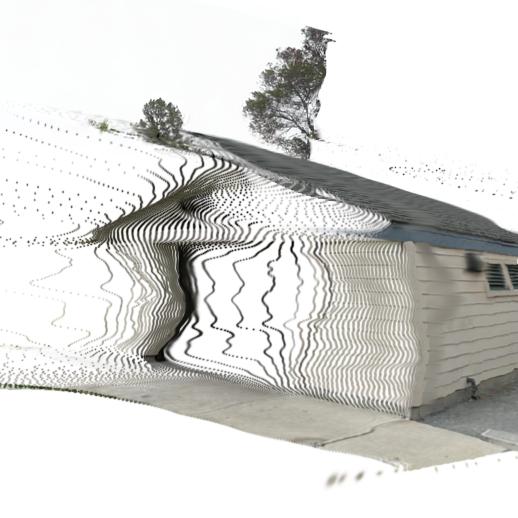}};
    \node[image, right=of img-04] (img-05) {\includegraphics[width=1\figurewidth]{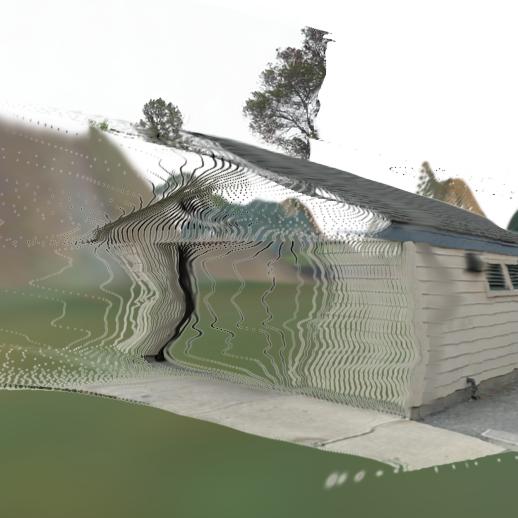}};

    %% Row 2
    \node[image, below=of img-00] (img-10) {\includegraphics[width=\figurewidth]{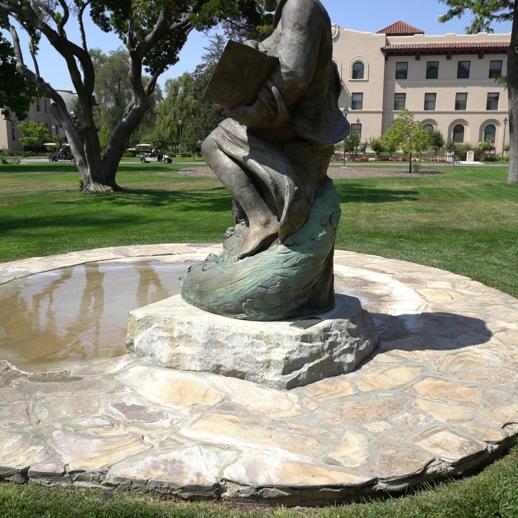}};
    \node[image, right=of img-10] (img-11) {\includegraphics[width=\figurewidth]{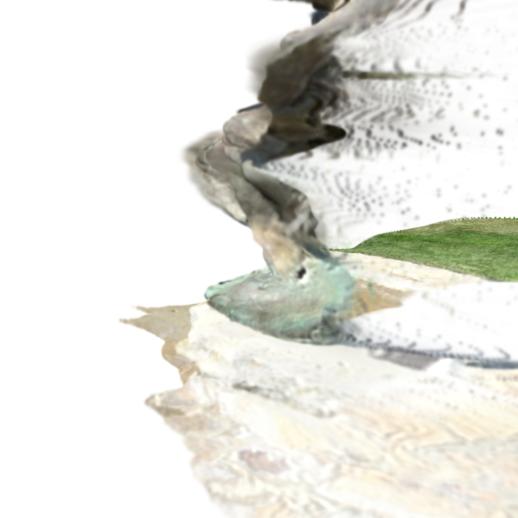}};
    \node[image, right=of img-11] (img-12) {\includegraphics[width=1\figurewidth]{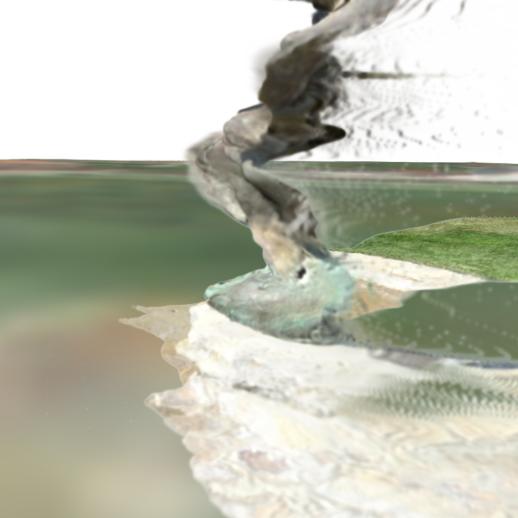}};
    \node[image, right=0.2cm of img-12] (img-13) {\includegraphics[width=\figurewidth]{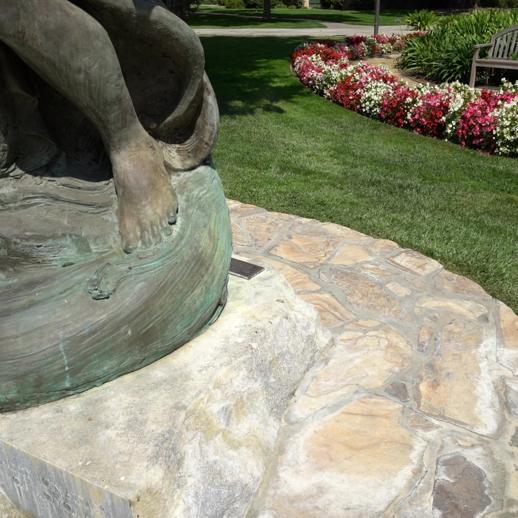}};
    \node[image, right=of img-13] (img-14) {\includegraphics[width=\figurewidth]{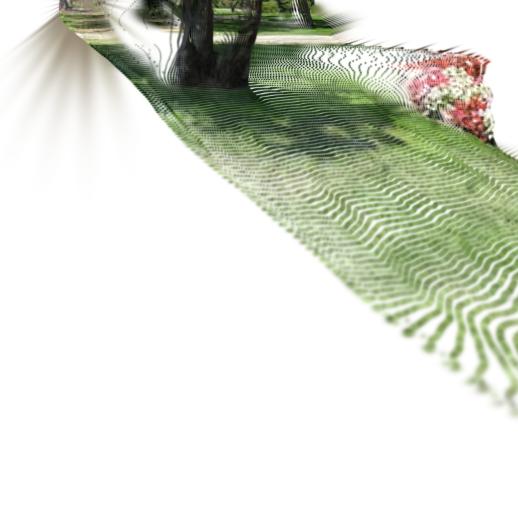}};
    \node[image, right=of img-14] (img-15) {\includegraphics[width=1\figurewidth]{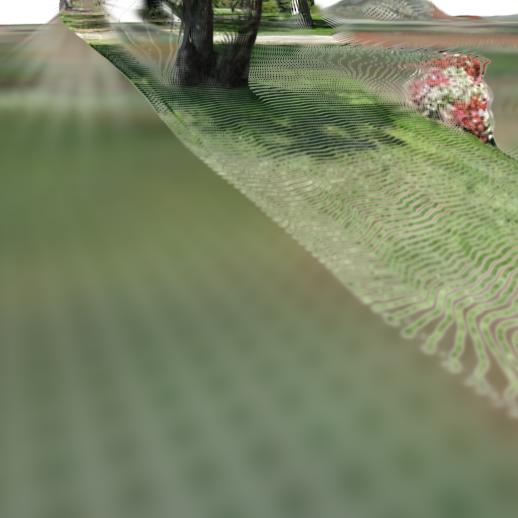}};

    %% Row 3
    \node[image, below=of img-10] (img-20) {\includegraphics[width=\figurewidth]{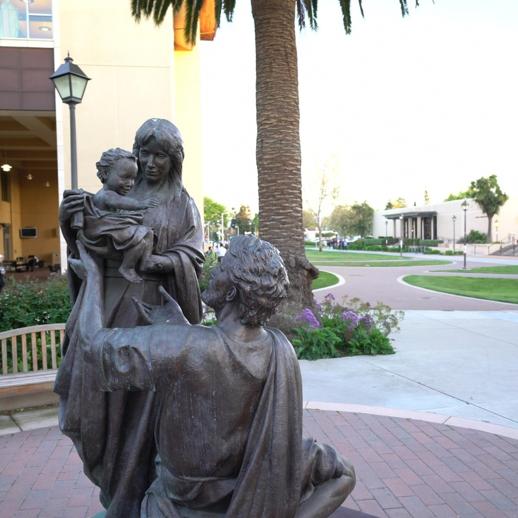}};
    \node[image, right=of img-20] (img-21) {\includegraphics[width=\figurewidth]{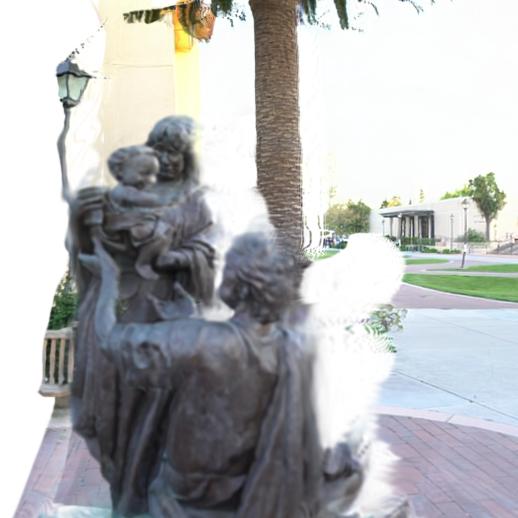}};
    \node[image, right=of img-21] (img-22) {\includegraphics[width=1\figurewidth]{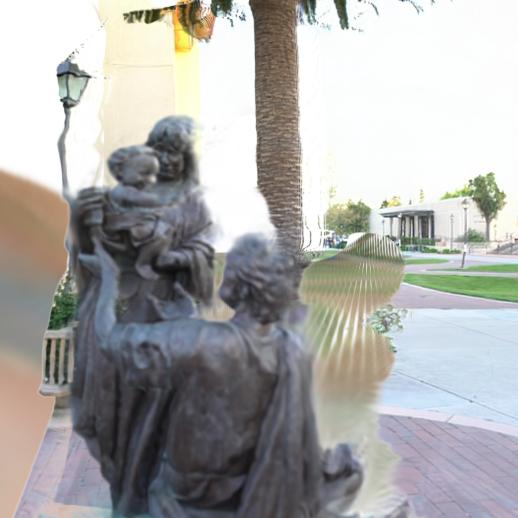}};
    \node[image, right=0.2cm of img-22] (img-23) {\includegraphics[width=\figurewidth]{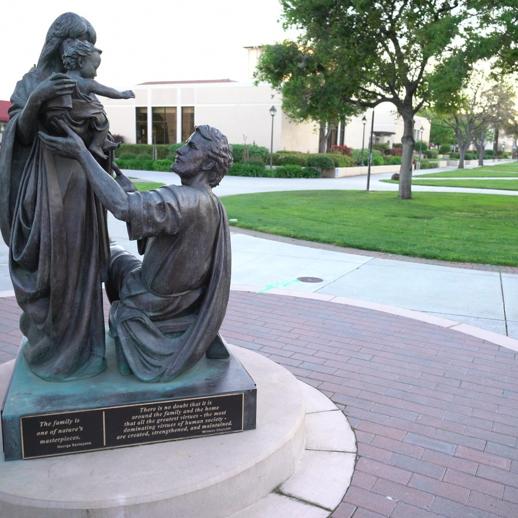}};
    \node[image, right=of img-23] (img-24) {\includegraphics[width=\figurewidth]{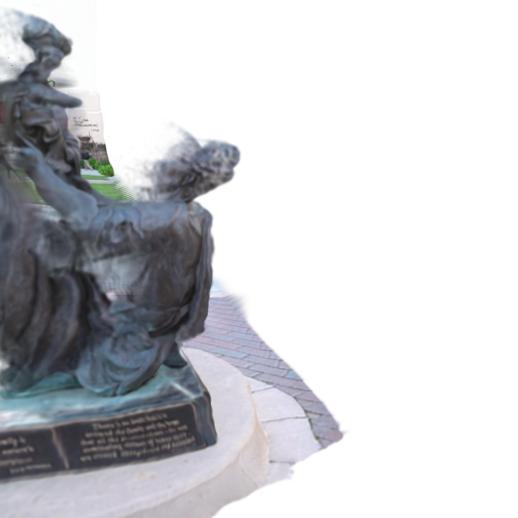}};
    \node[image, right=of img-24] (img-25) {\includegraphics[width=1\figurewidth]{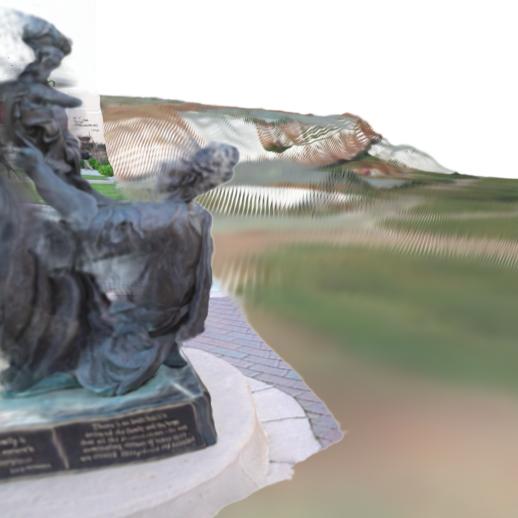}};

    %% Row 4
    \node[image, below=of img-20] (img-30) {\includegraphics[width=\figurewidth]{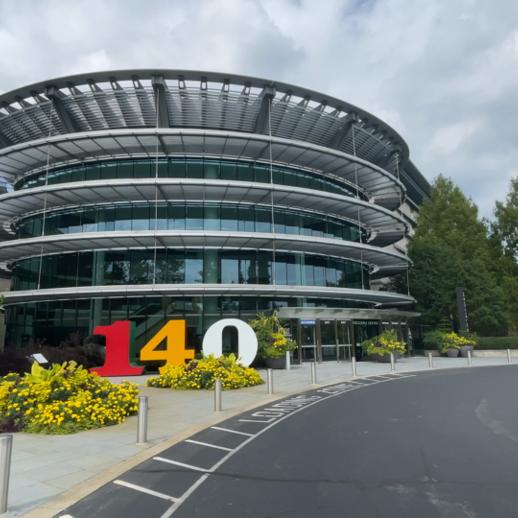}};
    \node[image, right=of img-30] (img-31) {\includegraphics[width=\figurewidth]{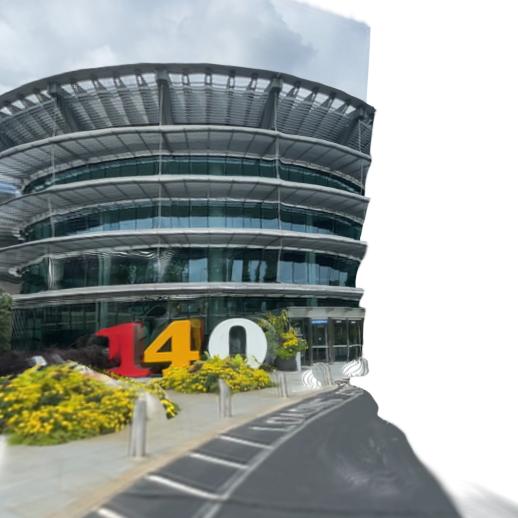}};
    \node[image, right=of img-31] (img-32) {\includegraphics[width=1\figurewidth]{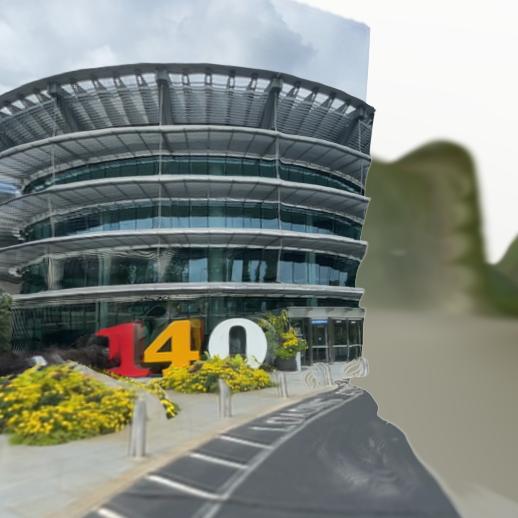}};
    \node[image, right=0.2cm of img-32] (img-33) {\includegraphics[width=\figurewidth]{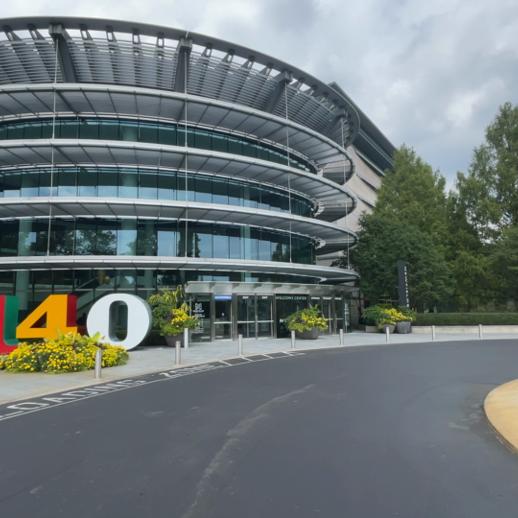}};
    \node[image, right=of img-33] (img-34) {\includegraphics[width=\figurewidth]{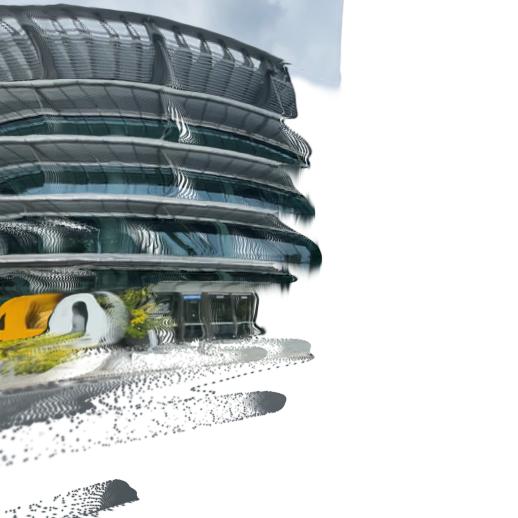}};
    \node[image, right=of img-34] (img-35) {\includegraphics[width=1\figurewidth]{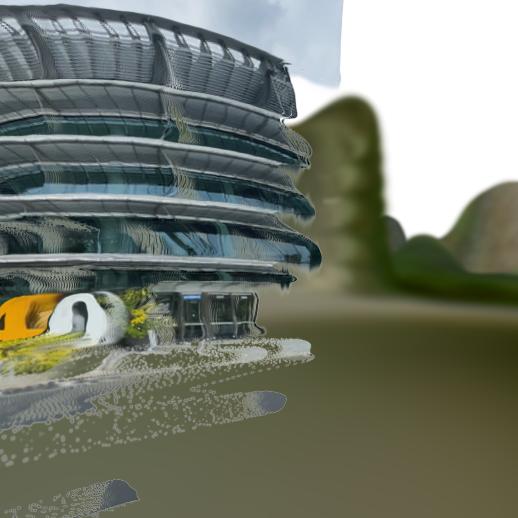}};

    % headings
    \node[label, yshift=0.15cm] (label1) at (img-00.north) {GT};
    \node[label, yshift=0.15cm] (label2) at (img-01.north) {Ground-Only};
    \node[label, yshift=0.15cm] (label3) at (img-02.north) {Full Model};
    \node[label, yshift=0.15cm] (label1) at (img-03.north) {GT};
    \node[label, yshift=0.15cm] (label2) at (img-04.north) {Ground-Only};
    \node[label, yshift=0.15cm] (label3) at (img-05.north) {Full Model};

    % scene labels
    \node[label, xshift=-0.15cm, rotate=90,] (scene1) at (img-00.west) {Barn};
    \node[label, xshift=-0.15cm, rotate=90] (scene2) at (img-10.west) {Ignatius};
    \node[label,xshift=-0.15cm,rotate=90] (scene3) at (img-20.west) {Family};
    \node[label,xshift=-0.15cm,rotate=90] (scene3) at (img-30.west) {cc08c0bdc34ddd};

    \end{tikzpicture}

    \caption{\textbf{\ours Ground-Only vs Full-Model}. We visualize the benefit of \ours's satellite branch on qualitative rendering on the Tanks and Temples and DL3DV benchmarks. Our Full-Model achieves better coverage and completeness compared to ground only imagery in sparse-view settings.}
    \label{fig:qualitative_ours_vs_ground}
\end{figure*}

\begin{figure*}[!t]
    \centering
    \begin{tikzpicture}[
        image/.style = {
            inner sep=0pt,
            outer sep=0pt,
            anchor=north west
        },
        node distance = 1pt and 1pt
    ]
    \setlength{\figurewidth}{0.135\textwidth}
    %% Row 1
    \node[image] (img-00) {\includegraphics[width=\figurewidth]{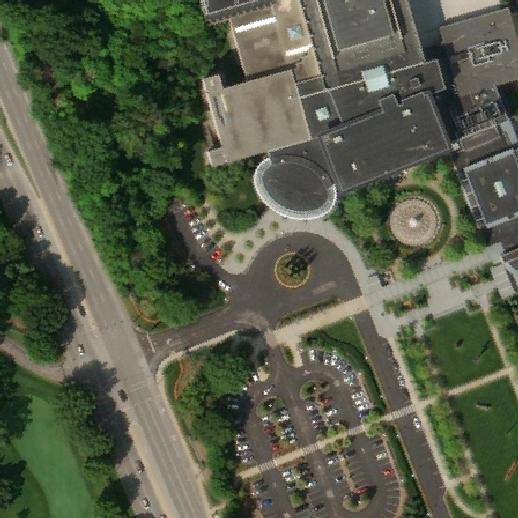}};
    \node[image, right=of img-00] (img-01) {\includegraphics[width=\figurewidth]{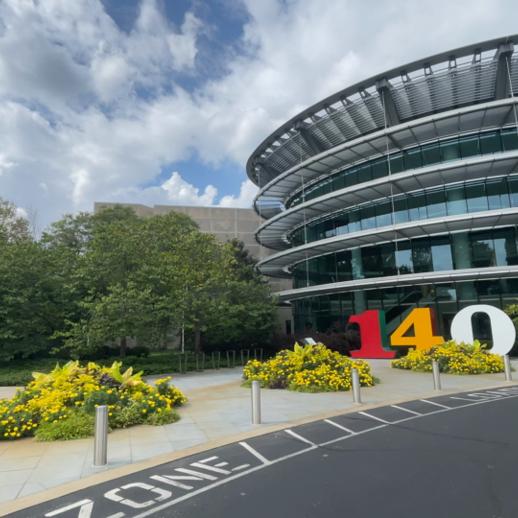}};
    \node[image, right=0.2cm of img-01] (img-02) {\includegraphics[width=\figurewidth]{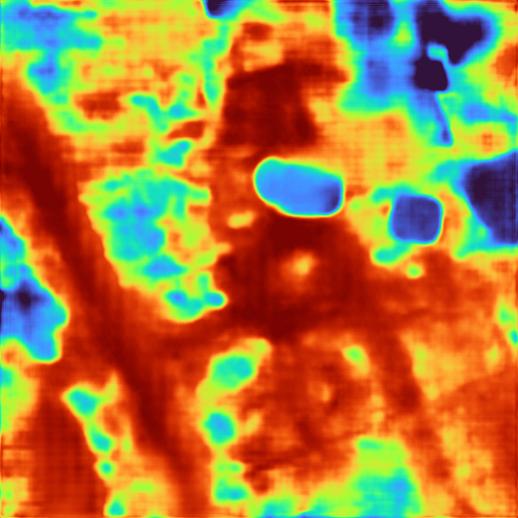}};
    \node[image, right=0.0cm of img-02] (img-03) {\includegraphics[width=\figurewidth]{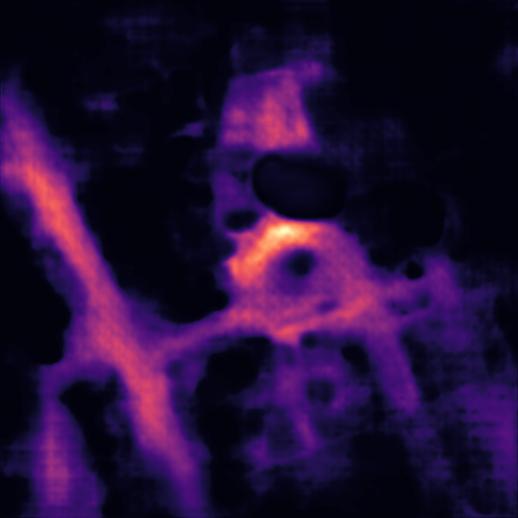}};
    \node[image, right=of img-03] (img-04) {\includegraphics[width=\figurewidth]{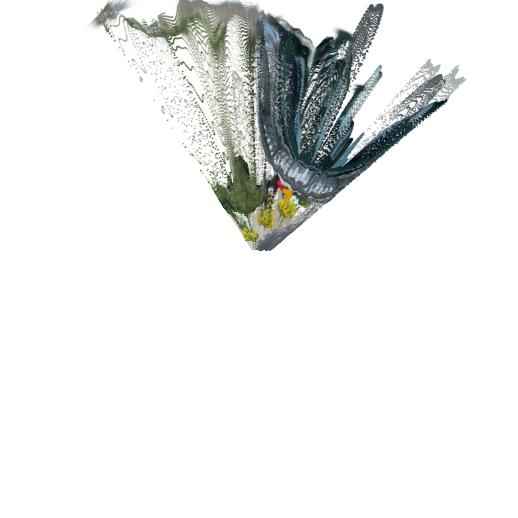}};
    \node[image, right=of img-04] (img-05) {\includegraphics[width=\figurewidth]{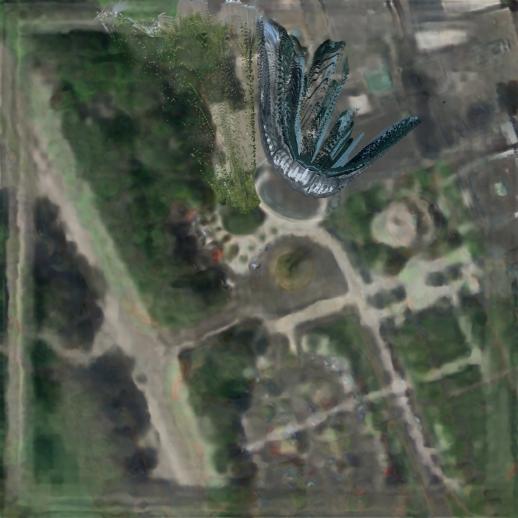}};
    \node[image, right=of img-05] (img-06) {\includegraphics[width=\figurewidth]{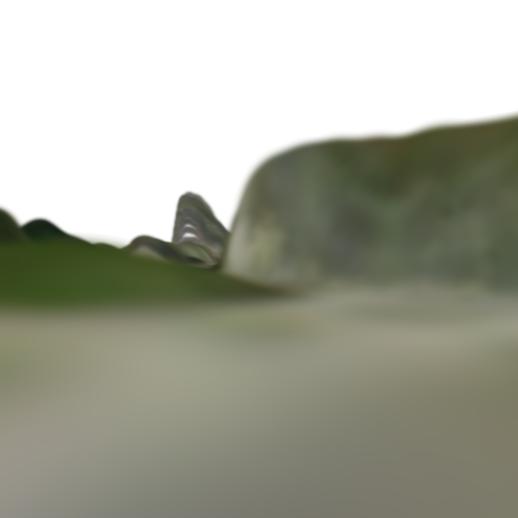}};

    %% Row 2
    \node[image, below=of img-00] (img-10) {\includegraphics[width=\figurewidth]{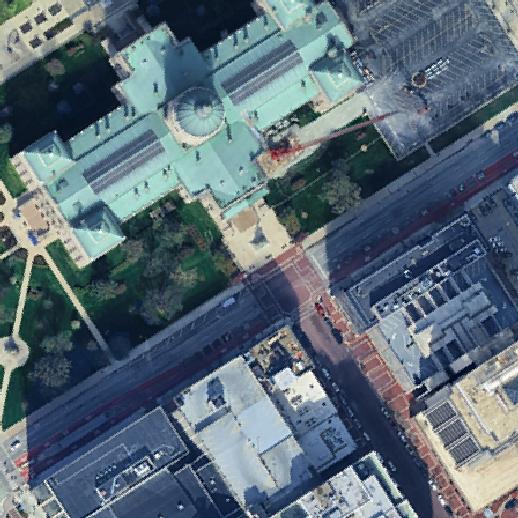}};
    \node[image, right=of img-10] (img-11) {\includegraphics[width=\figurewidth]{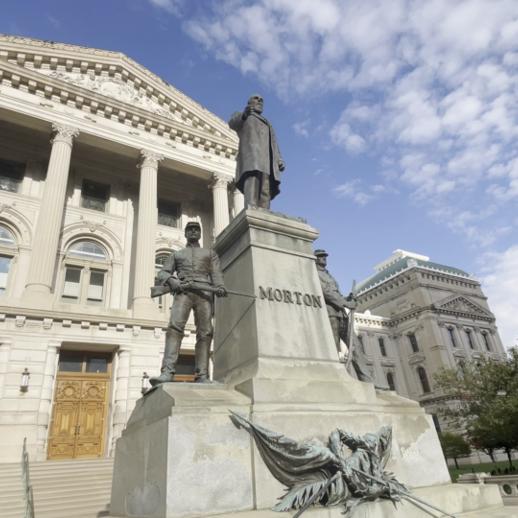}};
    \node[image, right=0.2cm of img-11] (img-12) {\includegraphics[width=\figurewidth]{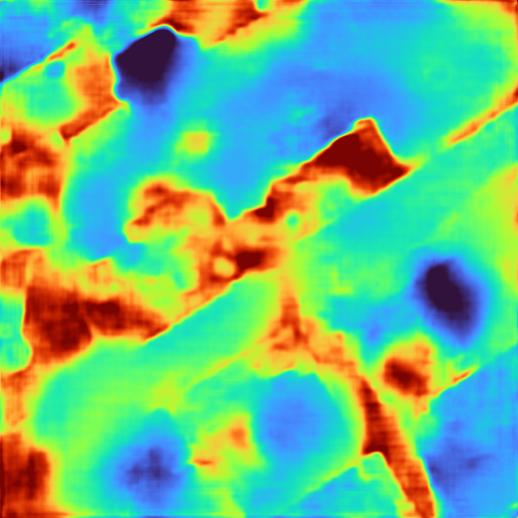}};
    \node[image, right=0.0cm of img-12] (img-13) {\includegraphics[width=\figurewidth]{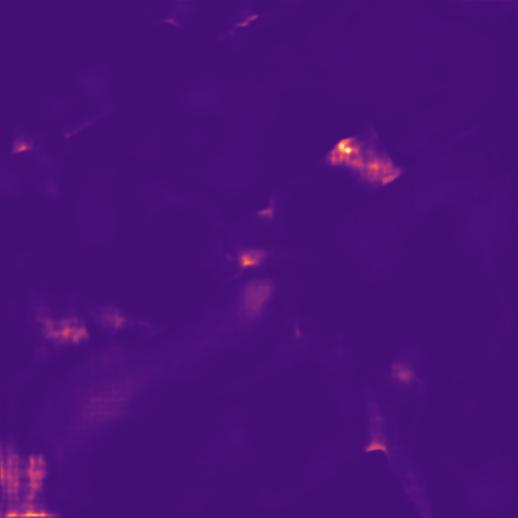}};
    \node[image, right=of img-13] (img-14) {\includegraphics[width=\figurewidth]{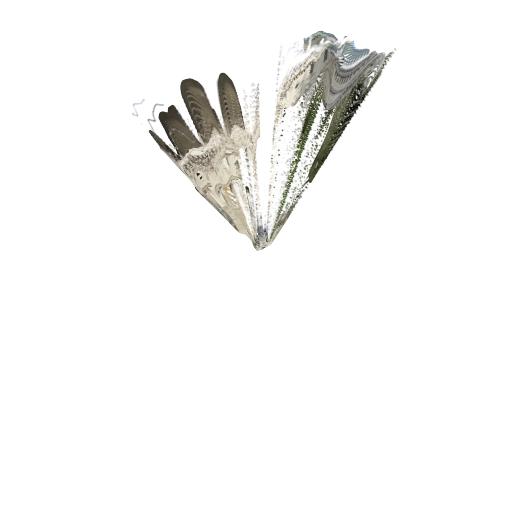}};
    \node[image, right=of img-14] (img-15) {\includegraphics[width=\figurewidth]{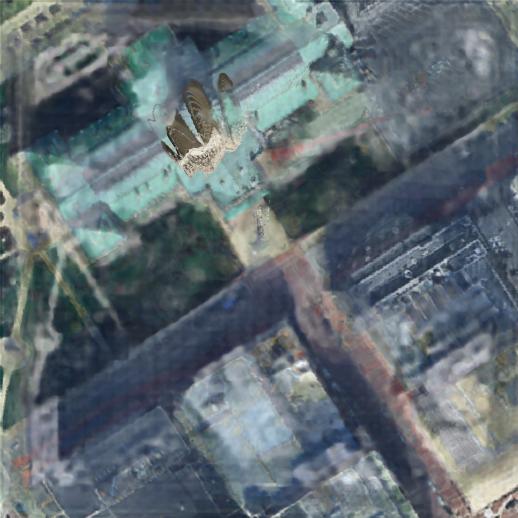}};
    \node[image, right=of img-15] (img-16) {\includegraphics[width=\figurewidth]{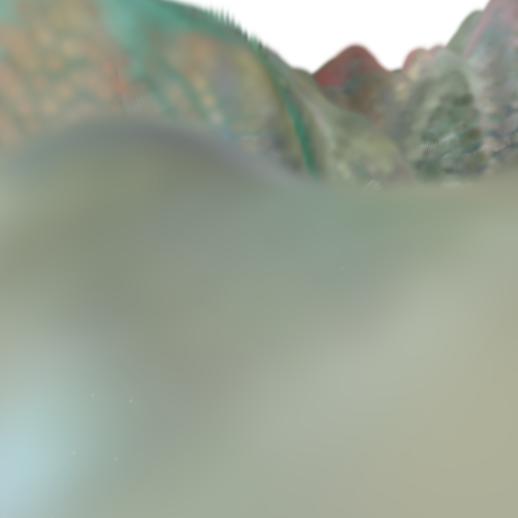}};

    %% Row 3
    \node[image, below=of img-10] (img-20) {\includegraphics[width=\figurewidth]{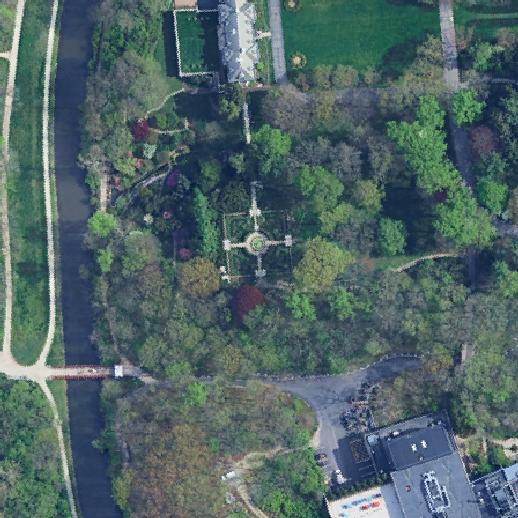}};
    \node[image, right=of img-20] (img-21) {\includegraphics[width=\figurewidth]{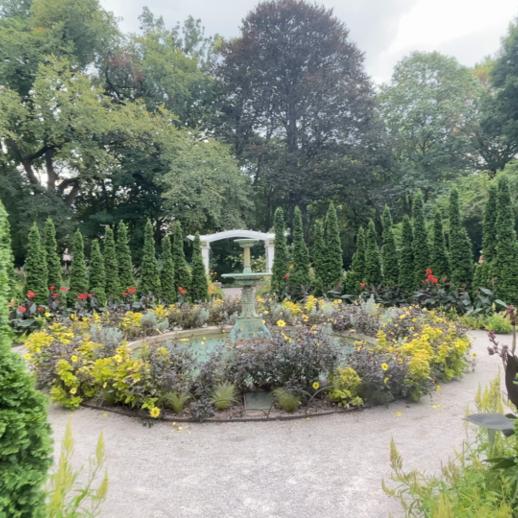}};
    \node[image, right=0.2cm of img-21] (img-22) {\includegraphics[width=\figurewidth]{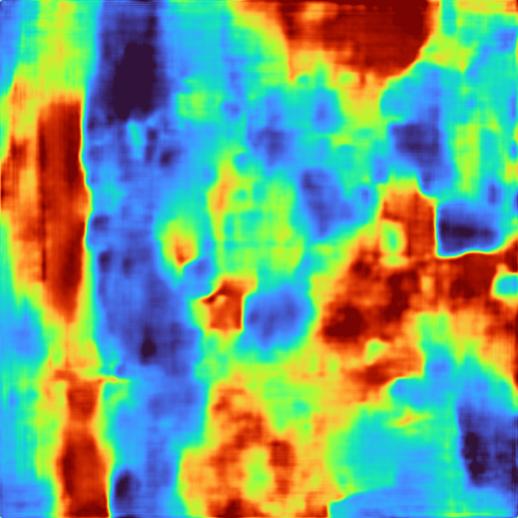}};
    \node[image, right=0.0cm of img-22] (img-23) {\includegraphics[width=\figurewidth]{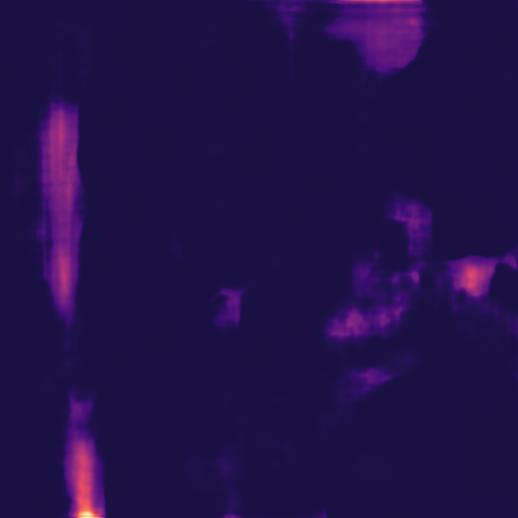}};
    \node[image, right=of img-23] (img-24) {\includegraphics[width=\figurewidth]{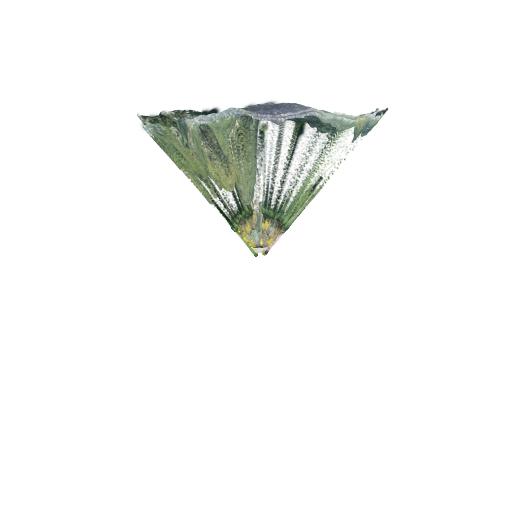}};
    \node[image, right=of img-24] (img-25) {\includegraphics[width=\figurewidth]{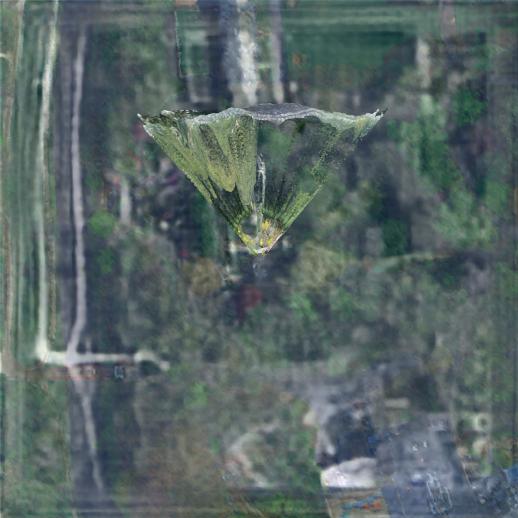}};
    \node[image, right=of img-25] (img-26) {\includegraphics[width=\figurewidth]{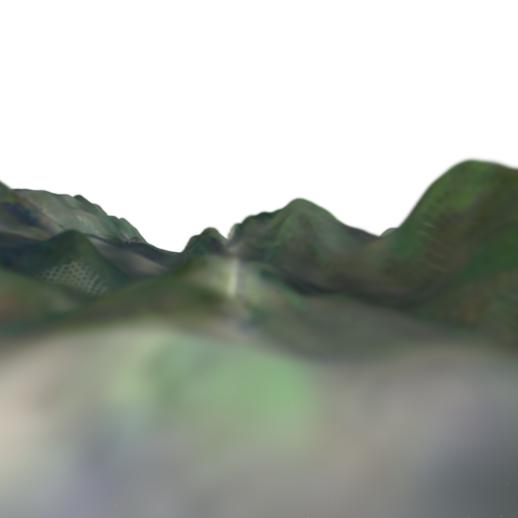}};

    %% Row 4
    \node[image, below=of img-20] (img-30) {\includegraphics[width=\figurewidth]{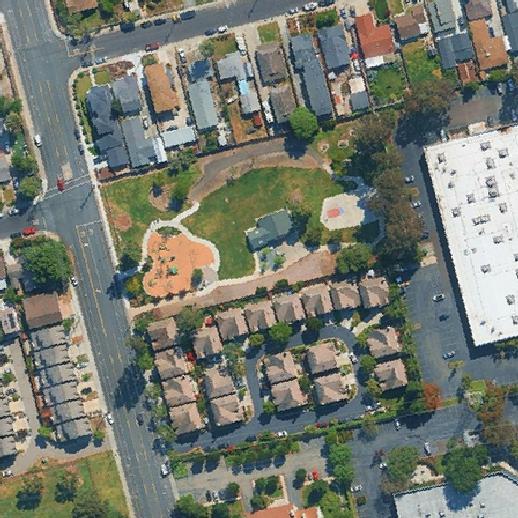}};
    \node[image, right=of img-30] (img-31) {\includegraphics[width=\figurewidth]{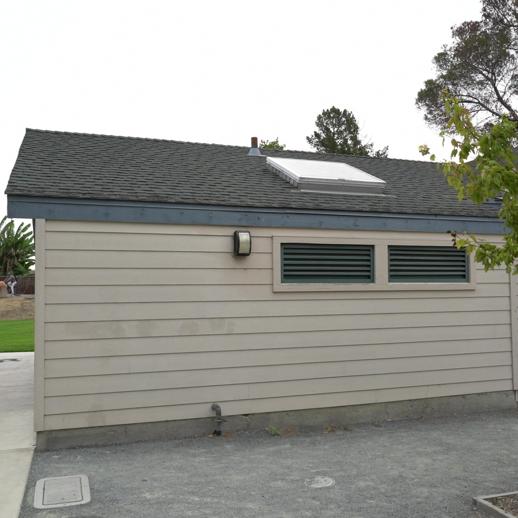}};
    \node[image, right=0.2cm of img-31] (img-32) {\includegraphics[width=\figurewidth]{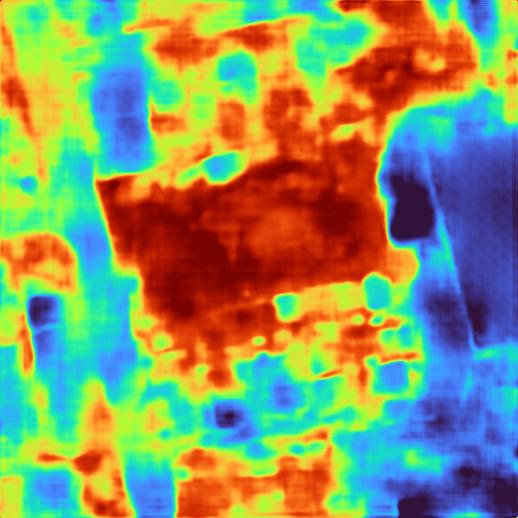}};
    \node[image, right=0.0cm of img-32] (img-33) {\includegraphics[width=\figurewidth]{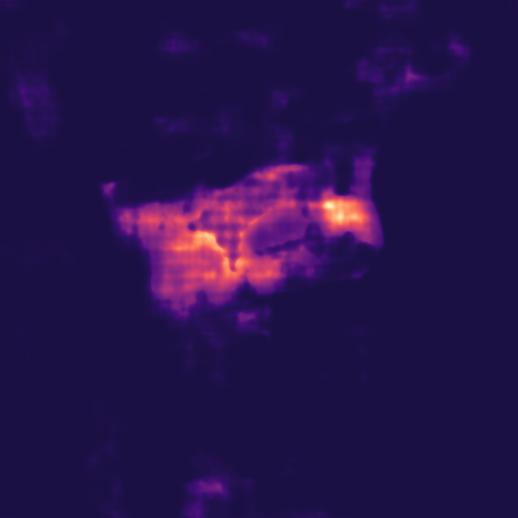}};
    \node[image, right=of img-33] (img-34) {\includegraphics[width=\figurewidth]{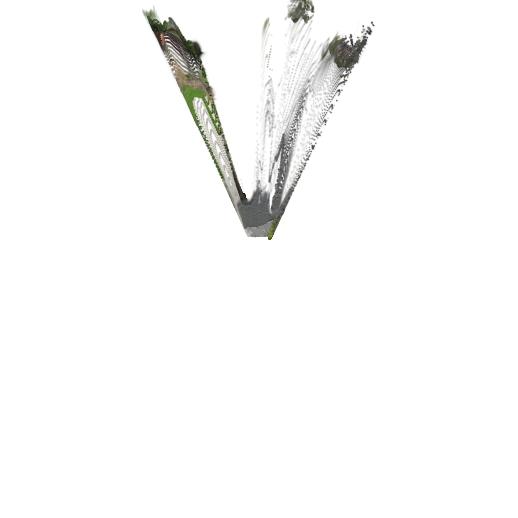}};
    \node[image, right=of img-34] (img-35) {\includegraphics[width=\figurewidth]{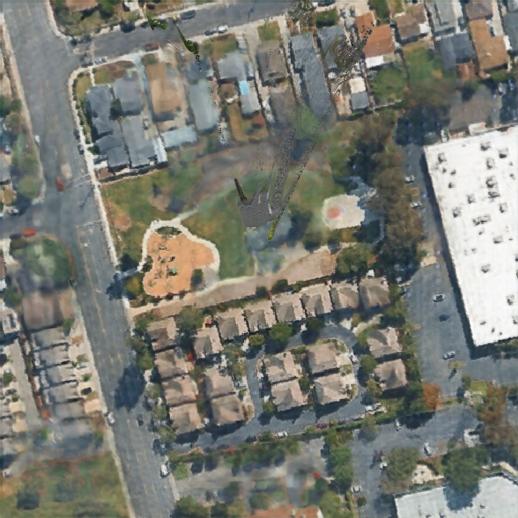}};
    \node[image, right=of img-35] (img-36) {\includegraphics[width=\figurewidth]{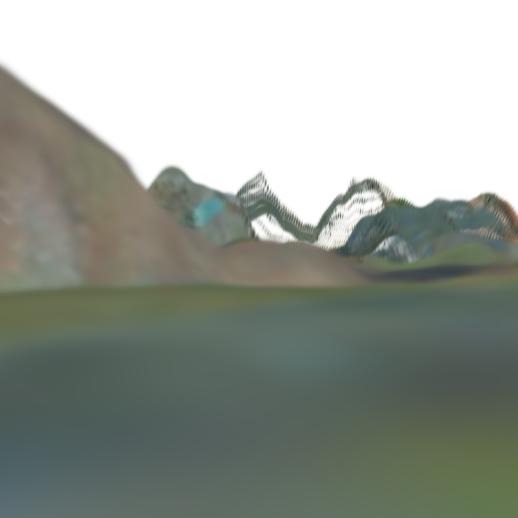}};
    
    %% Row 5
    \node[image, below=of img-30] (img-40) {\includegraphics[width=\figurewidth]{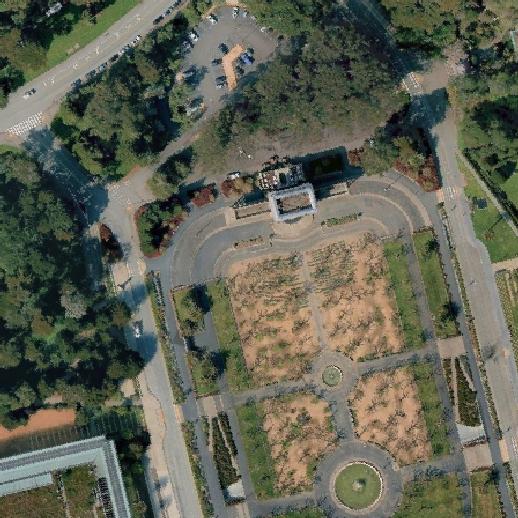}};
    \node[image, right=of img-40] (img-41) {\includegraphics[width=\figurewidth]{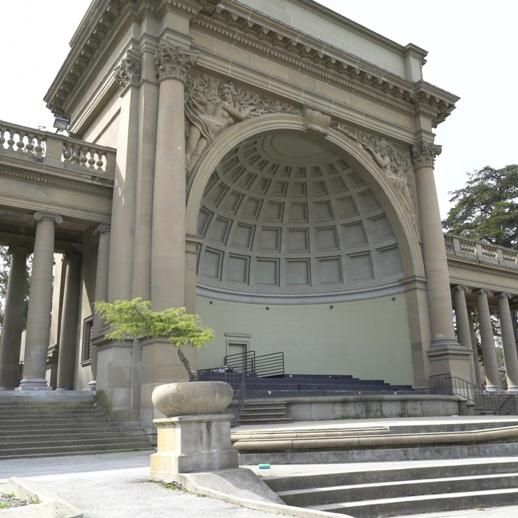}};
    \node[image, right=0.2cm of img-41] (img-42) {\includegraphics[width=\figurewidth]{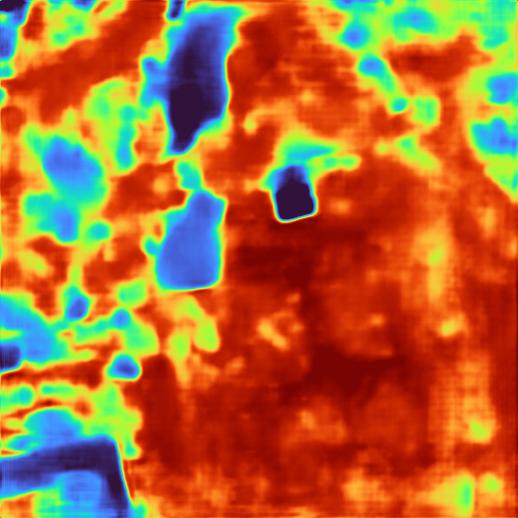}};
    \node[image, right=0.0cm of img-42] (img-43) {\includegraphics[width=\figurewidth]{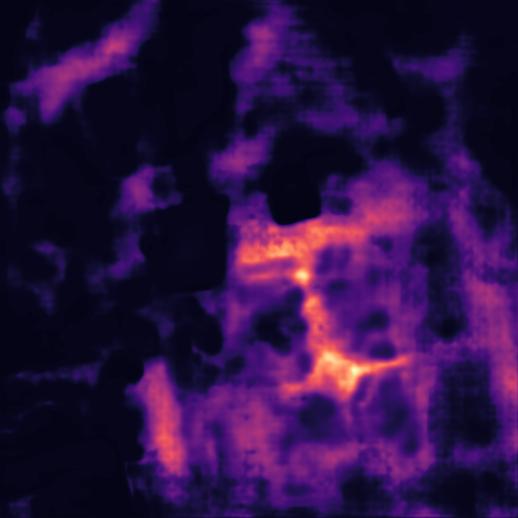}};
    \node[image, right=of img-43] (img-44) {\includegraphics[width=\figurewidth]{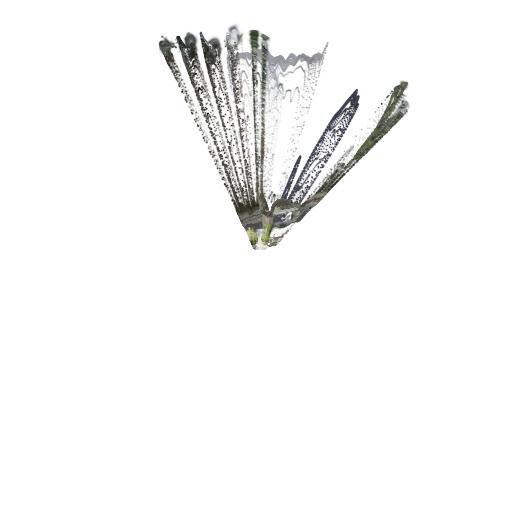}};
    \node[image, right=of img-44] (img-45) {\includegraphics[width=\figurewidth]{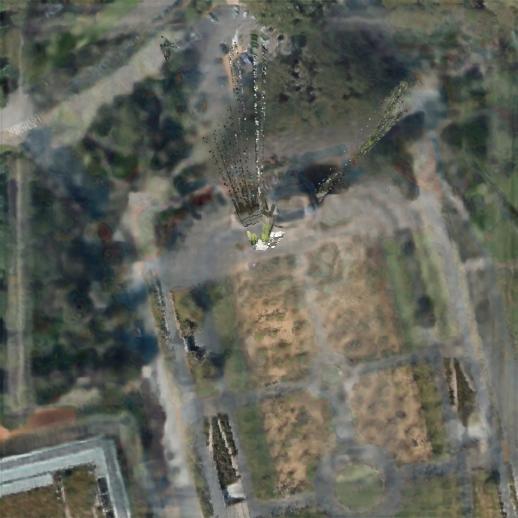}};
    \node[image, right=of img-45] (img-46) {\includegraphics[width=\figurewidth]{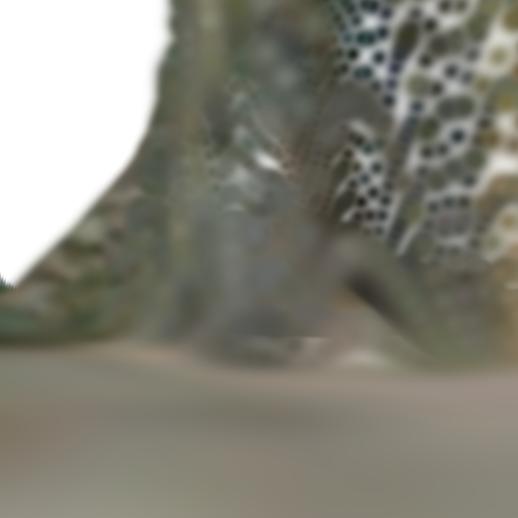}};
    
    %% Row 6
    \node[image, below=of img-40] (img-50) {\includegraphics[width=\figurewidth]{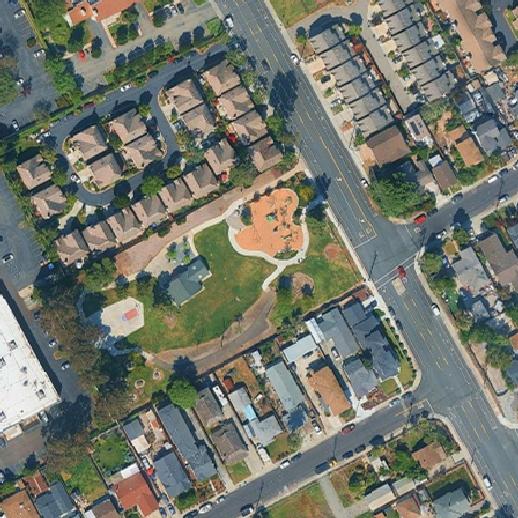}};
    \node[image, right=of img-50] (img-51) {\includegraphics[width=\figurewidth]{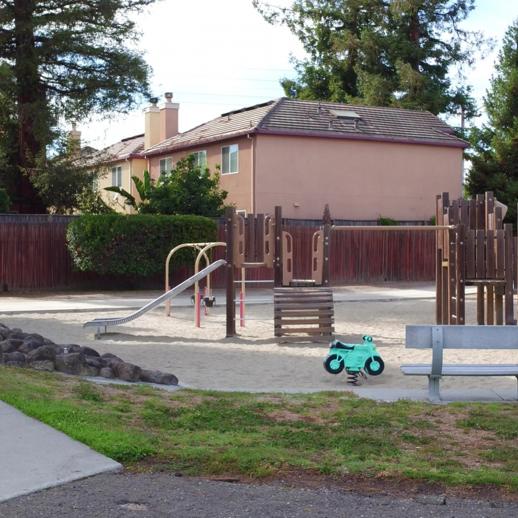}};
    \node[image, right=0.2cm of img-51] (img-52) {\includegraphics[width=\figurewidth]{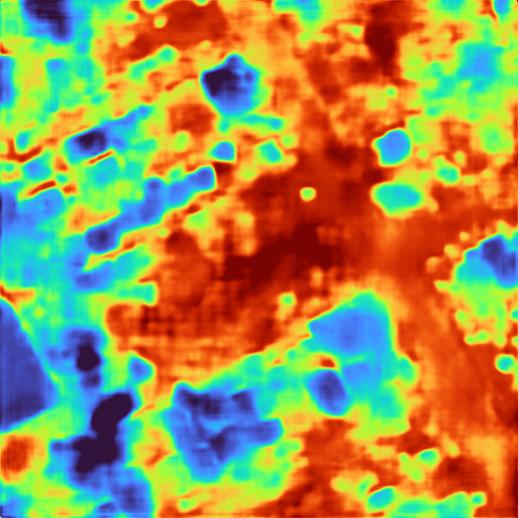}};
    \node[image, right=0.0cm of img-52] (img-53) {\includegraphics[width=\figurewidth]{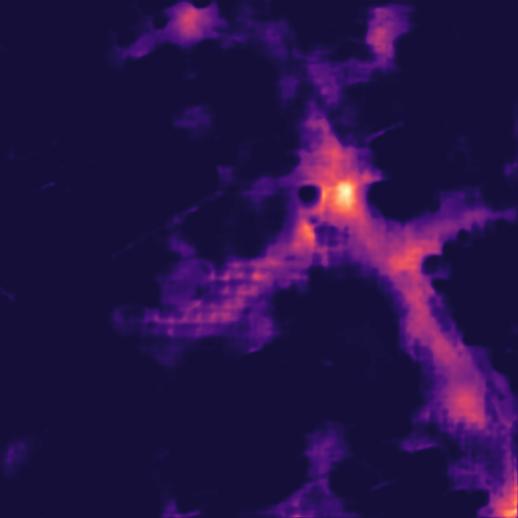}};
    \node[image, right=of img-53] (img-54) {\includegraphics[width=\figurewidth]{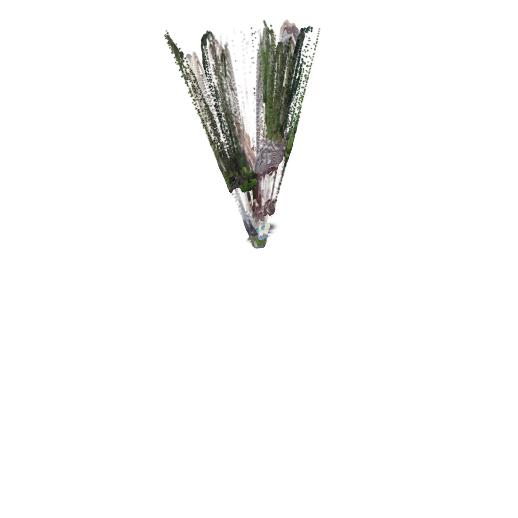}};
    \node[image, right=of img-54] (img-55) {\includegraphics[width=\figurewidth]{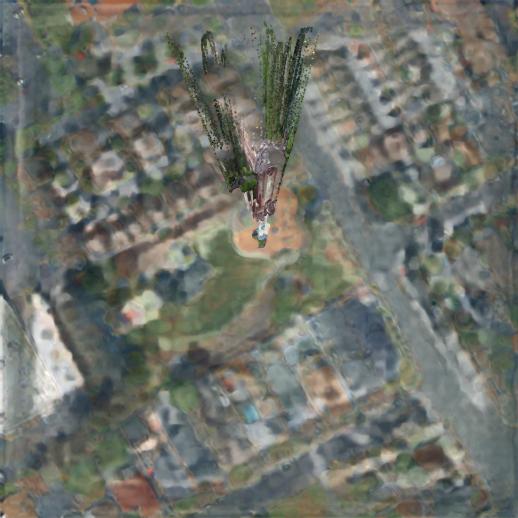}};
    \node[image, right=of img-55] (img-56) {\includegraphics[width=\figurewidth]{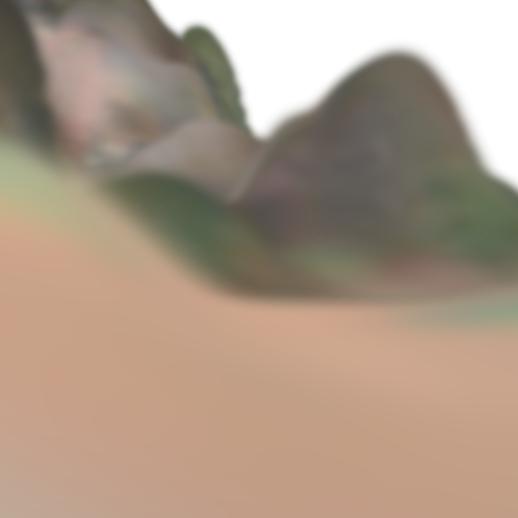}};

    % Labels
    \node[label, yshift=0.18cm] at (img-00.north) {Satellite};
    \node[label, yshift=0.18cm] at (img-01.north) {Ground Image};
    \node[label, yshift=0.18cm] at (img-02.north) {Pred Height};
    \node[label, yshift=0.18cm] at (img-03.north) {Confidence};
    \node[label, yshift=0.2cm] at (img-04.north) {$\mathcal{G}^{\text{ground}}$};
    \node[label, yshift=0.2cm] at (img-05.north) {$\mathcal{G}^{\text{combined}}$};
    \node[label, yshift=0.18cm] at (img-06.north) {Sat-to-Ground};

    % Vertical separator
    % midpoint between img-01 and img-02
    \coordinate (sep) at ($ (img-01.east)!0.5!(img-02.west) $);
    \draw[densely dashed, gray!70, line width=0.4pt]
        ([yshift=1.2ex]sep |- img-00.north) --
        ([yshift=-1.2ex]sep |- img-56.south);

    \end{tikzpicture}
\caption{\textbf{\ours satellite qualitative}. We show visuals of our full-model satellite head predictions on our benchmark scenes. The first two rows are the inputs to the model, i.e. a BEV perspective and a ground level image. We predict height maps, confidence values, ground level splats, and satellite splats that can then be rendered to ground level views.}
\label{fig:qualitative_terrain}
\end{figure*}

\begin{figure*}[!t]
    \centering
    \begin{tikzpicture}[
        image/.style = {
            inner sep=0pt,
            outer sep=0pt,
            anchor=north west
        },
        node distance = 1pt and 1pt
    ]
    \setlength{\figurewidth}{0.157\textwidth}

    % ----------------------------------------------------
    % New Row 1 (Scene: 513e4ea2e8477b06)
    % ----------------------------------------------------
    % Col 1: Our Height (formerly img-12)
    \node[image] (img-00) {\includegraphics[width=\figurewidth]{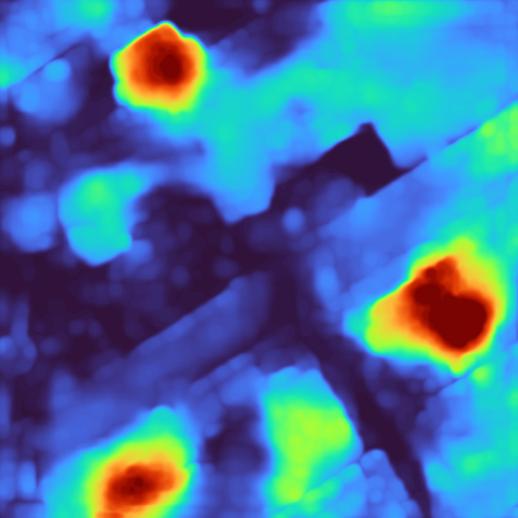}};
    % Col 2: Sat2Density Height (formerly img-13)
    \node[image, right=of img-00] (img-01) {\includegraphics[width=\figurewidth]{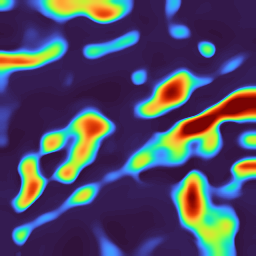}};
    % Col 3: Ours Sat2Grnd (formerly img-14)
    \node[image, right=0.0cm of img-01] (img-02) {\includegraphics[width=\figurewidth]{images/supp/terrain_qualitative/513e4ea2e8477b06/context_0.jpg}};
    % Col 4: Sat2Density Sat2Grnd (formerly img-15)
    \node[image, right=0cm of img-02] (img-03) {\includegraphics[width=\figurewidth]{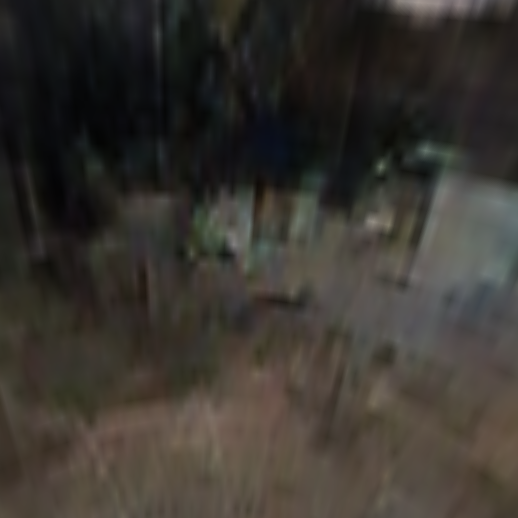}};
    % Col 5: Satellite (formerly img-10)
    \node[image, right=0.2cm of img-03] (img-04) {\includegraphics[width=\figurewidth]{images/supp/terrain_qualitative/513e4ea2e8477b06/sat.jpg}};
    % Col 6: Ground Image (formerly img-11)
    \node[image, right=of img-04] (img-05) {\includegraphics[width=\figurewidth]{images/supp/terrain_qualitative/513e4ea2e8477b06/context_0_gt.jpg}};

    % ----------------------------------------------------
    % New Row 2 (Scene: 1264931635e127)
    % ----------------------------------------------------
    % Col 1: Our Height (formerly img-12)
    \node[image, below=of img-00] (img-10) {\includegraphics[width=\figurewidth]{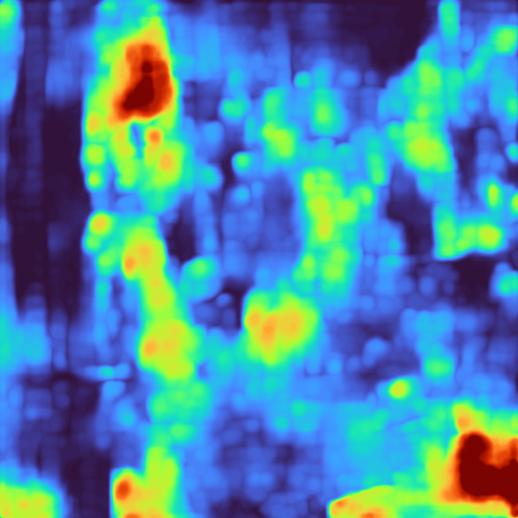}};
    % Col 2: Sat2Density Height (formerly img-13)
    \node[image, right=of img-10] (img-11) {\includegraphics[width=\figurewidth]{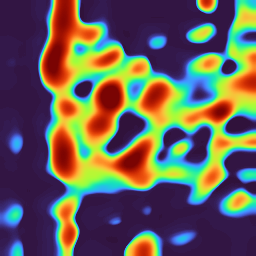}};
    % Col 3: Ours Sat2Grnd (formerly img-14)
    \node[image, right=0.0cm of img-11] (img-12) {\includegraphics[width=\figurewidth]{images/supp/terrain_qualitative/1264931635e127/context_0.jpg}};
    % Col 4: Sat2Density Sat2Grnd (formerly img-15)
    \node[image, right=0cm of img-12] (img-13) {\includegraphics[width=\figurewidth]{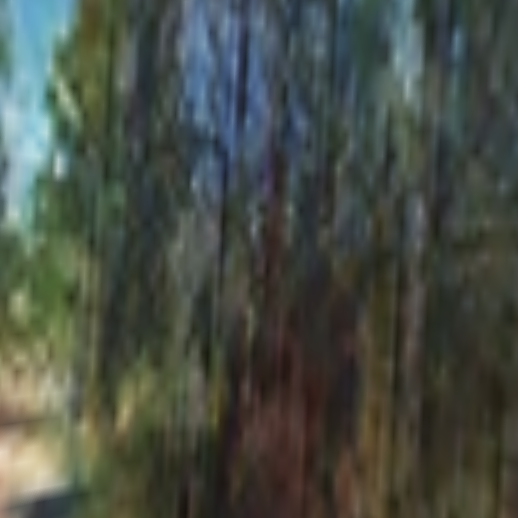}};
    % Col 5: Satellite (formerly img-10)
    \node[image, right=0.2cm of img-13] (img-14) {\includegraphics[width=\figurewidth]{images/supp/terrain_qualitative/1264931635e127/sat.jpg}};
    % Col 6: Ground Image (formerly img-11)
    \node[image, right=of img-14] (img-15) {\includegraphics[width=\figurewidth]{images/supp/terrain_qualitative/1264931635e127/context_0_gt.jpg}};

    % ----------------------------------------------------
    % New Row 3 (Scene: barn)
    % ----------------------------------------------------
    % Col 1: Our Height (formerly img-22)
    \node[image, below=of img-10] (img-20) {\includegraphics[width=\figurewidth]{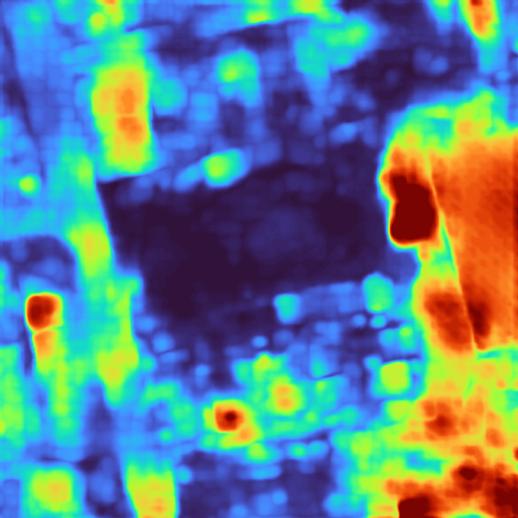}};
    % Col 2: Sat2Density Height (formerly img-23)
    \node[image, right=of img-20] (img-21) {\includegraphics[width=\figurewidth, viewport=21 21 234 232, clip]{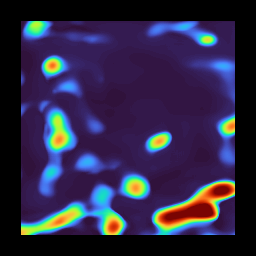}};
    % Col 3: Ours Sat2Grnd (formerly img-24)
    \node[image, right=0.0cm of img-21] (img-22) {\includegraphics[width=\figurewidth]{images/supp/terrain_qualitative/barn/context_0.jpg}};
    % Col 4: Sat2Density Sat2Grnd (formerly img-25)
    \node[image, right=0cm of img-22] (img-23) {\includegraphics[width=\figurewidth]{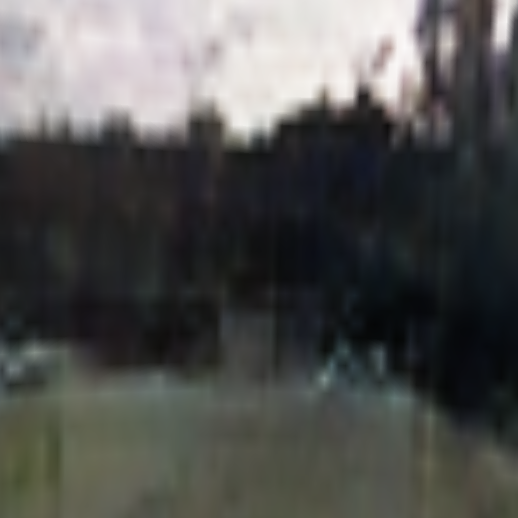}};
    % Col 5: Satellite (formerly img-20)
    \node[image, right=0.2cm of img-23] (img-24) {\includegraphics[width=\figurewidth]{images/supp/terrain_qualitative/barn/sat.jpg}};
    % Col 6: Ground Image (formerly img-21)
    \node[image, right=of img-24] (img-25) {\includegraphics[width=\figurewidth]{images/supp/terrain_qualitative/barn/context_0_gt.jpg}};

    % ----------------------------------------------------
    % New Row 4 (Scene: temple)
    % ----------------------------------------------------
    % Col 1: Our Height (formerly img-32)
    \node[image, below=of img-20] (img-30) {\includegraphics[width=\figurewidth]{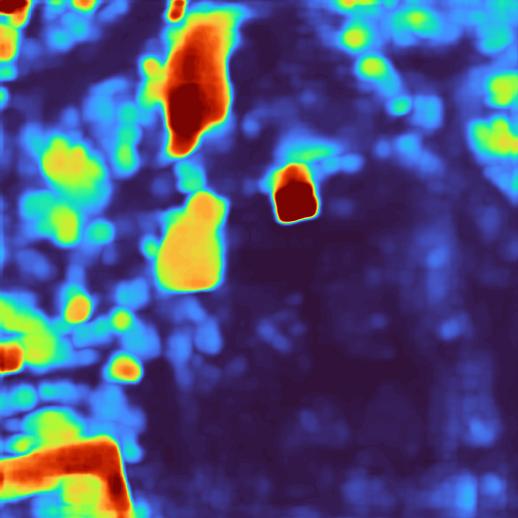}};
    % Col 2: Sat2Density Height (formerly img-33)
    \node[image, right=of img-30] (img-31) {\includegraphics[width=\figurewidth, viewport=21 21 234 232, clip]{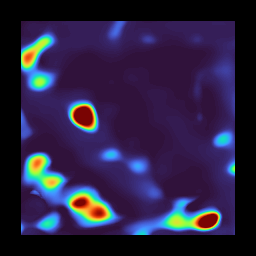}};
    % Col 3: Ours Sat2Grnd (formerly img-34)
    \node[image, right=0.0cm of img-31] (img-32) {\includegraphics[width=\figurewidth]{images/supp/terrain_qualitative/temple/context_6.jpg}};
    % Col 4: Sat2Density Sat2Grnd (formerly img-35)
    \node[image, right=0cm of img-32] (img-33) {\includegraphics[width=\figurewidth]{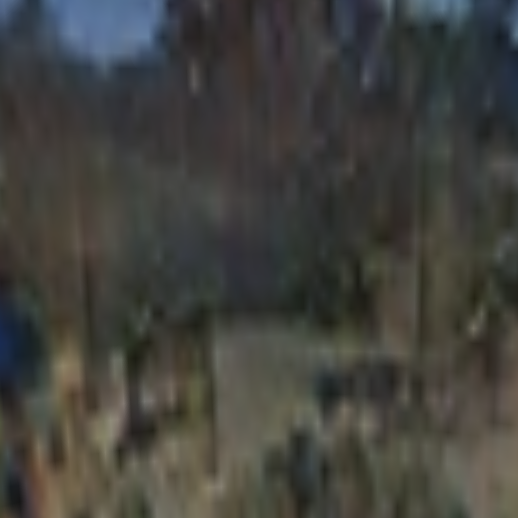}};
    % Col 5: Satellite (formerly img-30)
    \node[image, right=0.2cm of img-33] (img-34) {\includegraphics[width=\figurewidth]{images/supp/terrain_qualitative/temple/sat.jpg}};
    % Col 6: Ground Image (formerly img-31)
    \node[image, right=of img-34] (img-35) {\includegraphics[width=\figurewidth]{images/supp/terrain_qualitative/temple/context_6_gt.jpg}};

    % ----------------------------------------------------
    % New Row 5 (Scene: playground)
    % ----------------------------------------------------
    % Col 1: Our Height (formerly img-42)
    \node[image, below=of img-30] (img-40) {\includegraphics[width=\figurewidth]{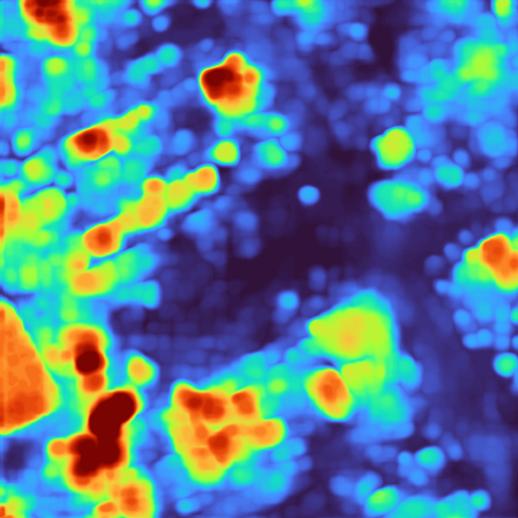}};
    % Col 2: Sat2Density Height (formerly img-43)
    \node[image, right=of img-40] (img-41) {\includegraphics[width=\figurewidth, viewport=21 21 234 232, clip]{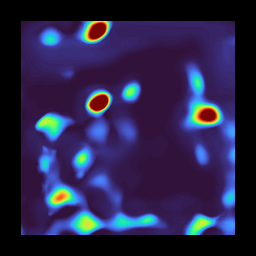}};
    % Col 3: Ours Sat2Grnd (formerly img-44)
    \node[image, right=0.0cm of img-41] (img-42) {\includegraphics[width=\figurewidth]{images/supp/terrain_qualitative/playground/context_0.jpg}};
    % Col 4: Sat2Density Sat2Grnd (formerly img-45)
    \node[image, right=0cm of img-42] (img-43) {\includegraphics[width=\figurewidth]{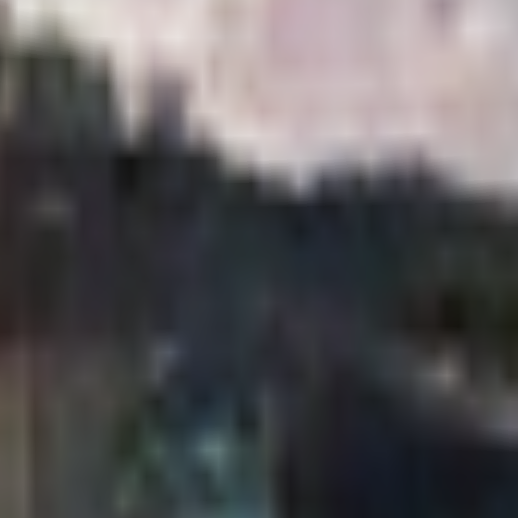}};
    % Col 5: Satellite (formerly img-40)
    \node[image, right=0.2cm of img-43] (img-44) {\includegraphics[width=\figurewidth]{images/supp/terrain_qualitative/playground/sat.jpg}};
    % Col 6: Ground Image (formerly img-41)
    \node[image, right=of img-44] (img-45) {\includegraphics[width=\figurewidth]{images/supp/terrain_qualitative/playground/context_0_gt.jpg}};

    % ----------------------------------------------------
    % New Row 6 (Scene: cc08c0bdc34ddd)
    % ----------------------------------------------------
    % Col 1: Our Height (formerly img-52)
    \node[image, below=of img-40] (img-50) {\includegraphics[width=\figurewidth]{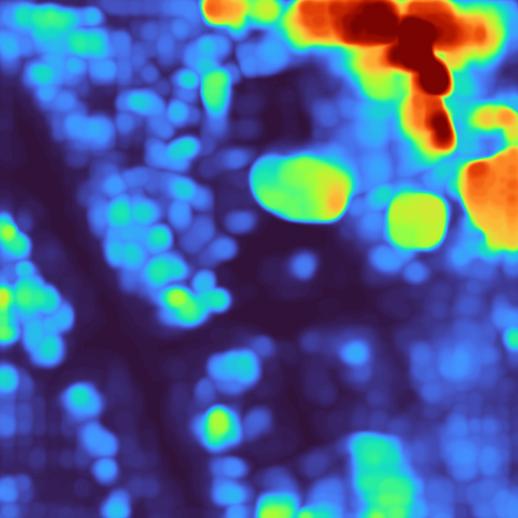}};
    % Col 2: Sat2Density Height (formerly img-53)
    \node[image, right=of img-50] (img-51) {\includegraphics[width=\figurewidth]{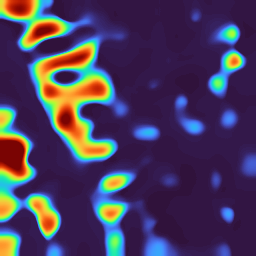}};
    % Col 3: Ours Sat2Grnd (formerly img-54)
    \node[image, right=0.0cm of img-51] (img-52) {\includegraphics[width=\figurewidth]{images/supp/terrain_qualitative/cc08c0bdc34ddd/context_0.jpg}};
    % Col 4: Sat2Density Sat2Grnd (formerly img-55)
    \node[image, right=0cm of img-52] (img-53) {\includegraphics[width=\figurewidth]{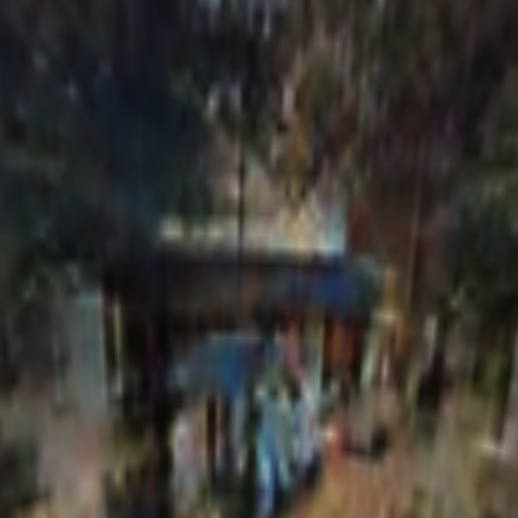}};
    % Col 5: Satellite (formerly img-50)
    \node[image, right=0.2cm of img-53] (img-54) {\includegraphics[width=\figurewidth]{images/supp/terrain_qualitative/cc08c0bdc34ddd/sat.jpg}};
    % Col 6: Ground Image (formerly img-51)
    \node[image, right=of img-54] (img-55) {\includegraphics[width=\figurewidth]{images/supp/terrain_qualitative/cc08c0bdc34ddd/context_0_gt.jpg}};
    
    % ---------------------------
    % Column Labels - UPDATED to reflect the new order
    % ---------------------------
    \node[label, yshift=0.18cm] at (img-00.north) {Ours};
    \node[label, yshift=0.18cm] at (img-01.north) {Sat2Density};
    \node[label, yshift=0.18cm] at (img-02.north) {Ours};
    \node[label, yshift=0.18cm] at (img-03.north) {Sat2Density};
    \node[label, yshift=0.18cm] at (img-04.north) {Satellite};
    \node[label, yshift=0.18cm] at (img-05.north) {Ground Image};

    % Group Labels (placed higher up, e.g., yshift=0.7cm)
    \node at ($ (img-00.north)!0.5!(img-01.north) $)[yshift=0.7cm] {\textbf{Height Predictions}};
    \node at ($ (img-02.north)!0.5!(img-03.north) $)[yshift=0.7cm] {\textbf{Sat2Ground Predictions}};
    \node at ($ (img-04.north)!0.5!(img-05.north) $)[yshift=0.7cm] {\textbf{GT}};

    % Vertical separator (position remains the same relative to the nodes)
    
    \coordinate (sep_new) at ($ (img-03.east)!0.5!(img-04.west) $);
    \draw[densely dashed, black, line width=1.0pt] % Thicker line for main separation
        ([yshift=1.2ex]sep_new |- img-00.north) --
        ([yshift=-1.2ex]sep_new |- img-54.south);
    \end{tikzpicture}

\caption{\textbf{Sat2Density~\cite{sat2density} comparison.} We compare our predictions (Columns 1-4) with Sat2Density height estimates and Sat2Density ground renders against the Ground Truth inputs (Columns 5-6). Both models get the same satellite image and ground image as inputs.}
\label{fig:sat2density_qualitative}
\end{figure*}

We show more qualitative renders from our model. In \cref{fig:baselines_qualitative} we show more comparisons to baseline methods. In \cref{fig:qualitative_ours_vs_ground} we visualize side-by-side comparisons of our ground only model (referred to as Ground in the main paper) and our satellite enabled full model (referred to as Combined in the main paper) to visually demonstrate the improved coverage and completeness obtained from our full model. In \cref{fig:qualitative_terrain} we visualize the outputs from the satellite branch independently. Finally, in \cref{fig:sat2density_qualitative} we compare our satellite branch predictions to those obtained from Sat2Density \cite{sat2density}. We observe that we obtain more detailed BEV height map estimates as well as more realistic geometry when we render ground views from the height estimates. 

\begin{figure}[!th]
    \centering
    \begin{tikzpicture}[
        image/.style = {
            inner sep=0pt,
            outer sep=0pt,
            anchor=north west
        },
        node distance = 1pt and 1pt
    ]
    \setlength{\figurewidth}{0.2\textwidth}
    %% Row 1
    \node[image] (img-00) {\includegraphics[width=\figurewidth,clip,trim=400 400 400 400]{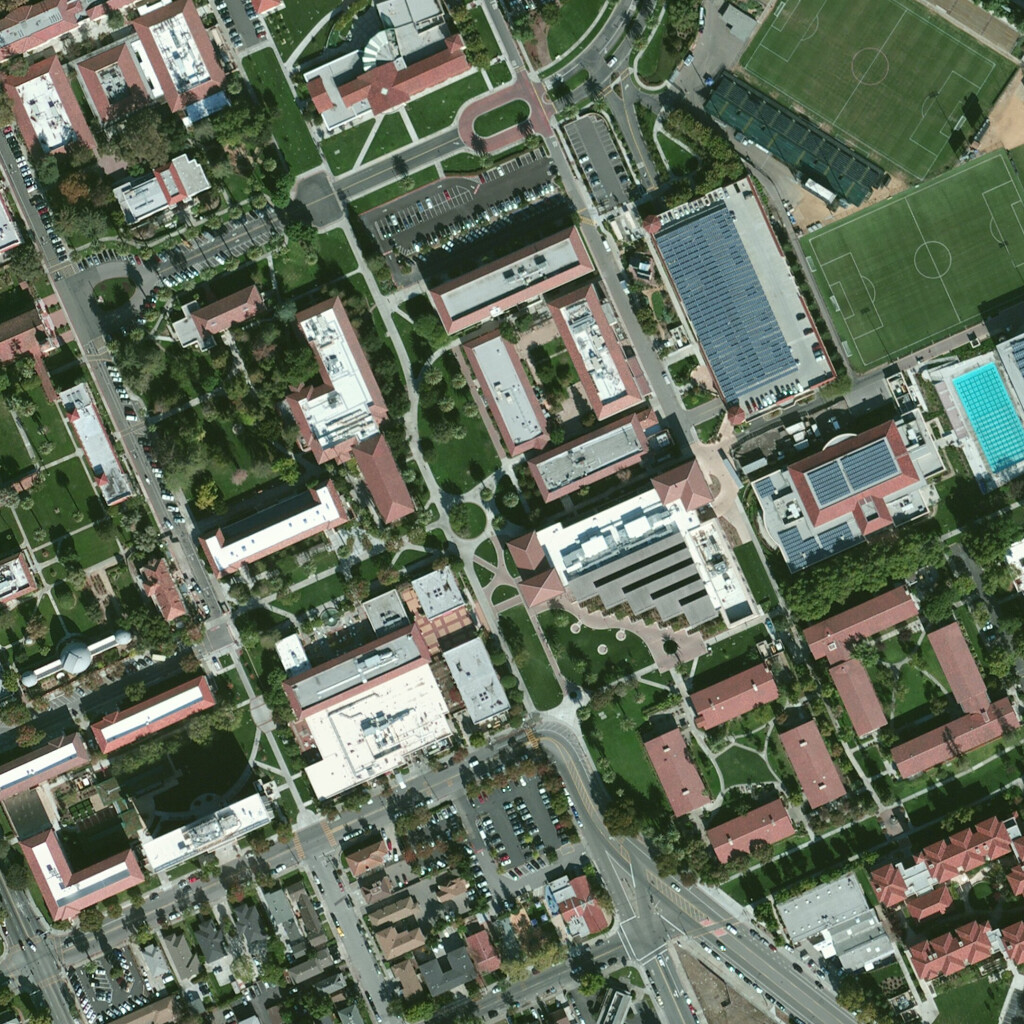}};
    \node[image, right=of img-00] (img-01) {\includegraphics[width=\figurewidth,clip,trim=400 400 400 400]{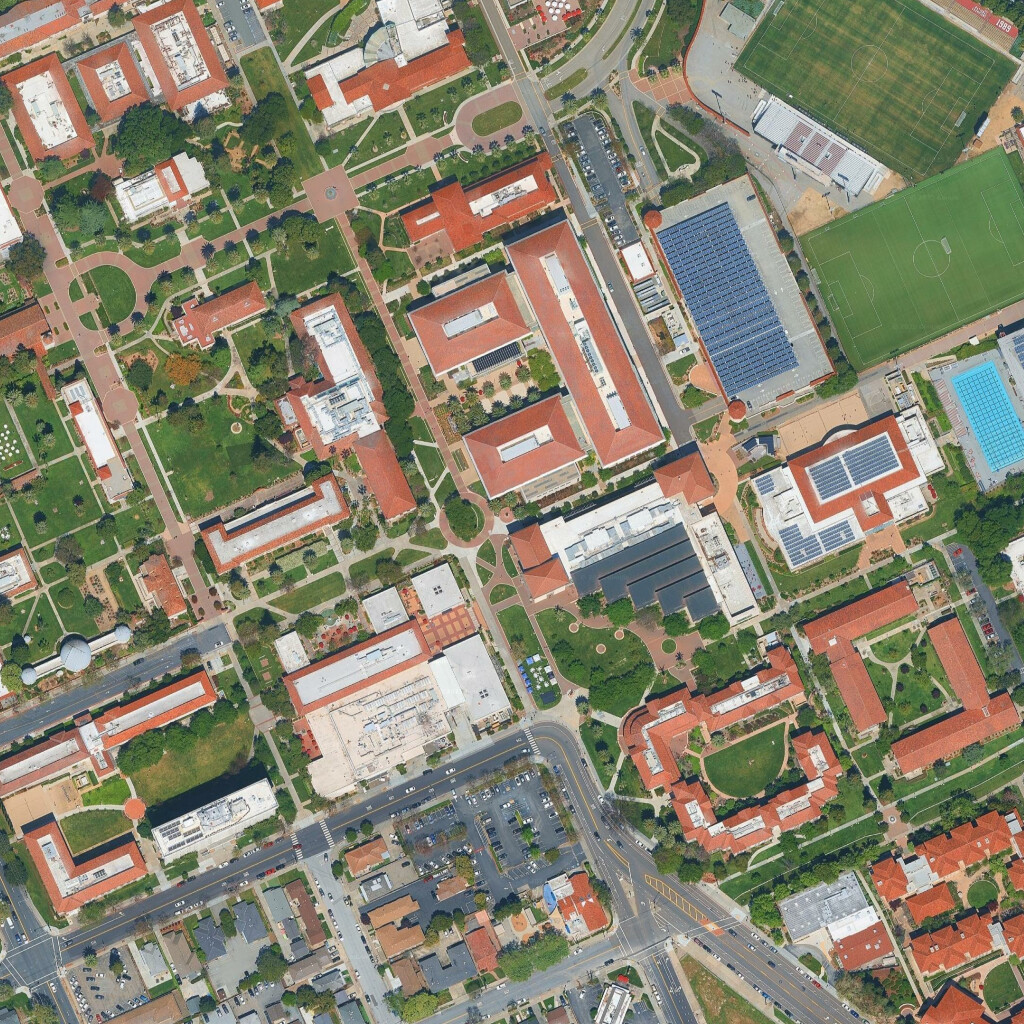}};

    \end{tikzpicture}
    \caption{\textbf{Limitations of satellite imagery}. Notice how a building has been rebuilt and expanded in the right frame compared to the left taken a few years ago. This is Family scene in Tanks and Temples.}
    \label{fig:limitations}
    \vspace{-1em}
\end{figure}

\section{Limitations}
Our method struggles in scenarios where the ground-level camera observes directions that fall outside the satellite orthographic view. For example, when looking upward toward the sky or downward at the ground. Because these look-at directions are not seen in the BEV views, their geometry cannot be reliably inferred. The approach is also unsuitable for environments that are not visible from above, such as indoor scenes, tunnels, underpasses, or structures with significant overhangs. Additionally, in small-baseline, high-overlap input views, the advantages of incorporating satellite information diminishes, since ground-level geometry alone already provides sufficient coverage. Finally, our BEV training data is sourced primarily from USA city regions, which limits the model’s generalization to geographic areas with different styles, satellite characteristics, or geography. There can also be inconsistencies in the satellite imagery itself; for example, if there has been a long temporal gap between satellite acquisition and ground-level capture. We illustrate such a case in \cref{fig:limitations}.

%% Sat2Density Visual - FINAL MODIFICATION

\clearpage
%%%%%%%%% REFERENCES
%{\small
%\bibliographystyle{ieee_fullname}
%\bibliography{egbib}
%}

\end{document}